\documentclass{article}

\usepackage{arxiv}

\usepackage{booktabs} % For formal tables
\usepackage[utf8]{inputenc}
\usepackage{subcaption}
\usepackage{csquotes}
\usepackage{xcolor}
\usepackage{graphicx}
\usepackage{amsmath}
\usepackage{algorithm}
\usepackage[noend]{algorithmic}
\usepackage{float}
\usepackage{soul}
\usepackage{natbib}
\usepackage{tikz}
\usepackage{pgfplots}
\usepackage{multirow}
\sethlcolor{lightgray}

\newcommand{\squeezeup}{\vspace{-1mm}}
\DeclareMathOperator*{\argmax}{arg\,max}
\DeclareMathOperator*{\argmin}{arg\,min}

\title{On the Use of Quality Diversity Algorithms for The Traveling Thief Problem}

\author{
 Adel Nikfarjam \\
Optimisation and Logistics\\School of Computer and Mathematical Sciences\\The University of Adelaide\\
  \texttt{adel.nikfarjam@adelaide.edu.au} \\
  \And
 Aneta Neumann \\
Optimisation and Logistics\\School of Computer and Mathematical Sciences\\The University of Adelaide\\
  \texttt{aneta.neumann@adelaide.edu.au} \\
    \And
 Frank Neumann \\
Optimisation and Logistics\\School of Computer and Mathematical Sciences\\The University of Adelaide\\
  \texttt{frank.neumann@adelaide.edu.au} \\}
\begin{document}
\maketitle

\begin{abstract}
%\frank{We agreed on TTP as the focus and QD to provide 1) insights into high %quality solutions in relation to TSP and DP score 2) better results for TTP. You %would have to write the paper with this focus. TTP first and then QD as an %approach to TTP to achieve 1) and 2)} \adel{done}

In real-world optimisation, it is common to face several sub-problems interacting and forming the main problem. There is an inter-dependency between the sub-problems, making it impossible to solve such a problem by focusing on only one component. The traveling thief problem~(TTP) belongs to this category and is formed by the integration of the traveling salesperson problem~(TSP) and the knapsack problem~(KP). In this paper, we investigate the inter-dependency of the TSP and the KP by means of quality diversity~(QD) approaches. QD algorithms provide a powerful tool not only to obtain high-quality solutions but also to illustrate the distribution of high-performing solutions in the behavioural space. 
%For the TTin this case the underlying scores of the TSP and KP solutions.    
% While the classic optimisation algorithms compute a single (near)-optimal solution, evolving a diverse set of high quality solutions has gained considerable attention in the community of evolutionary computation. Quality Diversity is a recently added paradigm in this area. Quality Diversity not only focuses on evolving a set of high performing solutions but also sheds a light on the search space by providing a view of the distribution of high-quality solutions in the behavioral space. However, Quality Diversity is only used in the continuous domain.  For the first time, we study a combinatorial optimisation problem, Traveling Thief Problem, in the context of Quality Diversity. Traveling Thief problem is a multi-component problem where two sub-problems inter-dependently forming the problem. Qaulity Diversity is a beneficial tool to study multi-component problems and investigate the inter-dependency by defining the sub-problems as behavioral descriptors.
We introduce a MAP-Elite based evolutionary algorithm using well-known TSP and KP search operators, taking the TSP and KP score as the behavioural descriptor. Afterwards, we conduct comprehensive experimental studies that show the usefulness of using the QD approach applied to the TTP. First, we provide insights regarding high-quality TTP solutions in the TSP/KP behavioural space. Afterwards, we show that better solutions for the TTP can be obtained by using our QD approach and it can improve the best-known solution for a number of TTP instances used for benchmarking in the literature.
%results found  second, to determine the performance of the algorithm in solving %TTP. The results show that the algorithm brings about very decent solutions; in %fact, it improves several best found solutions in the literature.  
\end{abstract}

%
% The code below should be generated by the tool at
% http://dl.acm.org/ccs.cfm
% Please copy and paste the code instead of the example below. 
%


% \frank{check whether MAP-elite should be capital letters. Ok leave it like this. Others such as behaviour descriptor etc should be lower case.} \adel{MAP-Elite is the name they use for the algorithm based on the map}
\section{Introduction}
\label{Sec:intro}
In many real-world optimisation problems, several NP-hard problems interact with each other. Such optimisation problems are complex due to the inter-dependencies between the sub-problems. The inter-dependencies make each sub-problem affect the quality and even the feasibility of solutions of the others. This complicates the decision-making process \cite{BonyadiM0019}. Vehicle routing problems, the traveling thief problem, and patient admission problems are examples of multi-component optimisation problems.

TTP was introduced in 2013 by \citet{BonyadiMB13}. TTP is the combination of the classical TSP and the KP. Both TSP and KP are well-known, well-studied combinatorial problems. 
In a nutshell, they integrate the TSP and the KP so that the traveling cost between two cities depends not only on the distance between the cities but also on the weight of the items collected so far.
% In nutshell, TTP compromises a thief stealing items from a number of cities (locations), each item has a corresponding weight and profit. The thief visits each city only once and carries a knapsack with limited capacity and pays rent for the knapsack proportional to travel duration. The goal is to maximise the profit. Since the thief's speed is considered non-liner dependent on the weight of knapsack (collected items), the two components, the TSP and the KP, are interdependent in the TTP.    
In recent years, several solution approaches have been introduced to TTP. This includes algorithms based on co-evolutionary strategies (\cite{BonyadiMPW14, YafraniA15}), local search heuristics (\cite{PolyakovskiyB0MN14, MaityD20}), profit-based heuristic that modifies tours based on the items' value (\cite{NamaziN0S19}),  simulated annealing~(\cite{YafraniA18}), swarm intelligence approaches (\cite{Wagner16, ZouariAT19}).  Furthermore, An adaptive surrogate model was proposed in \cite{NamaziSN020} to filter our non-promising tours. Exact methods based on dynamic programming have been introduced in~\cite{Wu0PN17}, but they are limited to solving only small instances. Moreover, \citet{WuijtsT19} investigated the fitness landscape of some small instances of TTP. They studied local search and genetic algorithms using a wide range of operators such as 2-opt, insertion, EAX, and PMX. They conclude that genetic algorithms using EAX can outperform the other algorithms under the investigation in those instances.  

In multi-component optimisation problems such as the TTP, it is beneficial to provide decision-makers with a diverse set of high-quality solutions differing in terms of the scores in the sub-problems. Such a set of solutions provides decision-makers with invaluable information about the inter-dependency of the sub-problems. It also enables them to involve their interests and choose between different alternatives.
Computing a diverse set of solutions has recently gained increasing attention in evolutionary computation literature. Traditionally, these works are dominated by research on multi-modal optimisation, which involves diversity preservation techniques such as niching. In this context, solution diversity is seen as a means to explore niches in the fitness landscape, which correspond to regions of local optima.    

%\subsection{Evolutionary Diversity Optimisation}

In contrast, evolutionary diversity optimisation ~(EDO) aims to explicitly maximise the structural diversity of the solutions, subject to quality constraints. In EDO approaches, some structural features are defined, and a measure is used to determine the diversity of a set of solutions. EDO was first introduced by \citet{ulrich2011maximizing} in the continuous domain. Afterwards, the concept has been used to generate a diverse set of images and benchmark instances for the TSP (\cite{alexander2017evolution,doi:10.1162/evcoa00274}). The star-discrepancy measure~(\cite{neumann2018discrepancy}) and indicators from evolutionary multi-objective optimisation~(\cite{neumann2019evolutionary}) have been used as diversity measures for the same problems. More recently, researchers used EDO for evolving a diverse set of high-quality solutions for combinatorial optimisation problems. Distance-based measures and entropy have been used in \cite{viet2020evolving, NikfarjamBN021a} for generating diverse sets of the TSP tours. \citet{NikfarjamB0N21b} studied the scenario that the optimal solution is unknown. In addition, the quadratic assignment problem~(\cite{DoGN021}), the minimum spanning tree problem~(\cite{Bossek021tree}), the knapsack problem~(\cite{BossekN021KP}), the optimisation of monotone sub-modular functions~(\cite{NeumannB021}), and traveling thief problem~(\cite{NikTTPEDO}) have been studied in this context.

%\subsection{Quality Diversity}

QD is another well-studied paradigm. QD focuses on exploring niches in the behavioural spaces and %In other words, QD 
seeks a set of high-quality solutions that differ in terms of a few user-defined features of interest. Having been provided with such a set of solutions, the users are able to choose the high-quality solution suiting their interests the most. QD has emerged from the concept of novelty search, where algorithms aim to find new behaviours without considering fitness (\cite{LehmanS11}). \citet{CullyM13} introduced a mechanism to only keep the best-performing solutions while seeking new behaviours. Concurrently, \citet{CluneML13} proposed a simple algorithm to plot the distribution of high-quality solutions over a feature/behavioural space. Interestingly, the proposed algorithm, named MAP-Elites, efficiently evolves behavioural repertoires. \citet{PughSSS15, PughSS16} formulated the concept of computing a diverse set of high-quality solutions differing in features or behaviours and named it QD. The paradigm has been widely applied to the areas of robotics (\cite{RakicevicCK21, ZardiniZZIF21, AllardSCC22}) and games (\cite{SteckelS21, FontaineTNH20, FontaineLKMTHN21}) as well as other continuous problems such as urban design (\cite{GalanosLYK21}). We refer interested readers to the review paper of \citet{kon21}. To the best of our knowledge, QD algorithms have not previously been  used for a combinatorial optimisation problem, and we provide the first study on this subject.

\subsection{Our contribution}

We employ the concept of QD for solving the TTP. By this means, we scrutinise the distribution of high-performing TTP solutions in the behavioural space  of the TSP and the KP and compute very high-quality solutions. We introduce a bi-level MAP-elite based evolutionary algorithm called BMBEA. The algorithm generates new solutions in a two-stage procedure. First, it generates new high-quality TSP tours from old ones by the well established EAX crossover operator~(\cite{nagata2013powerful}) for the TSP, or as an alternative by 2-OPT (\cite{croes1958method}). Second, it utilises dynamic programming (\cite{NeumannPSSW18}) or an $(1+1)$ evolutionary algorithm to compute an optimal (or near-optimal) packing list for the given TSP tour. Having generated a new solution, BMBEA applies a MAP-Elites based survival selection to achieve a diverse set of high-quality TTP solutions. To achieve diversity, MAP-Elites is applied with respect to the two-dimensional space given by the TSP and KP quality of the TTP solutions.  
We conduct a comprehensive experimental investigation to analyse and visualise the distribution of high-quality TTP solutions for different TTP instances. Furthermore, we show the capability of BMBEA to generate high-performing TTP solutions. The algorithm results in very high TTP values and improves the best-known TTP solution for some benchmark instances. These contributions are also included in the conference version that was published in the proceeding of GECCO 2022 (\cite{NikfarjamMap}).

This article extends its conference version~(\cite{NikfarjamMap}) as follows.
We propose a method that eliminates two influential input parameters that need to be tuned for each instance individually in Section \ref{subsec:(mu+1)}. Using the method requires a higher number of generations to converge compared to the previous method with the considered input values. However, it improves the results, especially for the larger instances. the experimental investigations can be found in Sections \ref{subsec:exp_map_relazed} and \ref{subsec:exp_bes_relaxed}.  We also investigate the impact of Map-Elite based survival selection. We show that the survival selection brings about a diverse set of solutions that prevents premature convergence in \ref{subsec:relaxed}, \ref{subsec:exp_map_mu+1} and  \ref{subsec:exp_bes_mu+1}. We also correct some incorrect experimental results from the conference version which were due to an implementation error.

The remainder of the paper is structured as follows. In Section \ref{Sec:prob_def}, we formally define the TTP problem. We introduce the MAP-Elites based approach for TTP and the BMBEA algorithm in Section~\ref{Sec:map}. We also propose a baseline algorithm to investigate the impact of MAP-Elitism in Section \ref{Sec:map}. We examine the high-quality TTP solutions in terms of their TSP and KP score and report on our results using BMBEA for solving the TTP are shown in Section \ref{Sec:EXP}. Finally, we finish with some concluding remarks. 

%\subsection{Traveling Thief Problem}

%In this study, we aim to empirically investigate the inter-dependency of the two %TTP's components, the TSP and the KP. For this purpose, we incorporate the Quality %Diversity~(QD) techniques.      

\section{The Traveling Thief Problem}
\label{Sec:prob_def}
The traveling thief problem (TTP) is formed by the integration of the traveling salesperson problem (TSP) and the knapsack problem (KP). %So, we first define the TSP and the KP first.
%\subsection{Traveling Salesperson Problem}
The TSP problem can be defined on a complete directed graph $G=(V, E)$ where $V$ is a set of nodes (cities) of size $n=|V|$ and $E$ is a set of pairwise edges between the nodes. 
There is a non-negative distance $d(e)$ associated with each edge $e = (u,v) \in E$. 
% Throughout the paper, we assume that TSP instances are symmetric, i.e. $d((u,v))=d((v,u))$ holds for all $u,v \in V$.
The goal is to find a permutation (tour) $x : V \to V$ that minimises the following cost function:

$$f (x) = d(x(n),x(1)) + \sum_{i=1}^{n-1} d(x(i),x(i+1)).$$

%\subsection{Knapsack Problem}

The KP is defined on a set of items $I$, where $m = |I|$. Each item $j$ has a profit $p_j$ and a weight $w_j$. In KP, the objective is to find a selection of items $y = (y_1, \cdots, y_m)$ (where $y_j$ is equal to $1$ if item $j$ is picked and otherwise, it is equal to $0$) that maximises the profit subject to the weight of the selected items not exceeding the capacity of the knapsack ($W$).  Formally, the goal is to maximise
\begin{align*}
& g (y) = \sum_{j=1}^{m} p_j y_j\\
&\text{subject to } \sum_{j=1}^{m} w_j y_j \leq W.
\end{align*}

%\subsection{Traveling Thief Problem}
The TTP is defined on the graph $G$ same as TSP and a set of items $I$ where items are scattered on the cities equally. Formally, every city $i$ except the first one contains a set of items $M_i$ (a subset of $I$). Same as KP, each item $k$ located in the city $i$ is associated with a profit $p_{ik}$ and a weight $w_{ik}$. 
To ease the presentation, we do not use the double subscripts for the profits and weights in the following but refer directly to the items at one particular city when required.

The thief should visit all the cities exactly once, pick some items into the knapsack, and return to the first city. A rent $R$ should be paid for the knapsack per time unit. The thief's speed non-linearly depends on the weight of the knapsack.
In TTP, we aim to find a solution $t=(x,y)$ consisting of a tour $x$ and a KP solution $y$ (called a packing list in the context of TTP) that maximises
%TODO
%DEFINE NOTATIONS FOR W_xi
\begin{align*}
& z(x,y) = g(y) - R \left( \frac{d_{x_n x_1}}{\nu_{max}-\nu W_{x_n}} + \sum_{i=1}^{n-1} \frac{d_{x_i x_{i+1}}}{\nu_{max}-\nu W_{x_i}} \right)\\
&\text{subject to } \sum_{j=1}^{m} w_j y_j \leq W.
\end{align*}
Here, $\nu_{\max}$ and $\nu_{\min}$ are the maximal and minimal traveling speed, $\nu = \frac{\nu_{max}-\nu_{min}}{W}$ is a constant, and $W_{x_i}$ is the cumulative weight of the items collected from the start of the tour up to city $x_i$.

In this study, $z$ serves as the fitness function, and $f$ and $g$ serve as the behavioural descriptor~(BD). Generally, the fitness function indicates how well a solution solves the given problem, while the BD shows how it solves the problem and behaves in terms of the features. In this case, the BD presents the length of the tour ($f$) and the value of items collected ($g$), whereby the fitness function returns the overall profit ($z$). Here, we aim to compute a diverse set of high-quality solutions differing in the BD. By this means, we can look into the distribution of high-performing TTP solutions over the 2D space of TSP and KP. %For this purpose, we employ the concept of MAP-Elites. 

\section{Bi-level Map-Elites-based Evolutionary Algorithm}
\label{Sec:map}

Map-Elites is an evolutionary computation approach where solutions compete with each other to survive. However, competition is only among solutions with a similar BD value in order to maintain diversity. We require a hyperparameter to define the similarity and the tolerance of acceptable differences between two descriptors. In the MAP-Elites algorithms, the BD space is discretised into a grid, where each cell is associated with one BD type. It means each solution belongs to at most one cell in the behavioural space (the map). Map-Elite algorithms typically keep only the best solution in each cell. When a solution is generated, it is assessed and potentially added to the cell with the associated BD. If the cell is empty, the solution occupies the cell; otherwise, the best solution is kept in the cell. The map aids in understanding and visualising the distribution of high-quality TTP solutions. For instance, how much we should move away from the optimal TSP tour and the optimal KP solution to generate high-performing TTP solutions? 
\begin{figure}[t]
    \centering
    \includegraphics[width=0.6\textwidth]{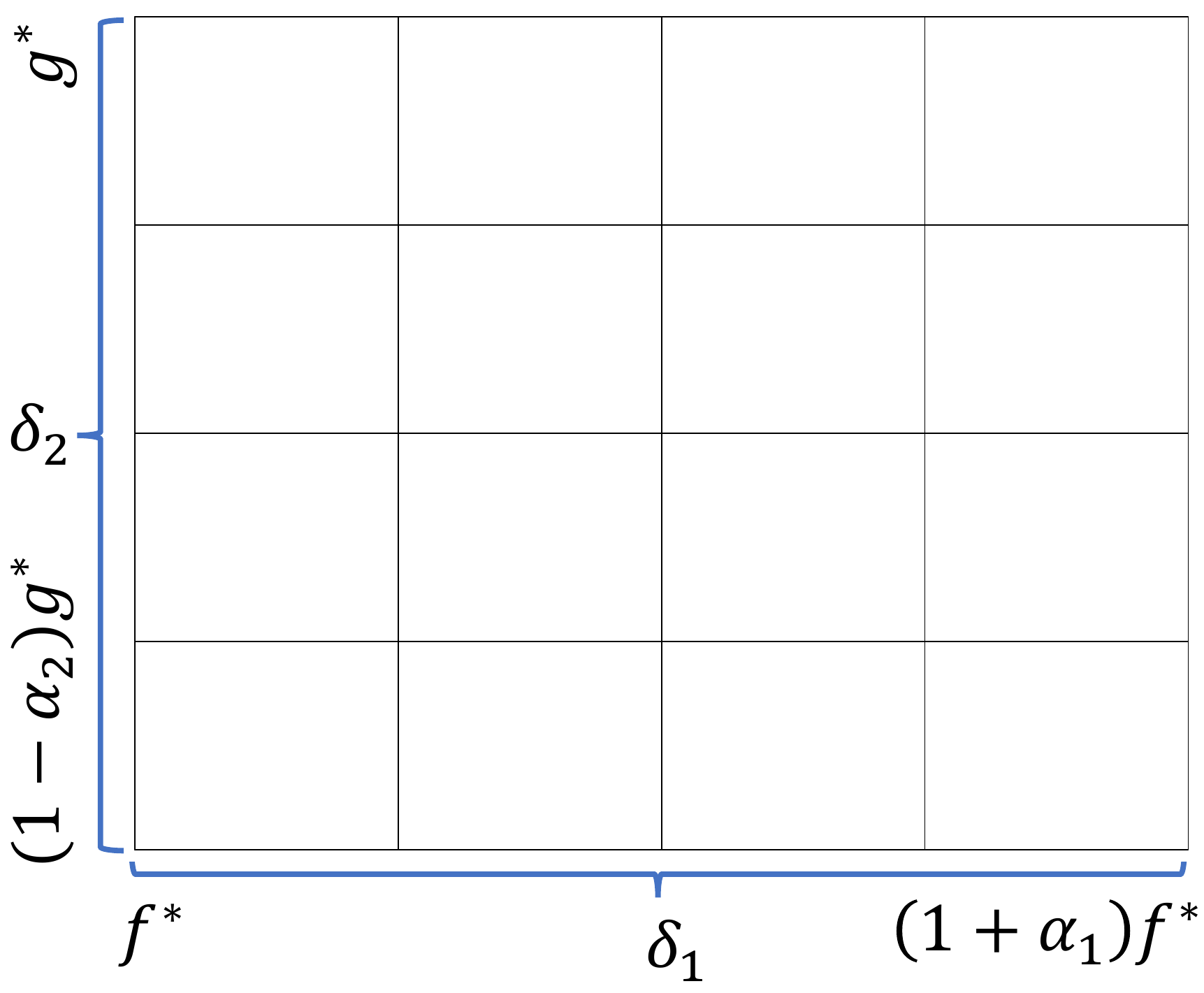}
    \caption{The representation of an empty map. There are $\delta_1 \times \delta_2$ cells within the map.} 
    % \frank{deltas just refer to width of 1 cell. Correct? Currently it looks like that it's taking the whole axis indicated by blue} \adel{They refer to the number of the cells in axis x and y. Is it confusing?} \frank{You would have to say that there are $\delta_1 \times delta_2$ cells}}
    \label{fig:example_map}
\end{figure}

Generally, the behavioural space can be extremely large. Thus, it is rational to limit the map to a promising part of the space; otherwise, either the number or the size of the cells increases severely, and as a result the performance and efficiency of the algorithms is undermined. As mentioned, a TTP solution consists of a tour and a packing list that belong to the TSP and the KP components of the problem. Although solving each sub-problems separately does not necessarily result in a high-quality TTP solution, a TTP solution should score fairly good in both features in order to gain high profits. Thus, we focus on solutions within $\alpha_1$ and $\alpha_2$ percents gap to the optimal TSP value ($f^*$) and the optimal KP value ($g^*$), respectively. In this study $\alpha_1$ and $\alpha_2$ are set to $5$ and $20$, respectively, based on initial experimental investigations. Figure~\ref{fig:example_map} depicts an empty map. There are $\delta_1 \times \delta_2$ cells. Cell $(i,j)$, $1 \leq i \leq \delta_1$, $1 \leq j \leq \delta_2$ contains the best found solution with TSP score in $\left[f^*+(i-1)\left(\frac{\alpha_1 f^*}{\delta_1}\right), f^*+(i)\left(\frac{\alpha_1 f^*}{\delta_1}\right)\right)$ and KP score in $\left[(1-\alpha_2)g^*+(j-1)\left(\frac{\alpha_2 g^*}{\delta_2}\right), (1-\alpha_2)g^*+(j)\left(\frac{\alpha_2 g^*}{\delta_2}\right)\right)$, the corresponding BD. 
% \frank{There are $\delta_1 \times \delta_2$ cells. Cell $(i,j)$, $1 \leq i \leq \delta_1$, $1 \leq j \leq \delta_2$ contains all solutions with TSP score in (give interval) and KP score in (give interval)}
The cell (1, $\delta_2$) consists of TTP solutions with TSP and KP values closest to the optimums. In this study, we require to know $f^*$ and $g^*$. For this purpose, we can use EAX in \cite{nagata2013powerful} and dynamic programming (DP) in \cite{DBLP:Toth80} to compute $f^*$ and $g^*$ for the TTP instances, respectively.

\begin{algorithm}[t!]
\begin{algorithmic}[1]
% REQUIRE{Initial Population $P$, and a limit for consecutive failures in improvement of the packing list $M$.}
\STATE Find the optimal/near-optimal values of the TSP and the KP by algorithms in \cite{nagata2013powerful,DBLP:Toth80}, respectively. 
\STATE Generate an empty map and populate it with the initialising procedure. 

\WHILE{termination criterion is not met}
\STATE Generate an offspring and calculate the TSP and the KP scores.
\IF{The TSP and the KP scores are within $\alpha_1\%$, and $\alpha_2\%$ gaps to the optimal values of BD.}
\STATE Find the corresponding cell to the TSP and the KP scores.
\IF{The cell is empty}
\STATE Store the offspring in the cell.
\ELSE
\STATE Compare the offspring and the individual occupying the cell and store the best individual in terms of TTP score in the cell.
\ENDIF
\ENDIF
\ENDWHILE
\end{algorithmic}
\caption{The MAP-Elites-Based Evolutionary Algorithm}
\label{alg:map}
\end{algorithm}

Algorithm \ref{alg:map} describes the BMBEA. The initialising procedure and the operators to generate a new TTP solution will be discussed later. Having generated an empty map, we populate it with an initialising procedure. After generating offspring, we calculate the TSP score and the KP score of the offspring. If the TSP and the KP scores are within $\alpha_1\%$ and $\alpha_2\%$ gap of the optimal values, respectively, we find the cell corresponding to those scores; otherwise, the offspring is discarded. If the corresponding cell is empty, the offspring is kept in the cell; otherwise, we compare the offspring and the individual in the cell and keep the individual with the highest TTP score. We repeat steps 3 to 10 until a termination criterion is met.           

Evolutionary algorithms require some operators to generate new solutions (offspring) from old ones (parents);  BMBEA is no exception. One can see the generating of TTP solutions as a bi-level process. First, new tours can be generated by mutation or crossovers; then, we can compute a suitable packing list for the new tours to have complete TTP solutions. 
\subsection{Search Operators for TSP}
\label{subSec:TSP_op}
\begin{figure}[t]
    \centering
    \includegraphics[width=\columnwidth]{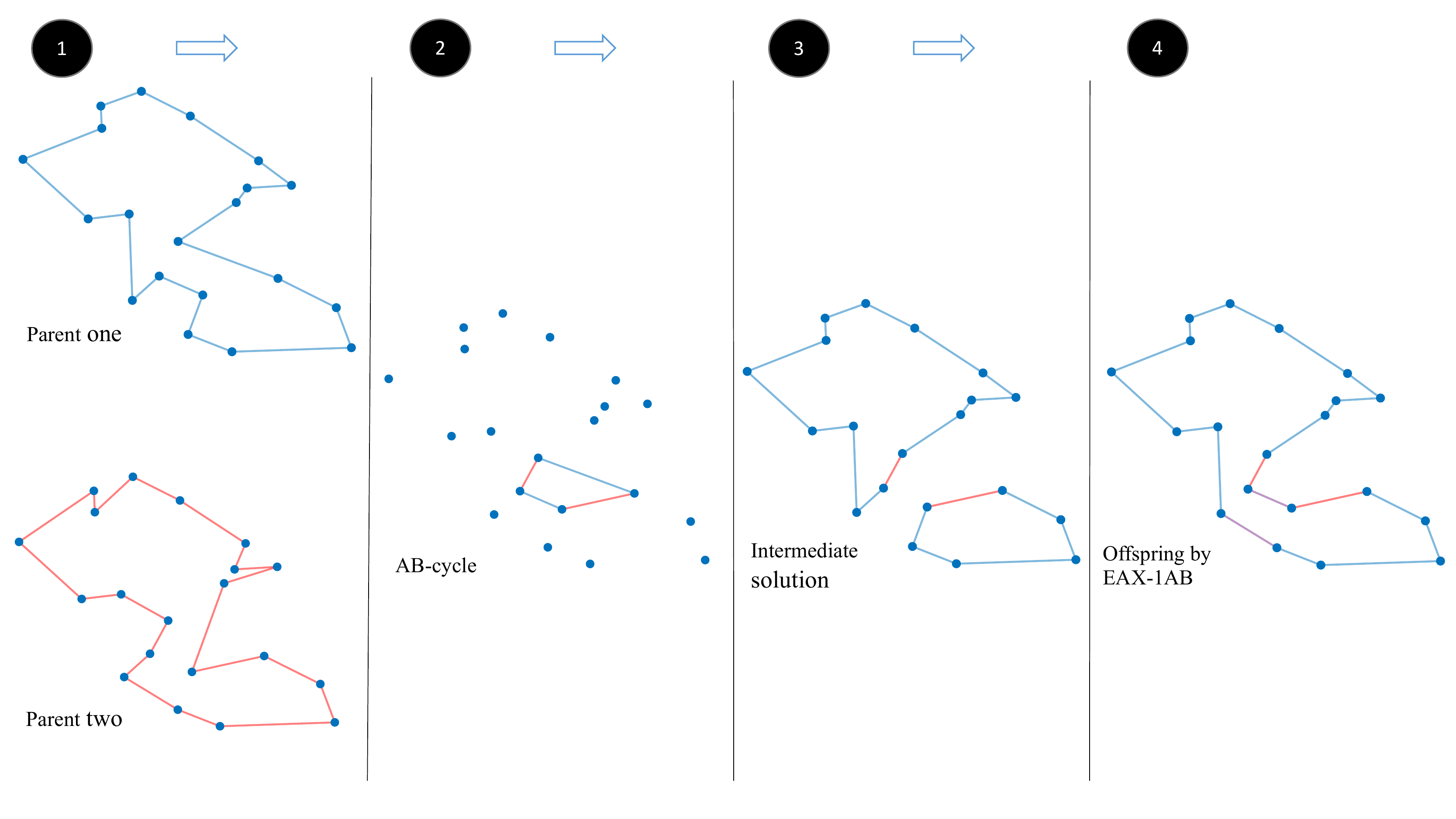}
    \caption{The representation of the steps to implement EAX.}
    \label{fig:example_eax_1ab}
\end{figure}

We consider EAX crossover~(\cite{nagata2013powerful}) to generate new TSP tours. EAX is a highly performing TSP crossover known as one of the state-the-of-the-art operators in solving TSP. The use of EAX has also been shown to lead to high-quality solutions for the TTP in \cite{WuijtsT19}. EAX has several variants; we incorporate the EAX-1AB due to its simplicity and efficiency. The EAX consists of three steps. Figure~\ref{fig:example_eax_1ab} depicts the three steps to implement the EAX-1Ab.
\begin{itemize}
    % \item Selection: Selecting two parents uniformly at random (Fig \ref{fig:example_eax_1ab}.1).
    \item AB-cycle: Generating one AB-cycle from the two parents by alternatively choosing edges from the first and second parents until a cycle is formed (Fig \ref{fig:example_eax_1ab}.2). 
    \item Intermediate Solution: Copying all edges of the first parent to the offspring; then removing the Ab-cycle's edges that belong to the first parent from the offspring, and adding the other edges of the AB-cycle to it (Fig \ref{fig:example_eax_1ab}.3).
    \item The Complete Tour: Connecting all sub-tours of the intermediate solution to form a complete tour (Fig \ref{fig:example_eax_1ab}.4).
\end{itemize}
To connect two sub-tours, we require to discard one edge from each sub-tour, and add two new edges to connect each end of the deleted edges. For Step 3,  we start with the sub-tour ($r$) with the minimum edge number. Then, we select the $4$-tuples of edges such that $\{e_1, e_2, e_3, e_4\} = \arg \min \{-d(e_1)-d(e_2)+d(e_3)+d(e_4)\}$ where $e_1 \in E(r)$ and  $e_2 \in E(t) \setminus E(r)$. $E(t)$ and $E(r)$ represent the set of edges formed the intermediate solution $t$ and sub-tour $r$, respectively. We refer the interested readers to \cite{nagata2013powerful} for more details about the process of generating a new tour by the EAX.
Alternatively, 2-OPT can be used to generate the TSP tours. 2-OPT is a random neighbourhood search, where two elements of a permutation are selected uniformly at random. Having these elements swapped, we reorder the elements in between in a backward direction.  
\subsection{Search Operators for KP}
\label{subSec:kp_op}
In the second phase, we optimise the packing list to match the TSP tour and form a good TTP solution. To this mean, inner algorithms are required to optimise the packing list. When the tour is fixed, and the packing list is optimised, the problem is referred to as Packing While Traveling~(PWT) \cite{PolyakovskiyB0MN14} in the literature. \citet{NeumannPSSW18} introduced a DP algorithm to solve the PWT problem to optimality.
\subsubsection{Dynamic Programming}
DP is a classical approach in solving the KP. Here, we employ the DP introduced in \cite{NeumannPSSW18} to solve the PWT problem. The DP includes a table $\beta$ consisting of $W$ rows and $m$ columns. In the DP, items are processed in the order that their corresponding node appears in the tour. For example, $I_i$ is processed sooner than $I_j$, if the node to which $I_i$ belongs is visited sooner than the node of $I_j$. If two items belong to the same node, they are processed according to their indices. The entry $\beta_{i,j}$ represents the maximal profit that the thief can obtain among all combinations of items $I_k$ with $I_k \preceq I_i$ bringing about the weight exactly equal to $j$. If no combinations lead to the weight $j$, $\beta_{i,j}$ is set to $-\infty$.

Let denote the profit of the empty set by $B(\emptyset)$, that is equal to traveling cost with an empty knapsack. Moreover, we denote the profit by $B(I_i)$ when only item $I_i$ is collected. Thus, for the first item $I_i$ (the first row of the table $\beta$) based on the order aforementioned, we have:
$$\beta_{i,0} = B(\emptyset),\quad\beta_{i,w_i} = B(I_i),\quad\beta_{i,j} = -\infty, \forall j\notin \{0,w_i\}$$

let show the predecessor of $I_i$ by $I_k$. For the rest of the table, each entry $\beta_{i,j}$ can be computed from $\max(\beta_{k,j}, T)$, where 
$$T = \beta_{k,j-w_i}+p_i-R\sum_{l=1}^n d_l\left(\frac{1}{\nu_{max}-\nu j}-\frac{1}{\nu_{max}-\nu j-w_i}\right)$$ 

The $\max_j \beta_{m,j}$ is reported as the optimal profit that the thief can gain from the given tour. Although DP can provide us with the optimal packing list for a given tour, the run-time is quite long. Considering that we compute the packing list in the second level of a bi-level optimisation, it can affect the time efficiency of the BMBEA. Therefore, we propose an EA as an alternative. The interested readers are referred to \cite{NeumannPSSW18}, which analysed the run-time of the DP.   
\subsubsection{$(1+1)$~Evolutionary Algorithm} The $(1+1)~EA$ is a well-known simple EA that converges fact since it only keeps the best-found solution. First, the new tour generated by the TSP operators inherits its parent's packing list. Next, a new packing list is generated by mutation. If the new packing list results in a higher TTP score, the new packing list is replaced with the old one. We continue these steps until a termination criterion is met. For mutation, the bit-flip is used, where each bit is independently flipped by mutation rate $\frac{1}{m}$. 

The mutation can result in packing lists violating the knapsack's capacity. We incorporate a repair function into the $(1+1)~EA$ to avoid the violation. After the offspring is mutated, the repair function fixes the offspring's violation. The repair function  removes collected items uniformly at random one by one until the packing list complies with the capacity constraint.

\subsection{Initialisation}
\label{subSec:Initial}

One may notice that it is doubtful to populate the map with random solutions. This is because, the map only accepts TTP individuals with fairly good TSP and KP scores. Therefore, a heuristic approach is required to populate the map initially. We can use the EAX-based algorithm in \cite{nagata2013powerful} to find the optimal/near optimal TSP tours in terms of length. Having extracted the tours, we can compute a good quality packing list for each tour by one of the KP operators mentioned in section ~\ref{subSec:kp_op}. This results in TTP solutions with high TSP and KP scores, let denote the set of solutions by $P_0$. This initial population enables us to populate the map at the beginning of the BMBEA. In this study, we use a target length ($f^*$) as a termination criterion for the EAX-based algorithm in \cite{nagata2013powerful}, so as to increase the time-efficiency and diversity of tours. If we do not have the target values it is important to tune the running time of the algorithm for each instance individually. Note that in case of using $(1+1)$EA as the KP operator, it will boost the performance of BMBEA, if we start with the optimal packing plan obtained by \cite{DBLP:Toth80}. 

\subsection{A More Relaxed Map}
\label{subsec:relaxed}
Treating $\alpha_1$ and $\alpha_2$ as input required intensive tuning and preliminary experimental investigations. Randomly selecting values for $\alpha_1$ and $\alpha_2$ not only affects the algorithm's efficiency, but it may also bring about an infeasible map that is impossible to fill. As a result, the algorithm cannot produce any solution in such cases. For this purpose, we propose another method in which we set the thresholds after the initialization. Having computed $P_0$ as mentioned above, we set the $(1+\alpha_1)f^* $ to $\argmax_{x \in P_0} f(x)$ and $(1-\alpha_2)g^*$ to $\argmin_{y \in P_0} g(y)$.
\subsection{($\mu+1$) EA}
\label{subsec:(mu+1)}

We require a similar algorithm with a conventional survival selection to investigate the impact of QD and MAP-Elitism on the results. We consider $(\mu+1)$EA since it has the same offspring size as the introduced algorithm. Here, we generate an initial population $P_0$, same as Section \ref{subSec:Initial}. The parents are selected uniformly at random; then, an offspring is generated as described in Section \ref{subSec:TSP_op} and \ref{subSec:kp_op}. After adding the offspring to the population, we remove one individual with the worst TTP score. Algorithm \ref{alg:mu} outlines steps required for $(\mu+1)$EA.  

\begin{algorithm}
\begin{algorithmic}[1]
% REQUIRE{Initial Population $P$, and a limit for consecutive failures in improvement of the packing list $M$.}
 
\STATE Generate an initial population as explained in Section.  

\WHILE{termination criterion is not met}
\STATE Generate an offspring and add it to the population $P$.

\STATE Discard one individual $p$ from $P$, where $\argmin_{p\in P} z(p)$.

\ENDWHILE
\end{algorithmic}
\caption{$(\mu+1)$ Evolutionary Algorithm}
\label{alg:mu}
\end{algorithm}

\begin{table}[t]
% \begin{scriptsize}
\renewcommand{\tabcolsep}{2pt}
\renewcommand{\arraystretch}{0.6}
    \centering
    \begin{tabular}{c l | c l}
    \toprule
    No. & Original Name & No. & Original Name\\
    \midrule
         1 & eil51\_n50\_bounded-strongly-corr\_01 & 18 & a280\_n279\_uncorr\_01\\
         2 & eil51\_n150\_bounded-strongly-corr\_01 & 19 & rat575\_n574\_bounded-strongly-corr\_01\\
         3 & eil51\_n250\_bounded-strongly-corr\_01 & 20 & rat575\_n574\_uncorr-similar-weights\_01\\
         4 & eil51\_n50\_uncorr-similar-weights\_01 & 21 & rat575\_n574\_uncorr\_01\\
         5 & eil51\_n150\_uncorr-similar-weights\_01 & 22 & dsj1000\_n999\_bounded-strongly-corr\_02\\
         6 & eil51\_n250\_uncorr-similar-weights\_01 & 23 & dsj1000\_n999\_uncorr-similar-weights\_06\\
         7 & eil51\_n50\_uncorr\_01 & 24 & dsj1000\_n999\_uncorr\_04\\
         8 & eil51\_n150\_uncorr\_01 & 25 & u2152\_n2151\_bounded-strongly-corr\_01\\
         9 & eil51\_n250\_uncorr\_01 & 26 & u2152\_n2151\_uncorr-similar-weights\_01\\
         10 & pr152\_n151\_bounded-strongly-corr\_01 & 27 & u2152\_n2151\_uncorr\_01\\
         11 & pr152\_n453\_bounded-strongly-corr\_01 & 28 & fnl4461\_n4460\_bounded-strongly-corr\_01\\
         12 & pr152\_n151\_uncorr-similar-weights\_01 & 29 & fnl4461\_n4460\_uncorr-similar-weights\_01\\
         13 & pr152\_n453\_uncorr-similar-weights\_01 & 30 & fnl4461\_n4460\_uncorr\_01\\
         14 & pr152\_n151\_uncorr\_01 & 31 & dsj1000\_n999\_uncorr\_02\\
         15 & pr152\_n453\_uncorr\_01 & 32 & dsj1000\_n999\_uncorr\_03\\
         16 & a280\_n279\_bounded-strongly-corr\_01 & 33 & dsj1000\_n999\_uncorr-similar-weights\_03\\
         17 & a280\_n279\_uncorr-similar-weights\_01 & 34 & dsj1000\_n999\_uncorr-similar-weights\_04\\
         \bottomrule
    \end{tabular}
    % \end{scriptsize}
    \caption{The names of the TTP instances are used in the paper.}
    \label{tab:names}
\end{table}

\section{Experimental Investigation}
\label{Sec:EXP}
In this section, we use the BMBEA to compute a set of solutions for several TTP instances; then, we plot the map to illuminate the distribution of the solutions over the space of $f$ and $g$. Moreover, we comprehensively compare different search operators and their effects on the distributions and the final maps. We consider the EAX and the 2-OPT for generating tours and the DP and the $(1+1)$EA for computing the packing lists. Employing the operators alternatively, we have four different operator settings. The algorithms are terminated when they reach either of $10000$ iterations or 72 hours CPU time. Here, iteration is referred to as the main loop of the BMBEA. We use the TTP instances developed in \cite{PolyakovskiyB0MN14}. Table \ref{tab:names} presents the names of the instances used in the paper. Please note that we select the first instance of each sub-group except for the dsj1000. The renting price ($R$) is set to zero in those instances; the issue makes the TTP instances turn to KPs. We separate the instances into two categories, small and medium. Since DP's time efficiency is correlated with the number of items, we select instances where $m \leq 500$ for small instances.  
\begin{figure*}
\centering
\begin{tikzpicture}

%\node (tit) at (-4.4,0) {\scriptsize{Frequency:}};
\node (EAX-DP) at (-6,-0.3) {\scriptsize{\textcolor{gray!90}{EAX-DP}}};
\node (EAX-EA) at (-2,-0.3) {\scriptsize{\textcolor{gray!90}{EAX-EA}}};
\node (2-OPT-DP) at (2,-0.3) {\scriptsize{\textcolor{gray!90}{2-OPT-DP}}};
\node (2-OPT-EA) at (6,-0.3) {\scriptsize{\textcolor{gray!90}{2-OPT-EA}}};
\end{tikzpicture}
% \scalebox{0.9}{
\includegraphics[width=.23\columnwidth]{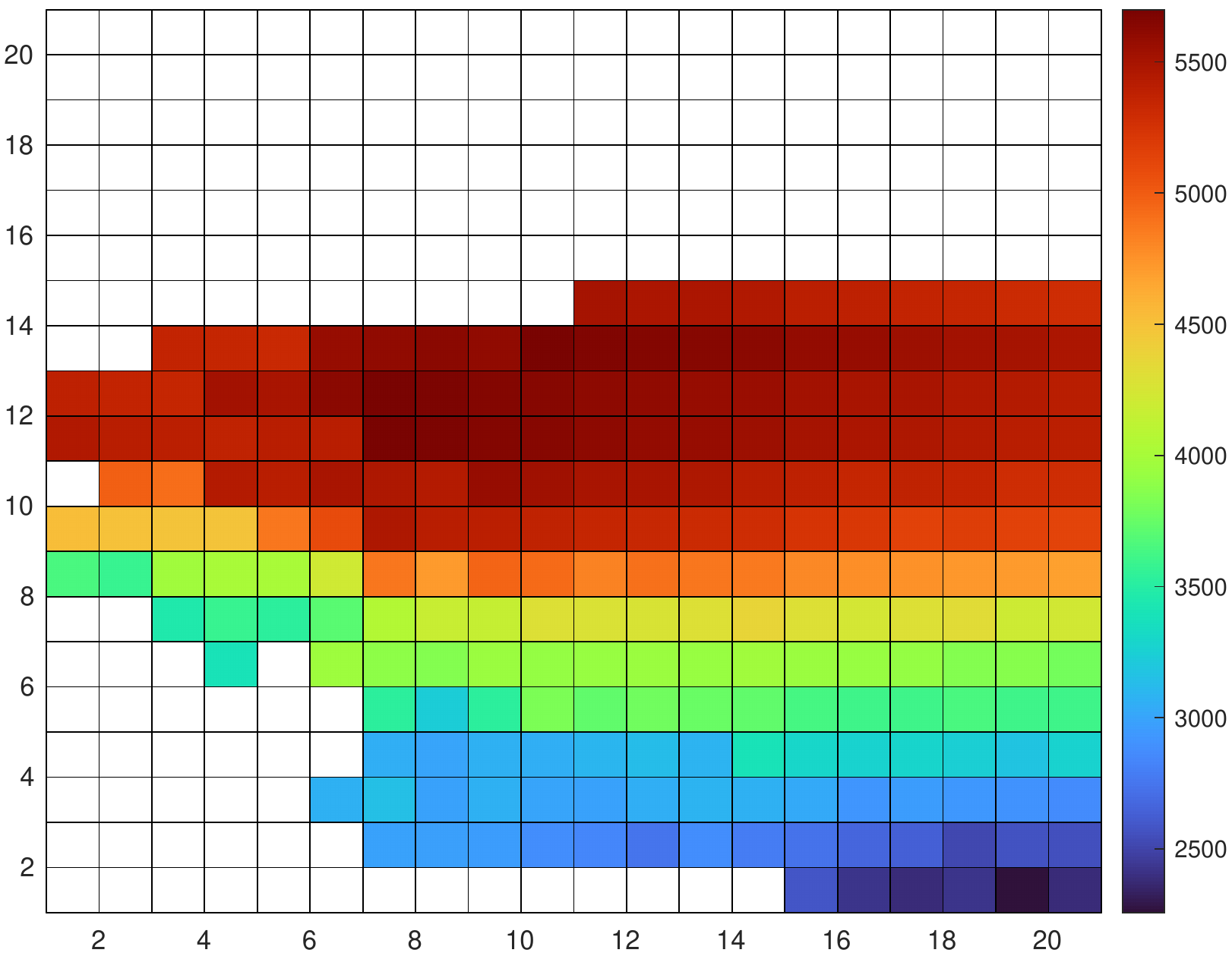}
\hskip5pt
\includegraphics[width=.23\columnwidth]{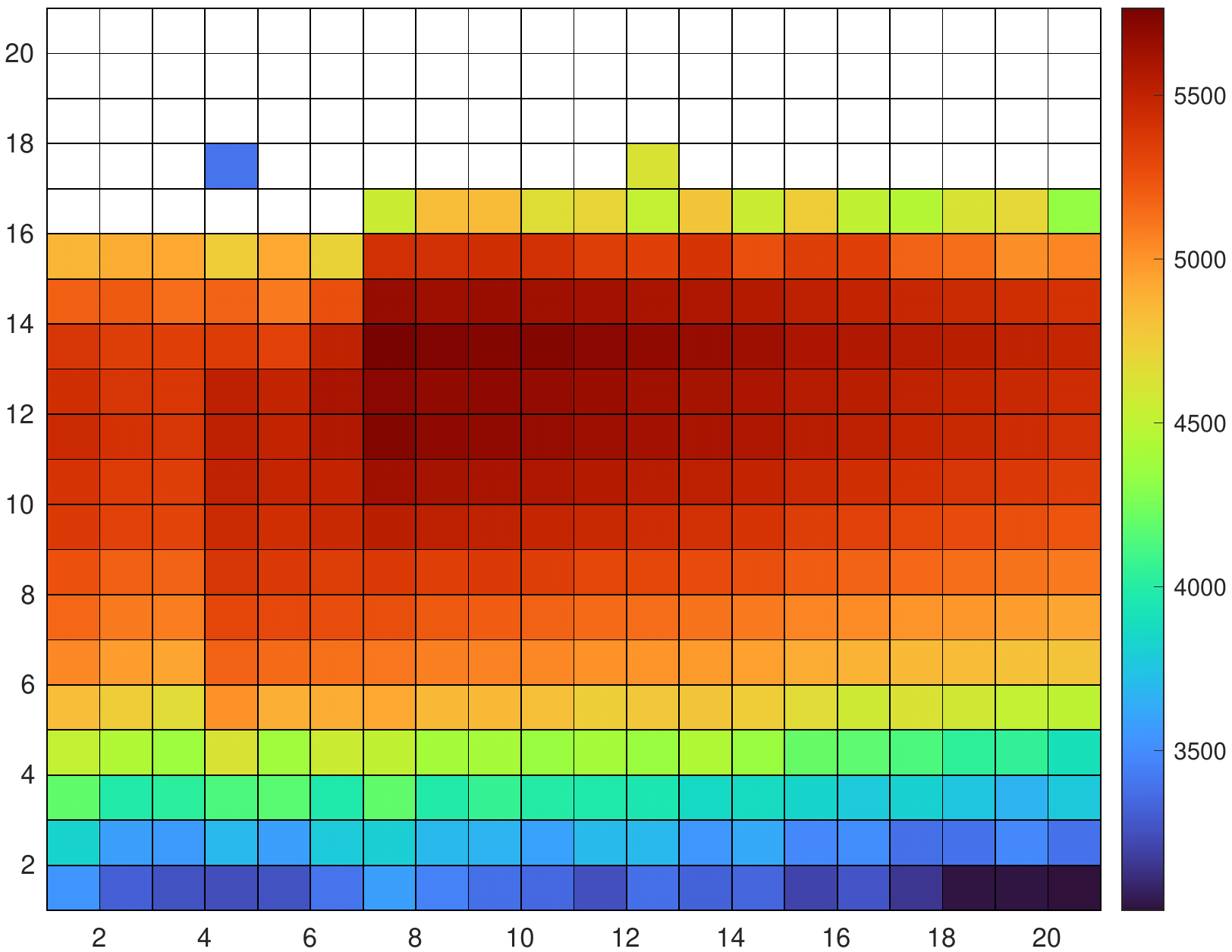}
\hskip5pt
\includegraphics[width=.23\columnwidth]{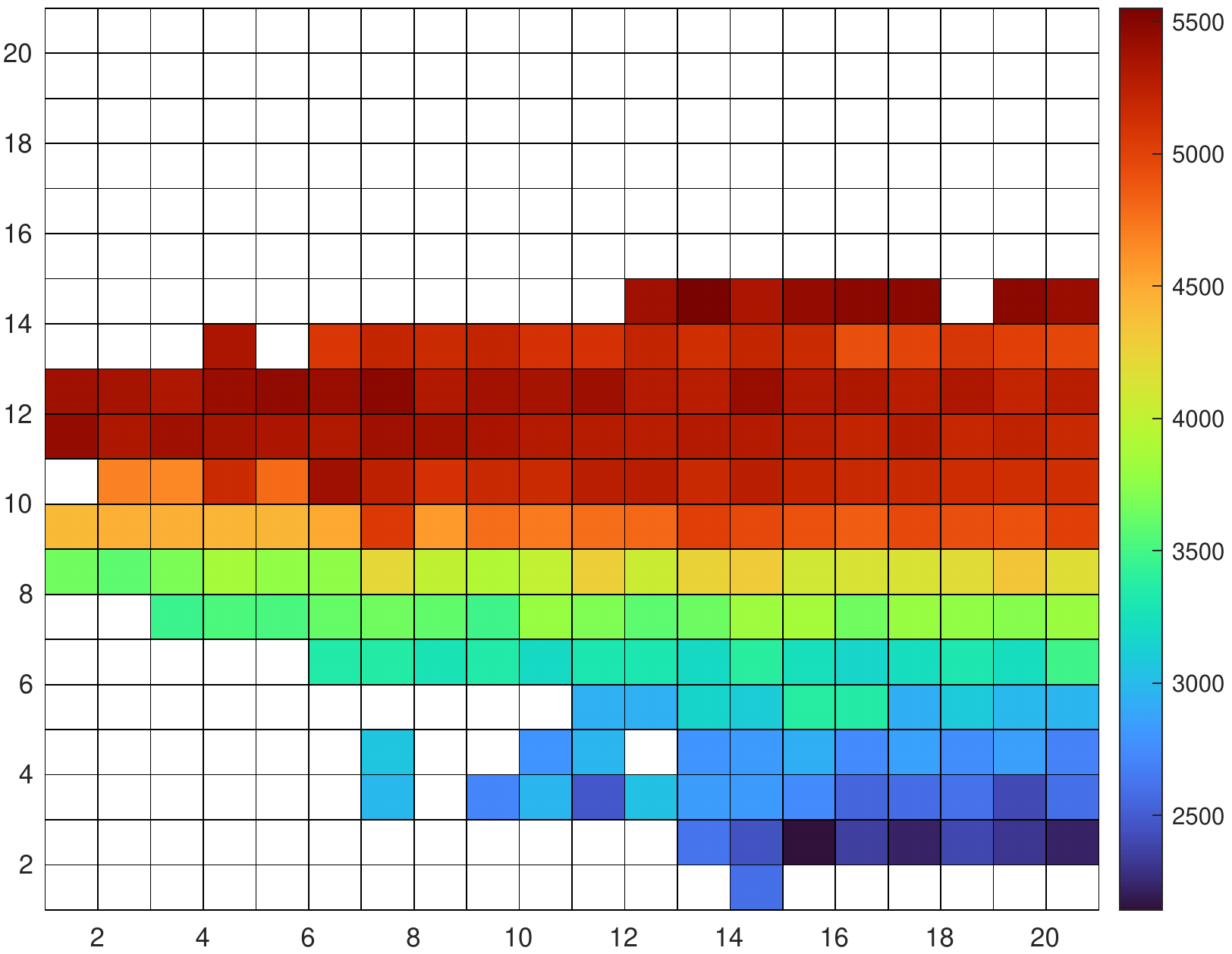}
\hskip5pt
\includegraphics[width=.23\columnwidth]{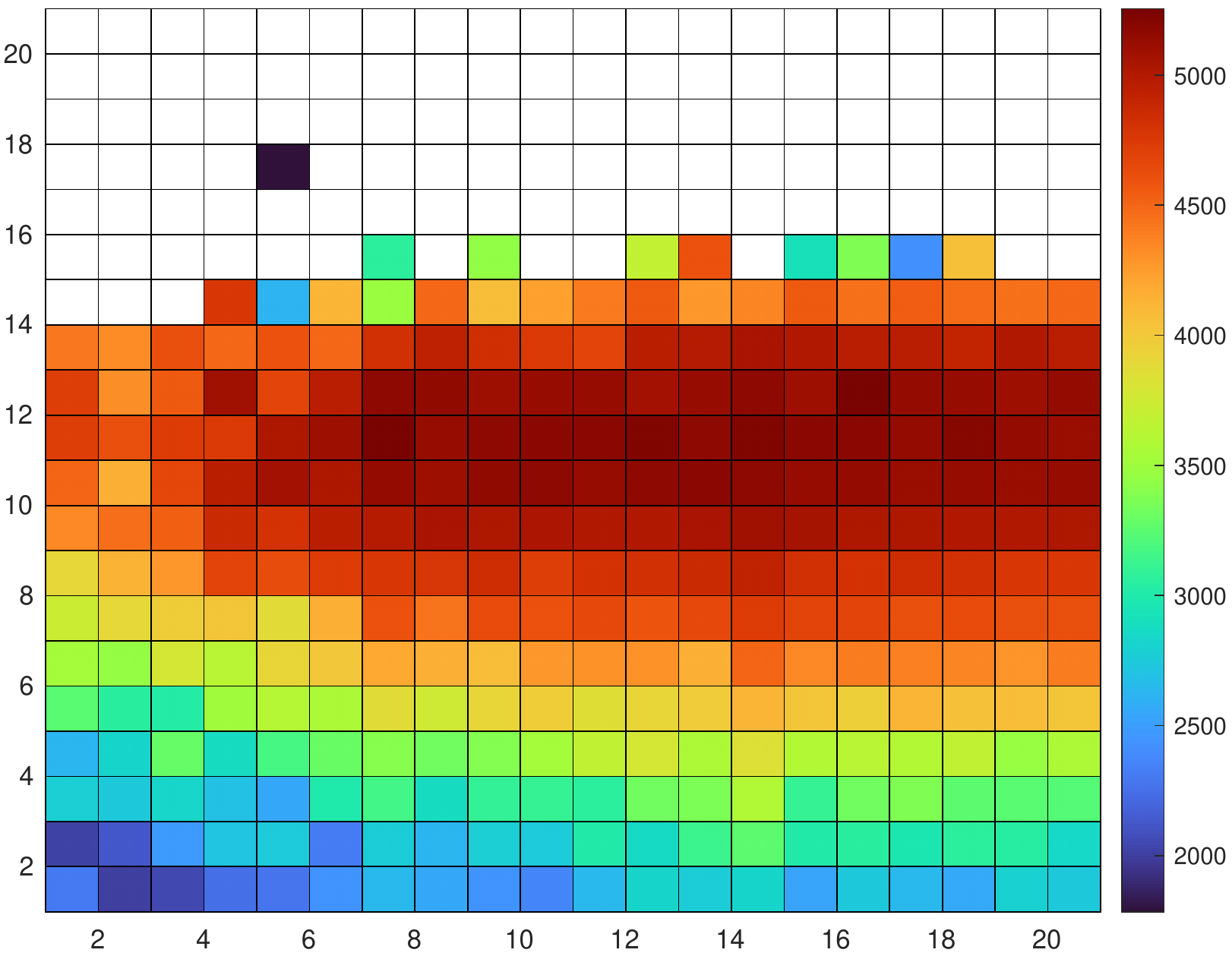}
\hskip5pt
\includegraphics[width=.23\columnwidth]{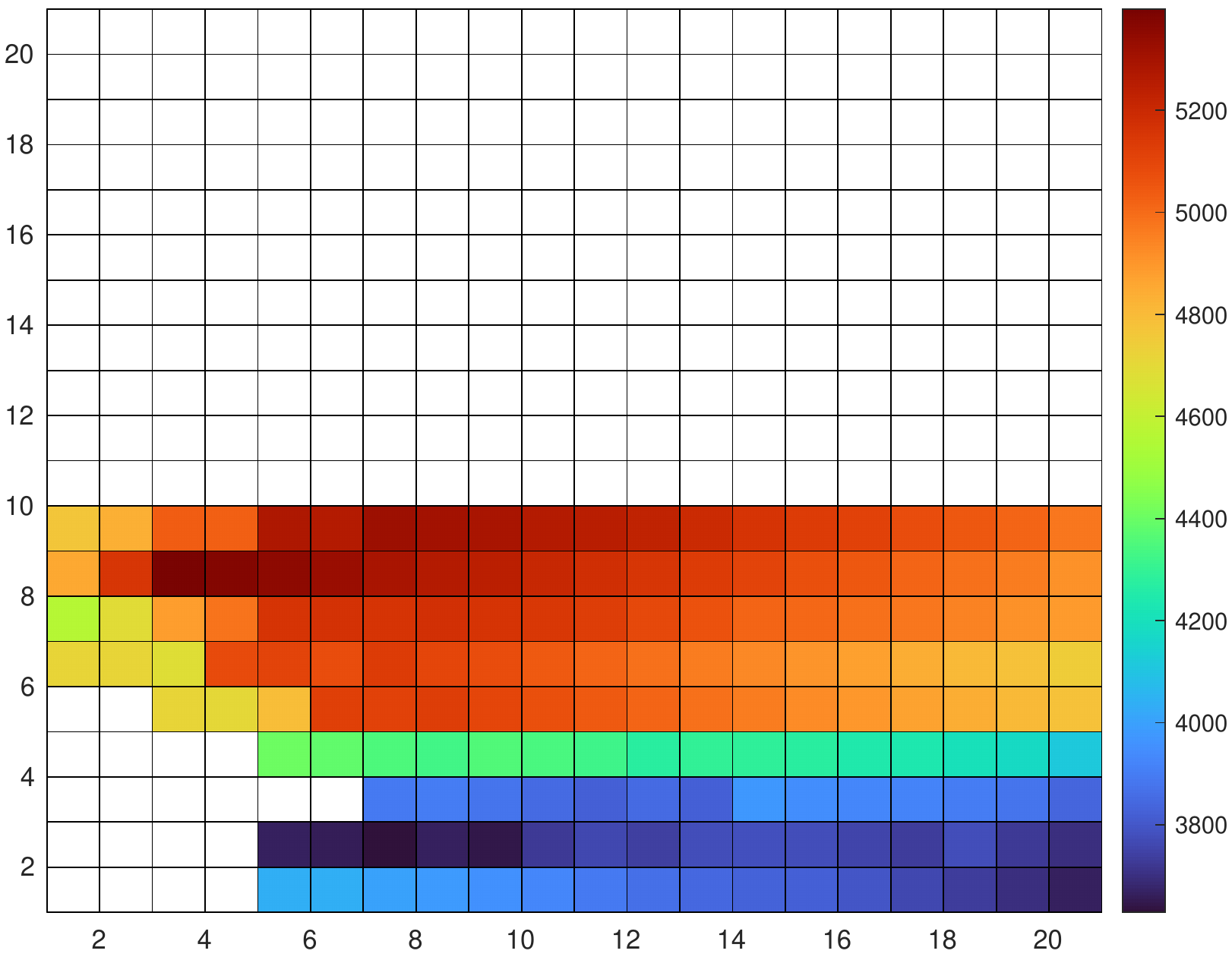}
\hskip5pt
\includegraphics[width=.23\columnwidth]{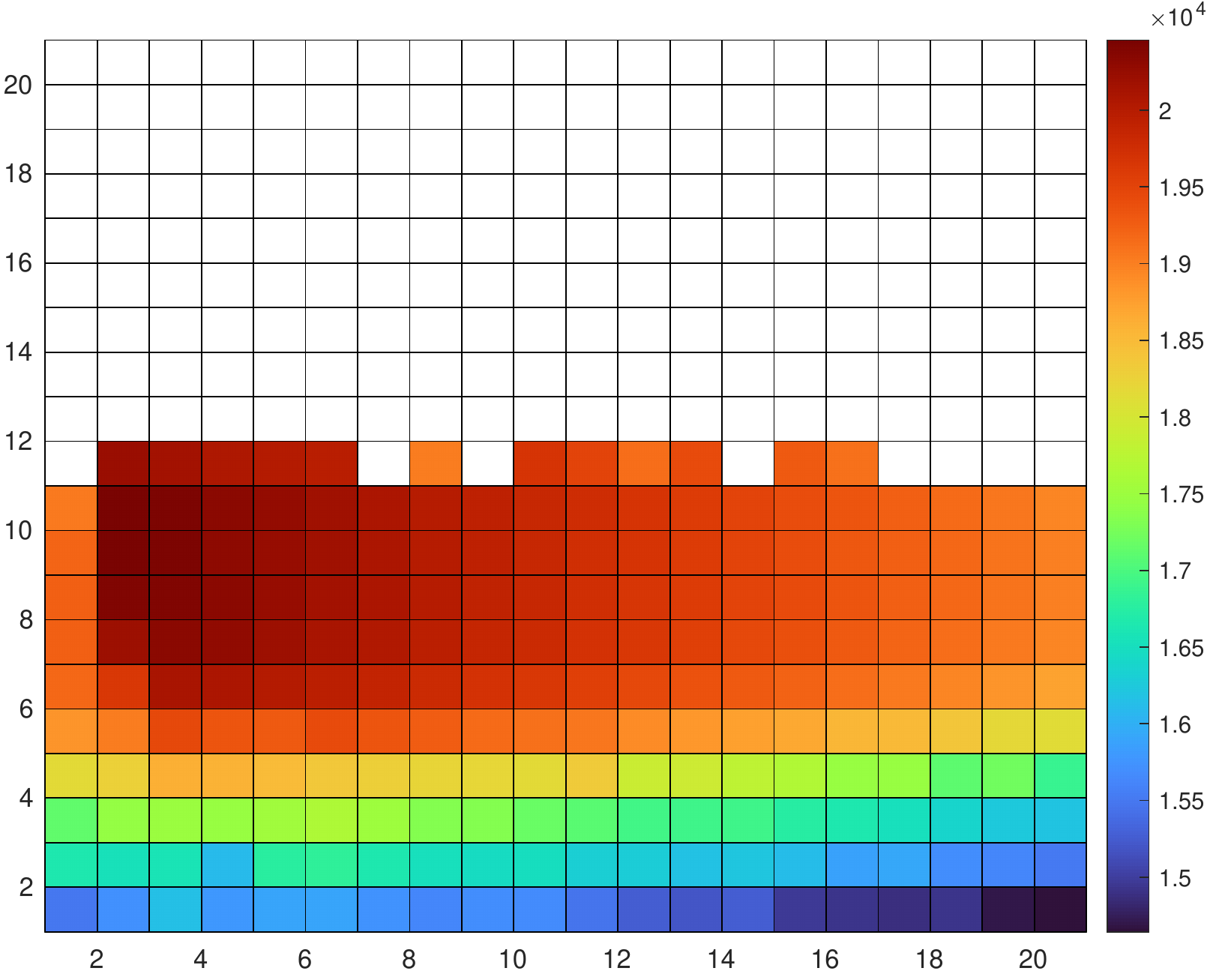}
\hskip5pt
\includegraphics[width=.23\columnwidth]{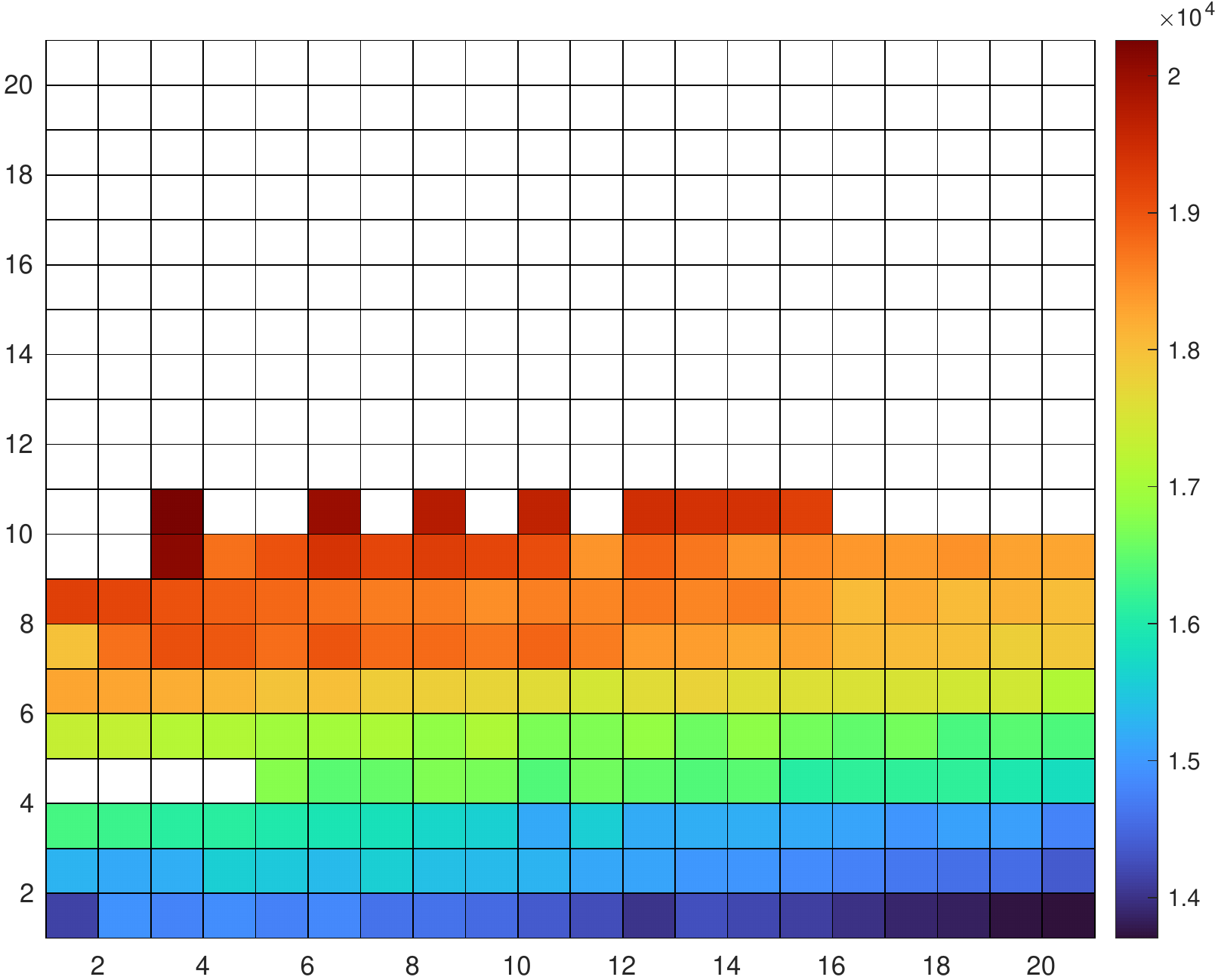}
\hskip5pt
\includegraphics[width=.23\columnwidth]{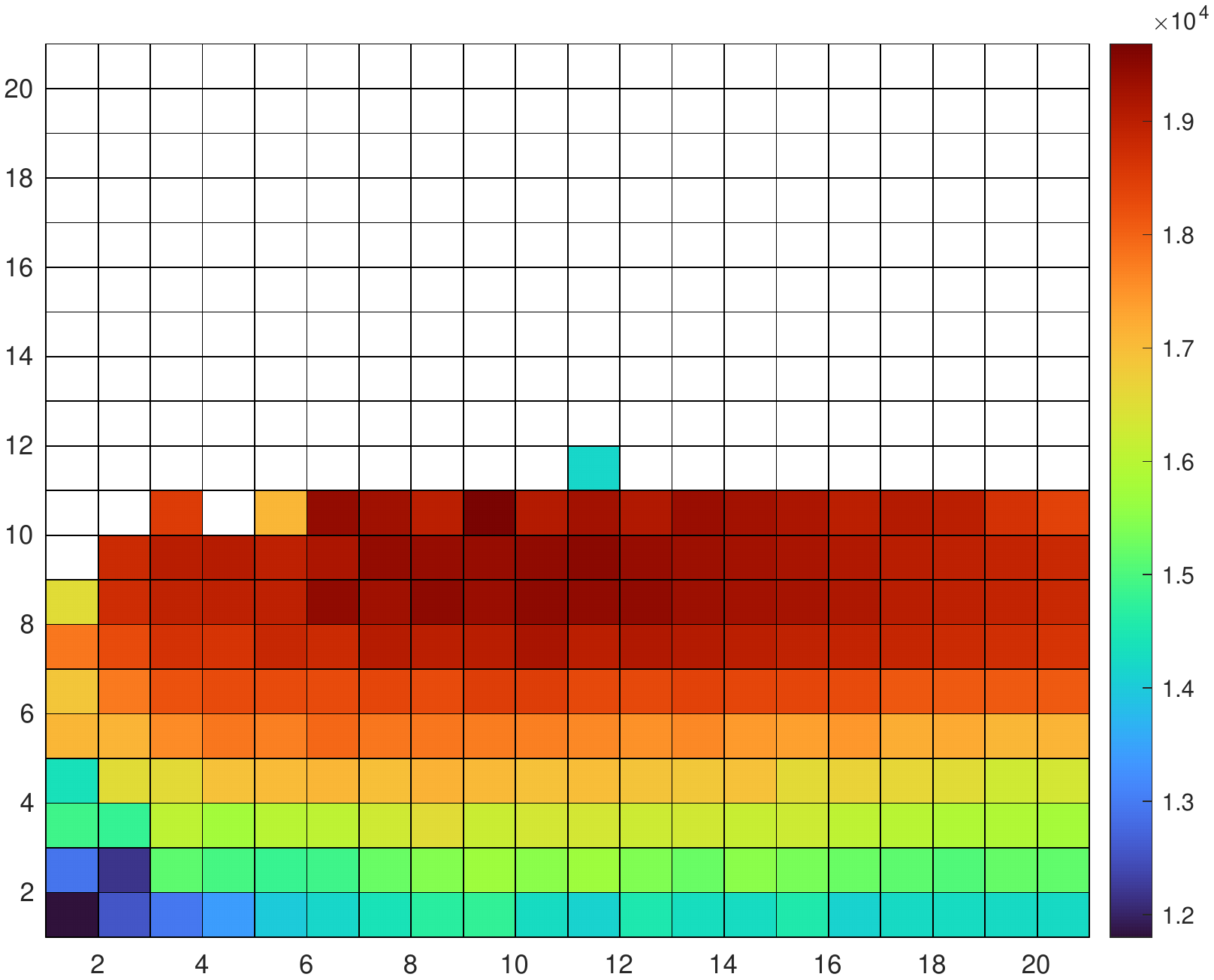}
\hskip5pt
\includegraphics[width=.23\columnwidth]{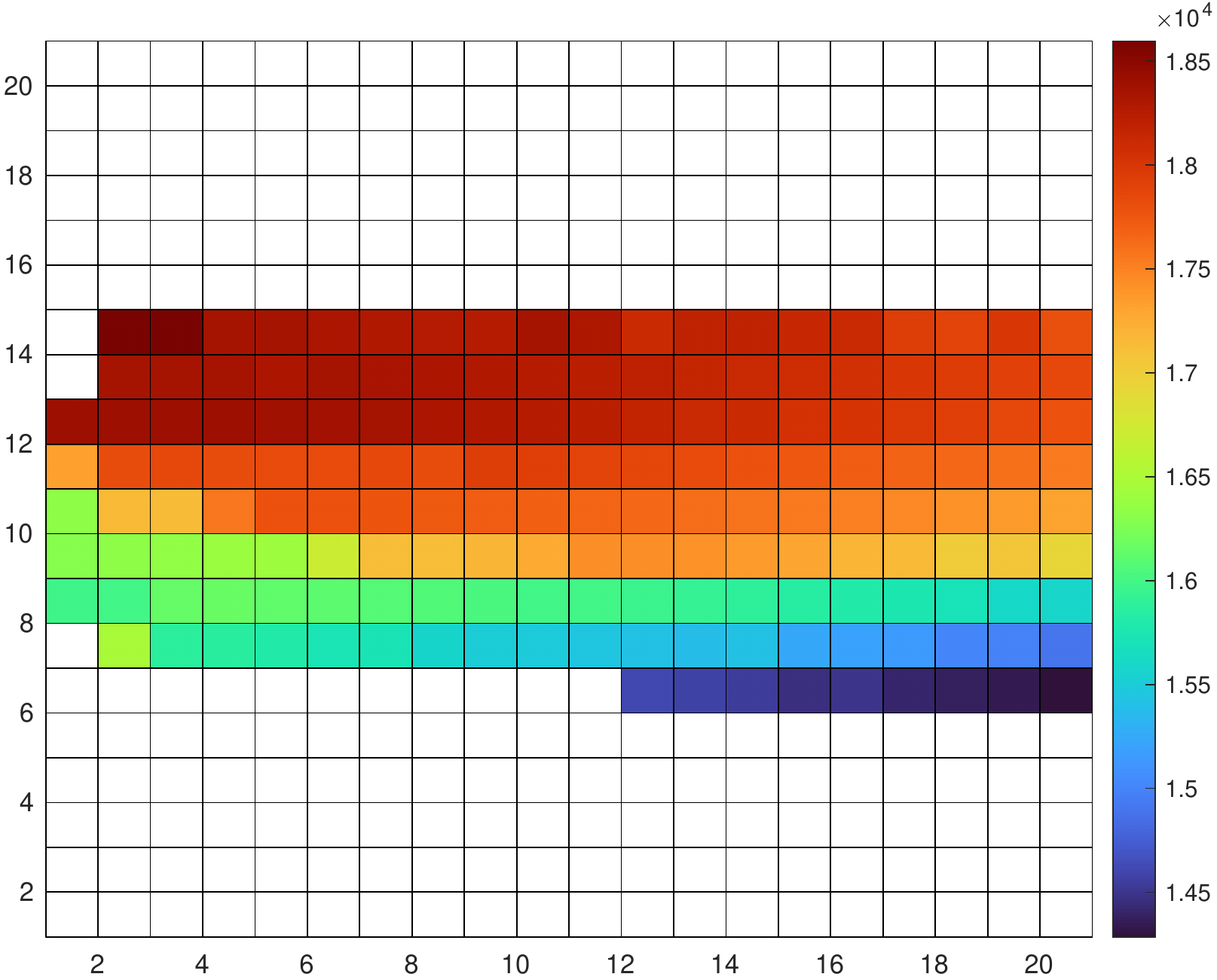}
\hskip5pt
\includegraphics[width=.23\columnwidth]{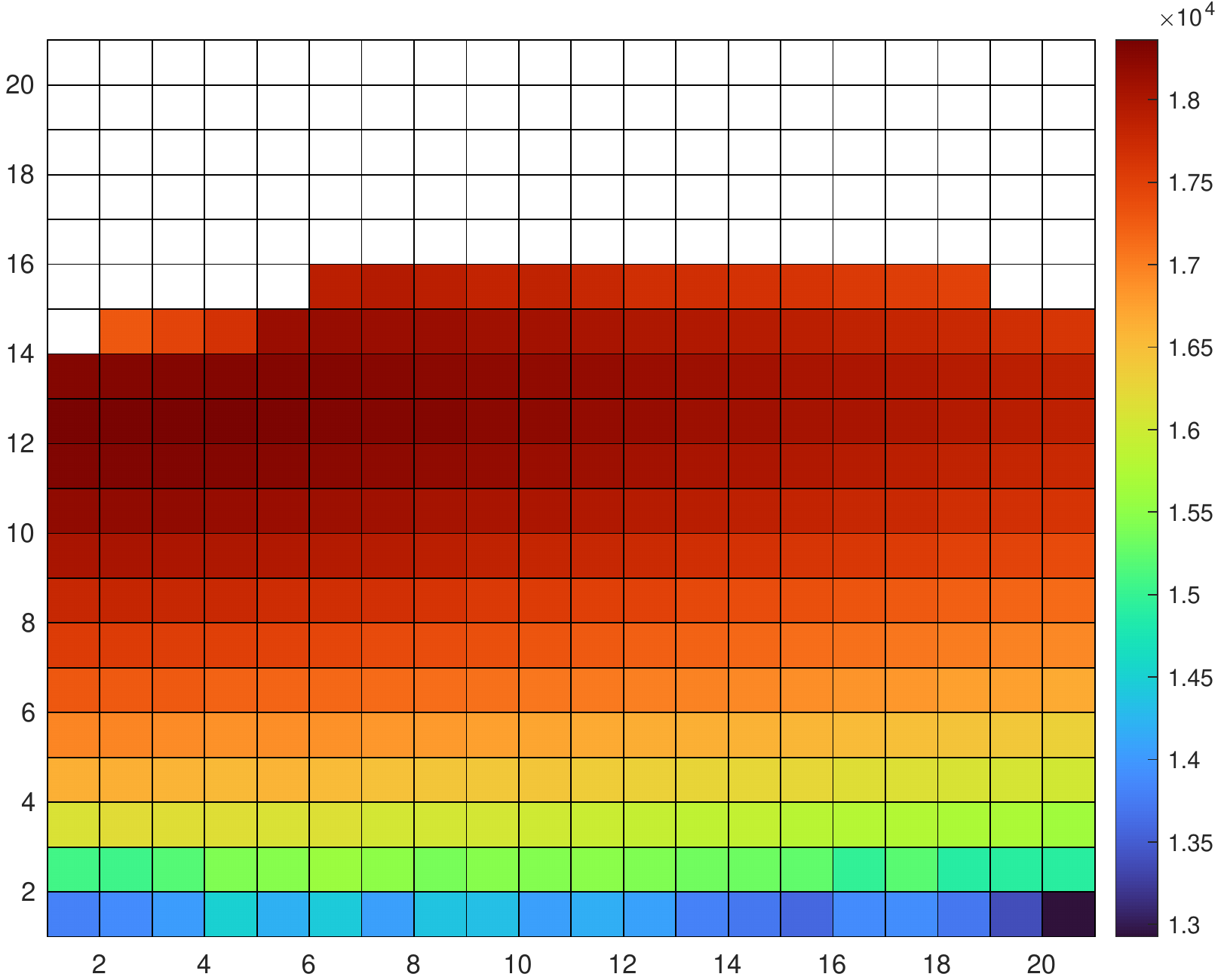}
\hskip5pt
\includegraphics[width=.23\columnwidth]{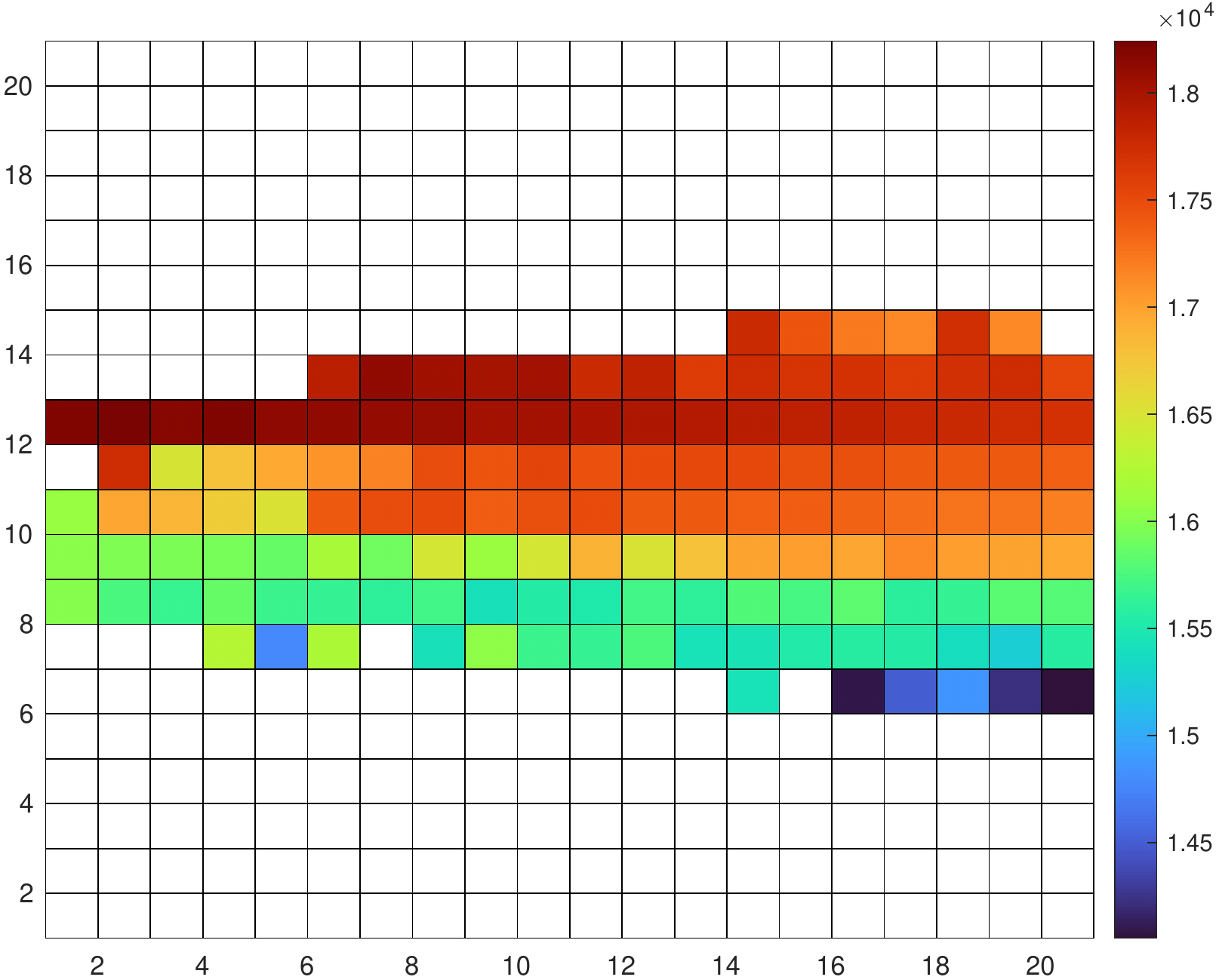}
\hskip5pt
\includegraphics[width=.23\columnwidth]{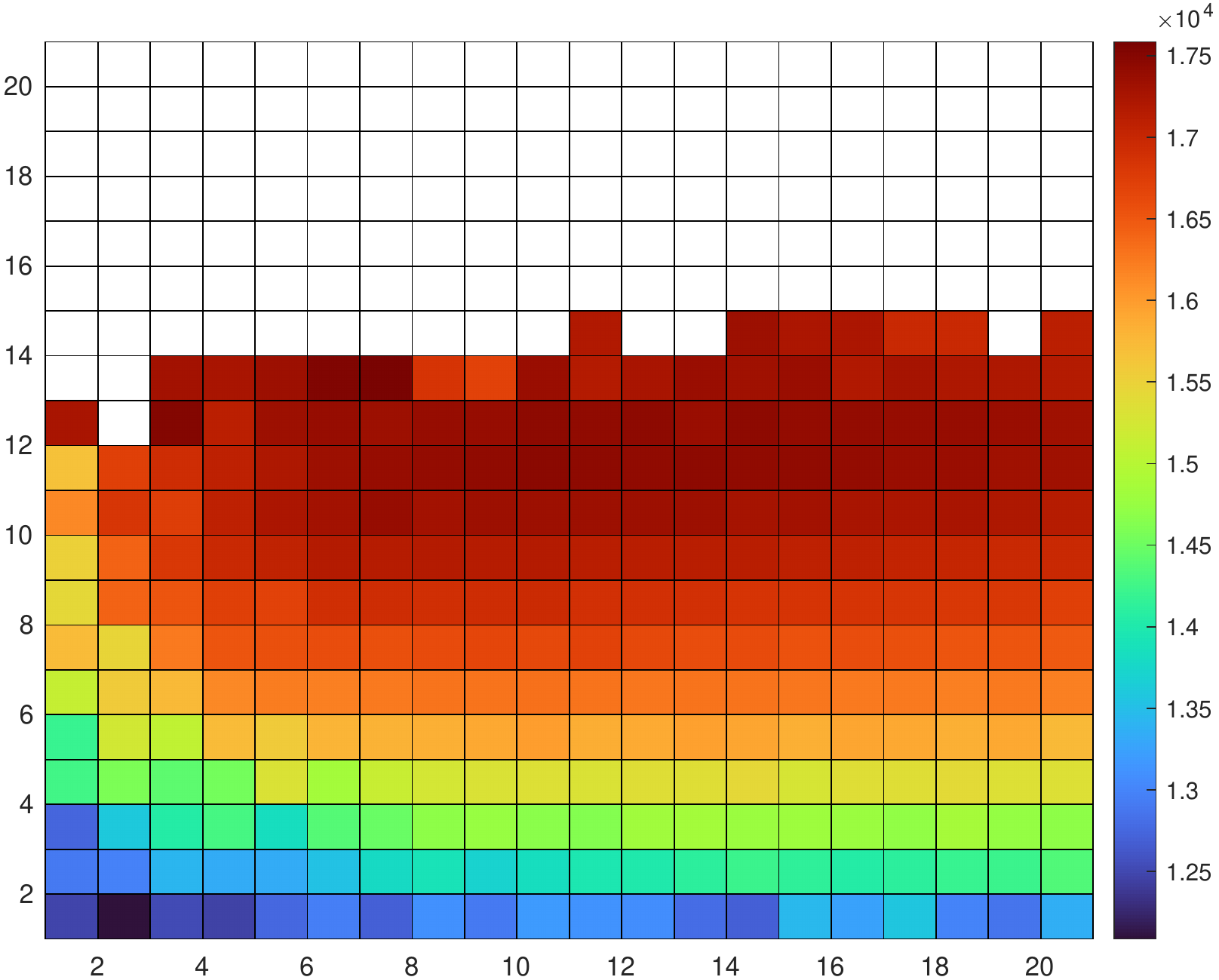}
\hskip5pt

% } % scalebox
% \squeezeup
% \squeezeup
% \squeezeup
% \squeezeup
\caption{The distribution of TTP solutions of the four competitors over the behaviour space on instance eil51\_n250\_uncorr-similar-weights\_01 (top), pr152\_n453\_uncorr\_01 (middle), and a280\_n279\_bounded-strongly-corr\_01 (bottom). The cells are coloured based on the average TTP scores of the solutions in the cell over ten independent runs. }
\label{fig:Maps_Quality}
% \squeezeup
% \squeezeup
% \squeezeup
% \squeezeup
\end{figure*}
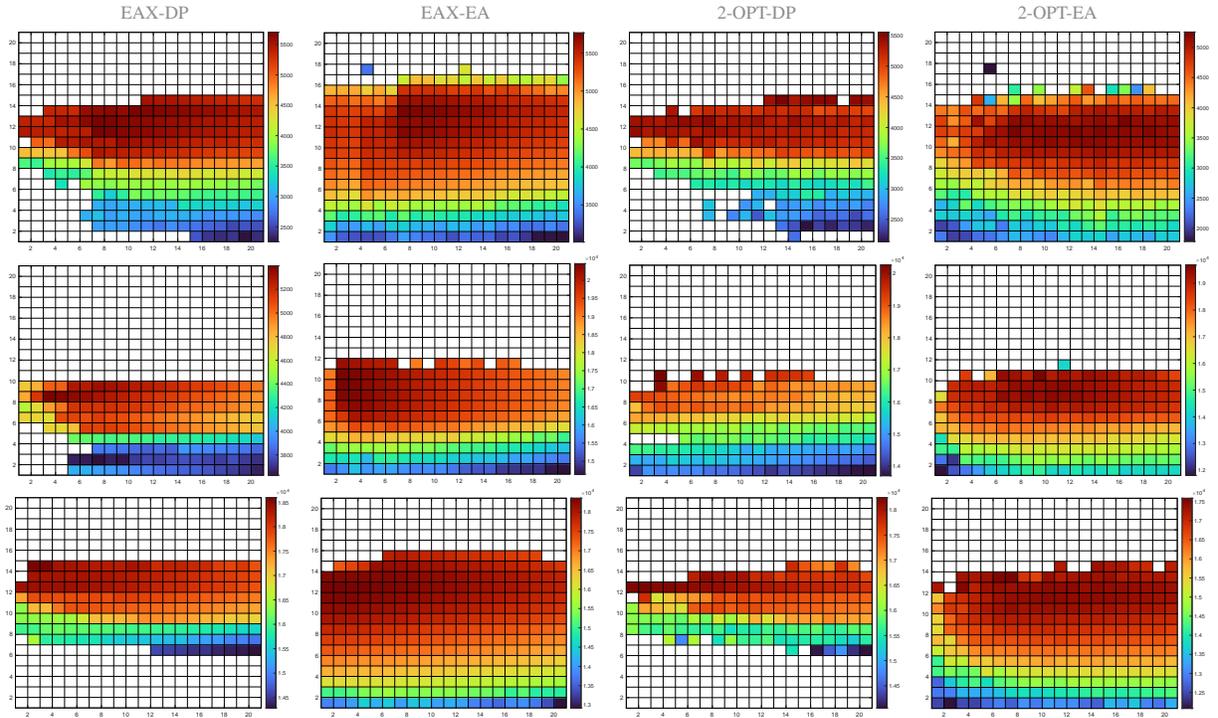

\begin{figure*}
\centering
\begin{tikzpicture}
\node (EAX-DP) at (-6.,-0.3) {\scriptsize{\textcolor{gray!90}{EAX-DP}}};
\node (EAX-EA) at (-2,-0.3) {\scriptsize{\textcolor{gray!90}{EAX-EA}}};
\node (2-OPT-DP) at (2,-0.3) {\scriptsize{\textcolor{gray!90}{2-OPT-DP}}};
\node (2-OPT-EA) at (6,-0.3) {\scriptsize{\textcolor{gray!90}{2-OPT-EA}}};
\end{tikzpicture}
% \scalebox{0.9}{
\includegraphics[width=.23\columnwidth]{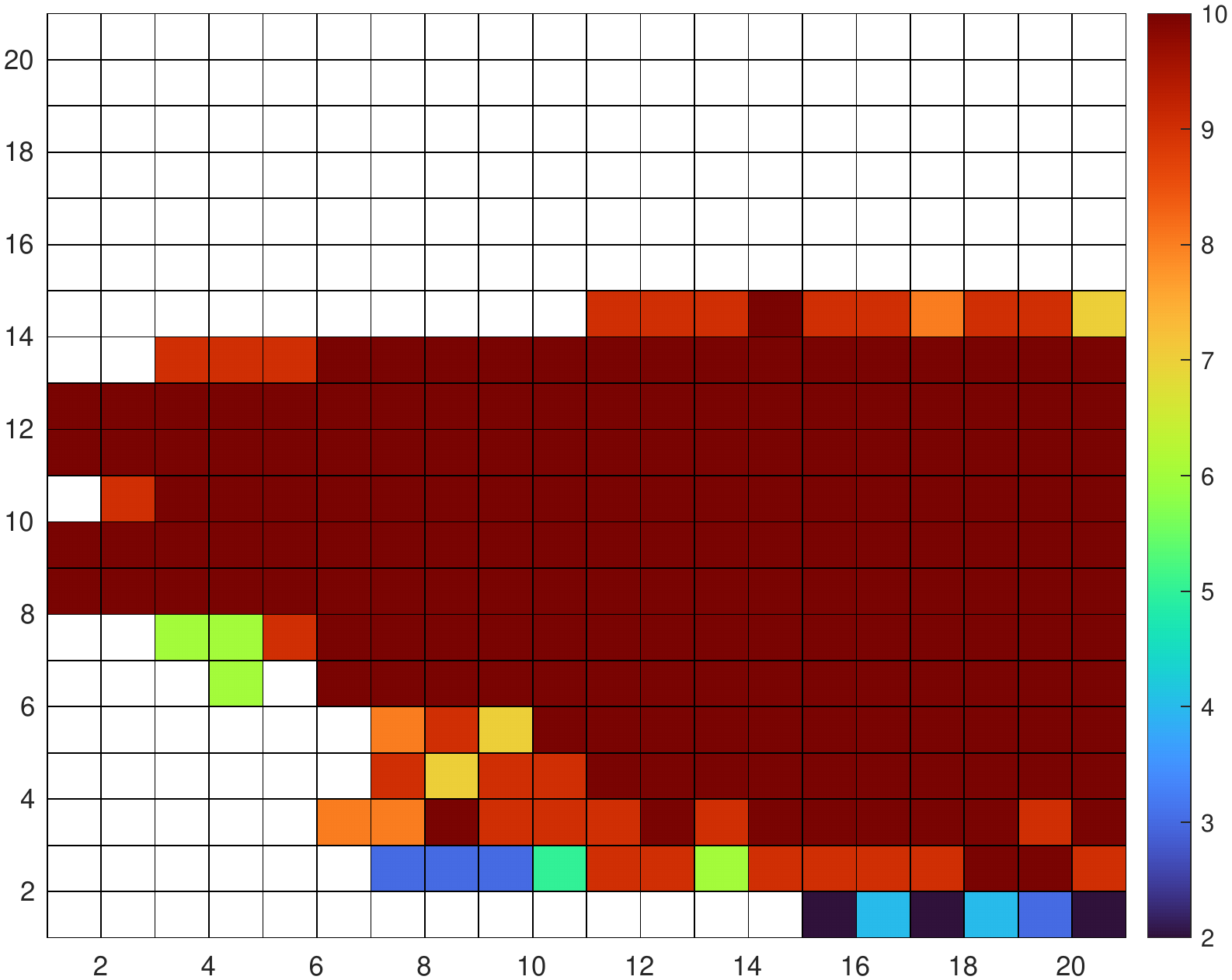}
\hskip5pt
\includegraphics[width=.23\columnwidth]{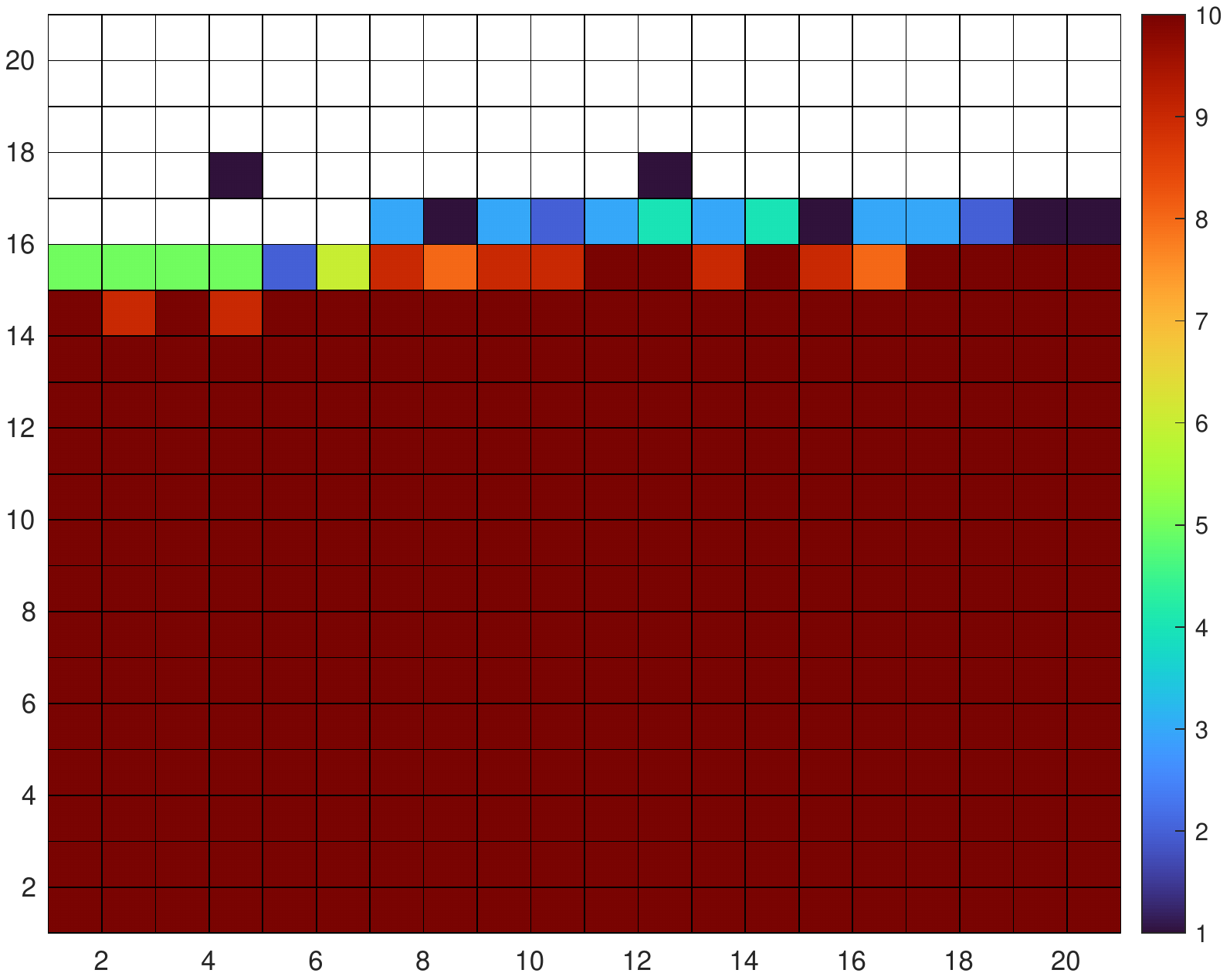}
\hskip5pt
\includegraphics[width=.23\columnwidth]{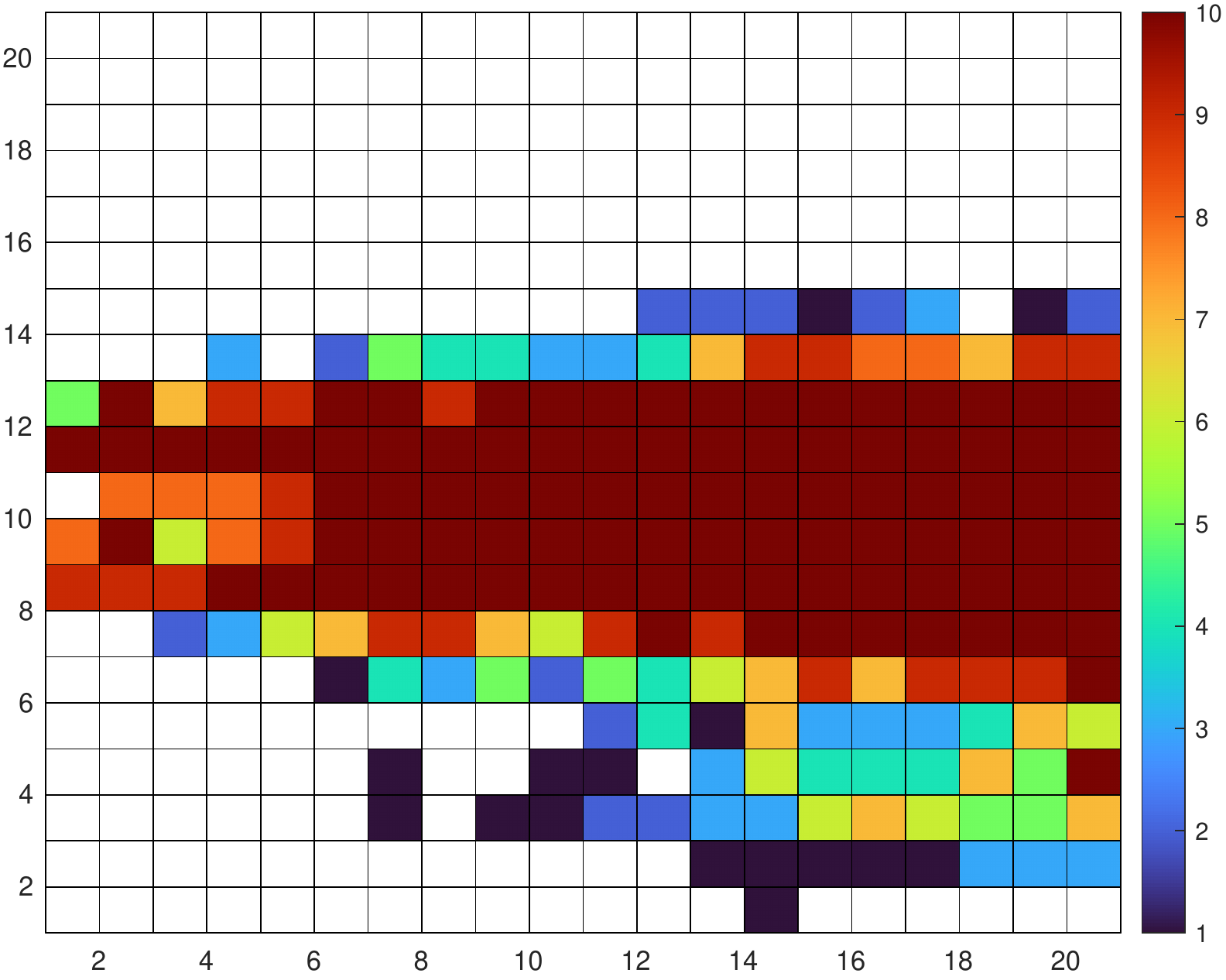}
\hskip5pt
\includegraphics[width=.23\columnwidth]{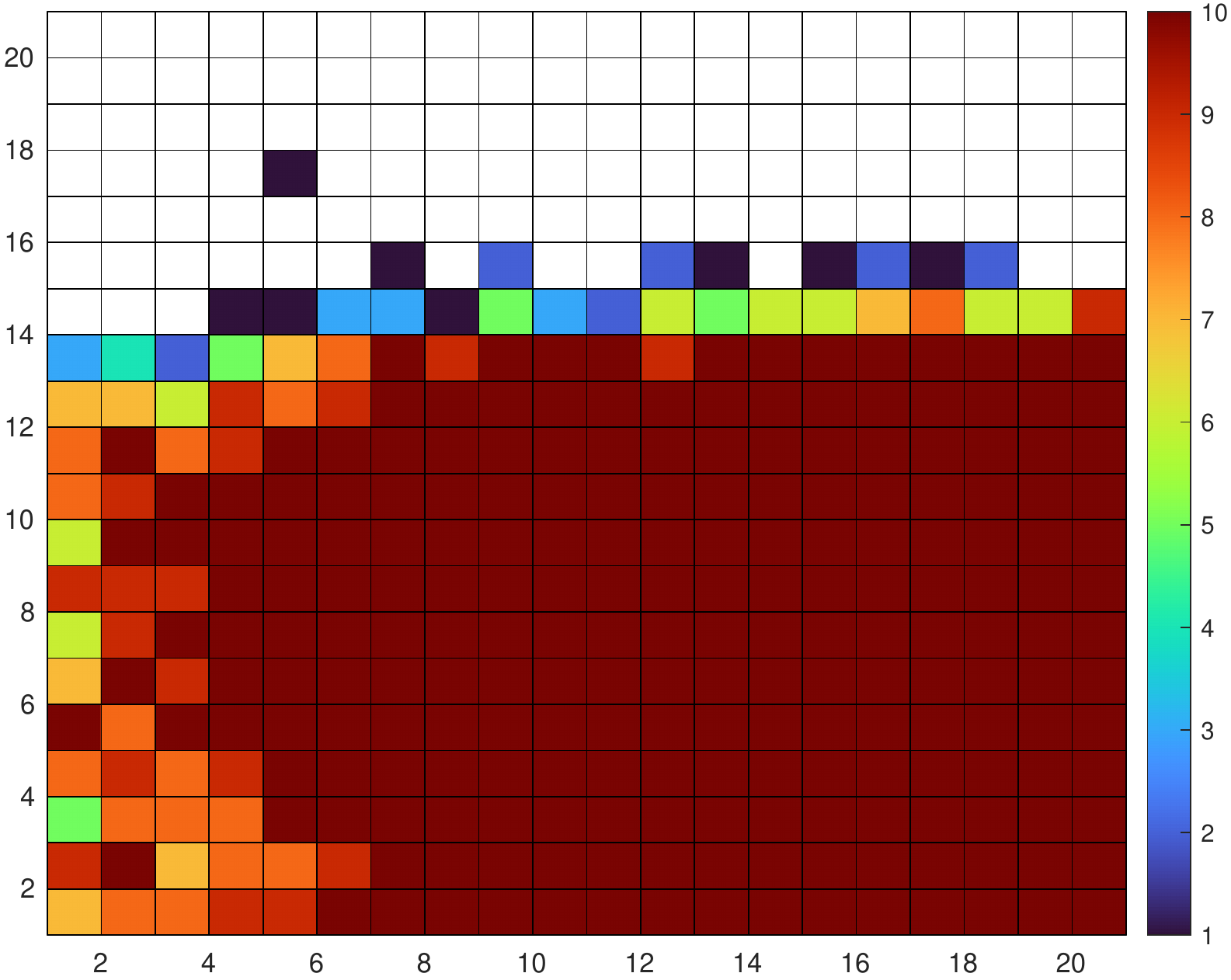}
\hskip5pt
\includegraphics[width=.23\columnwidth]{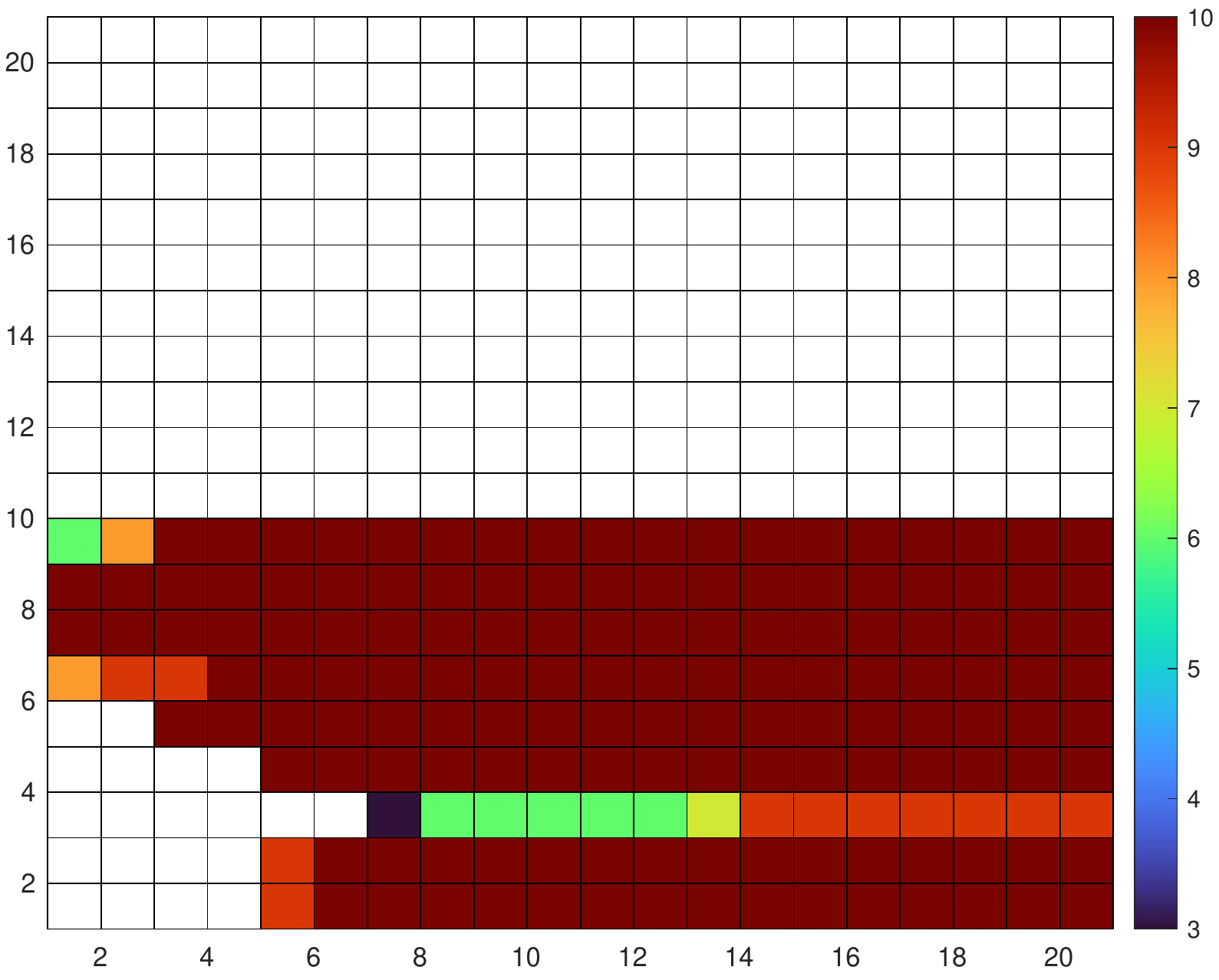}
\hskip5pt
\includegraphics[width=.23\columnwidth]{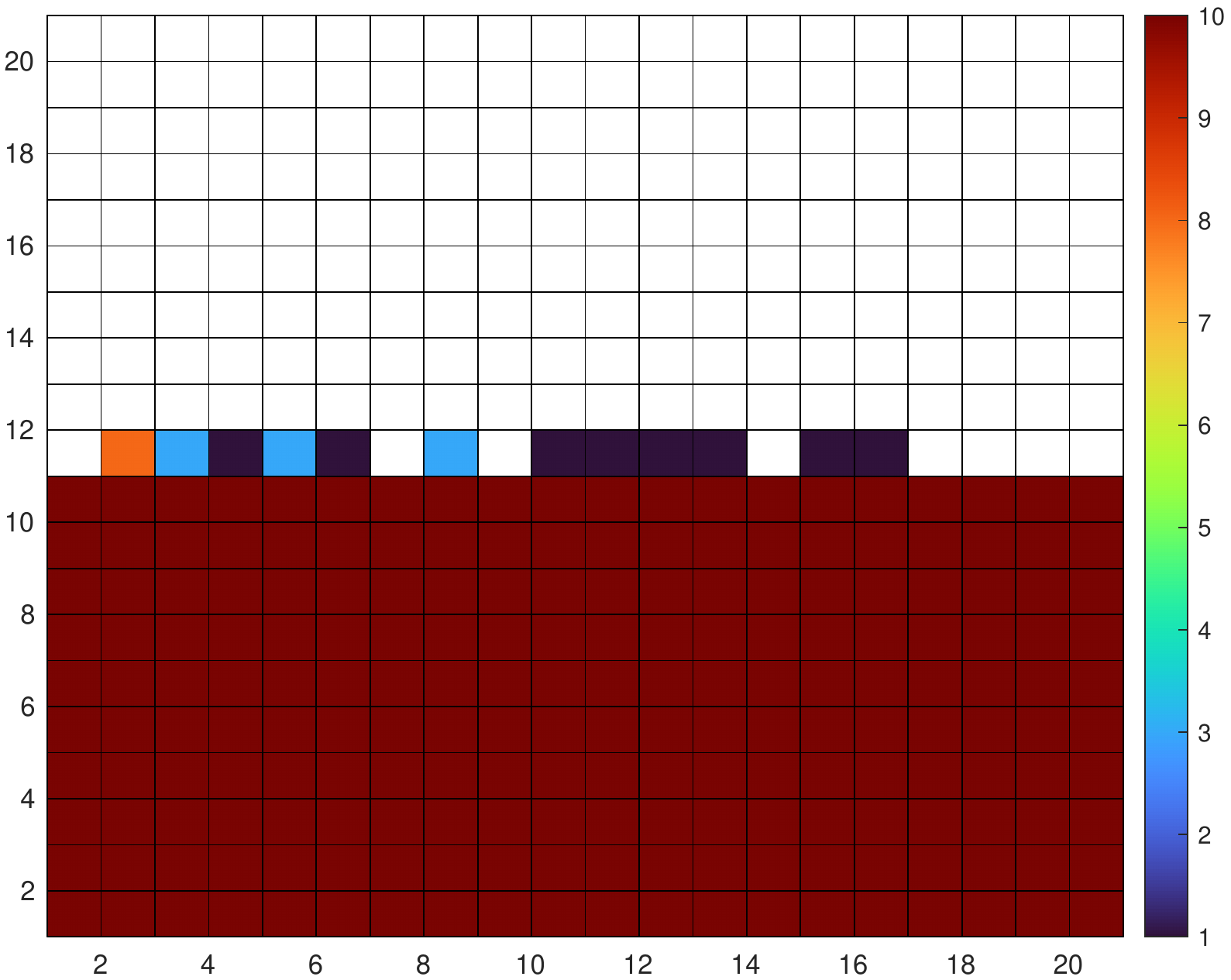}
\hskip5pt
\includegraphics[width=.23\columnwidth]{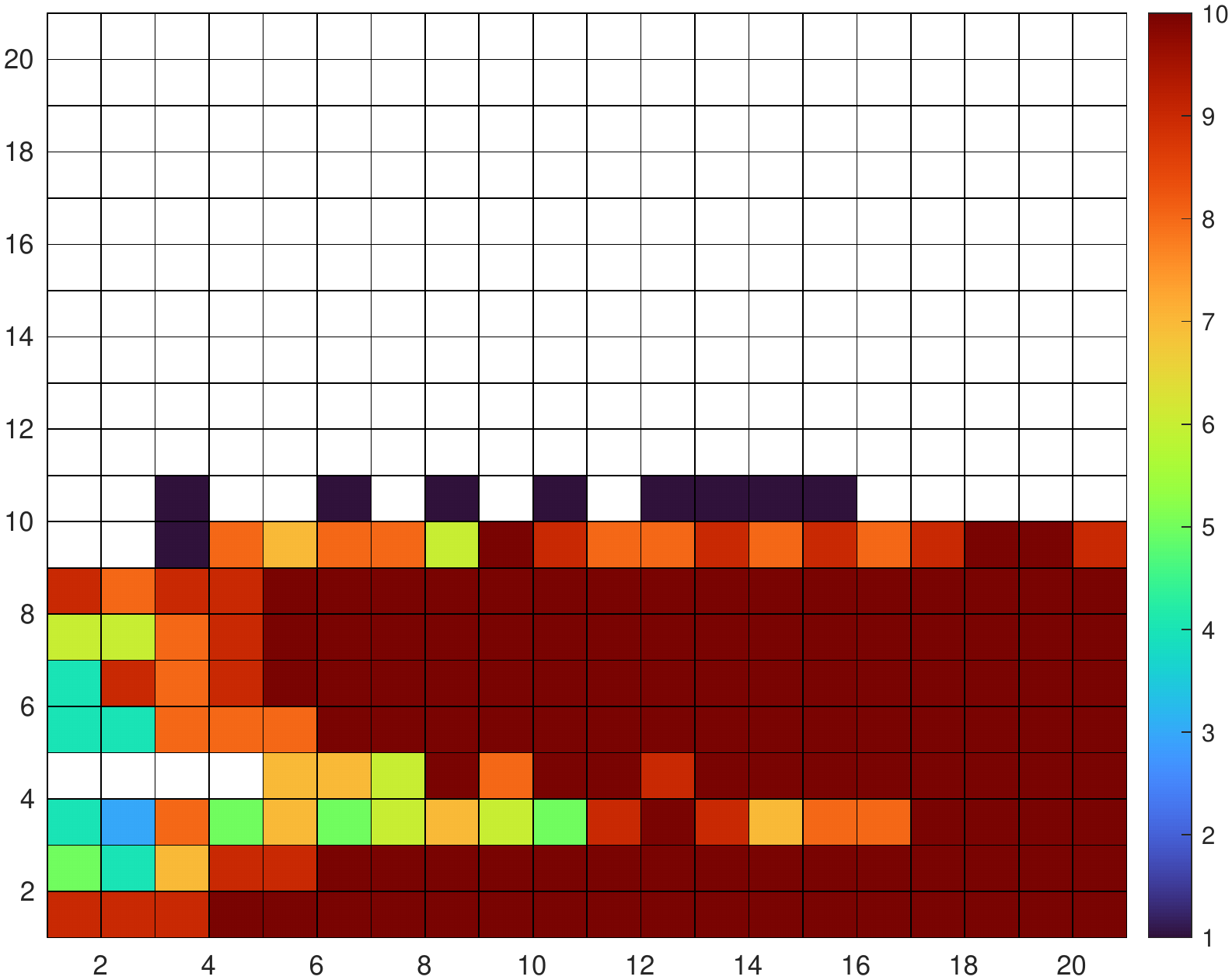}
\hskip5pt
\includegraphics[width=.23\columnwidth]{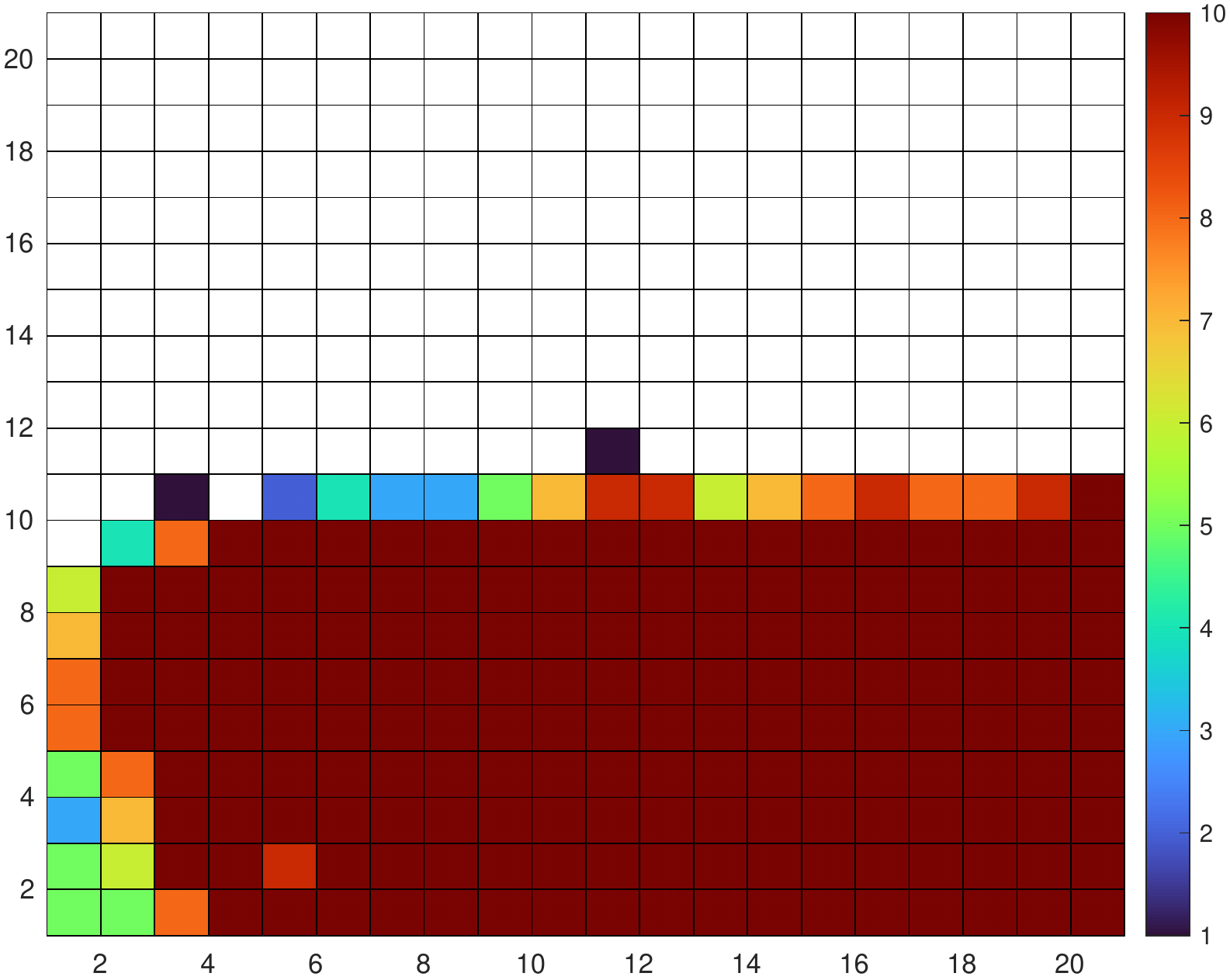}
\hskip5pt
\includegraphics[width=.23\columnwidth]{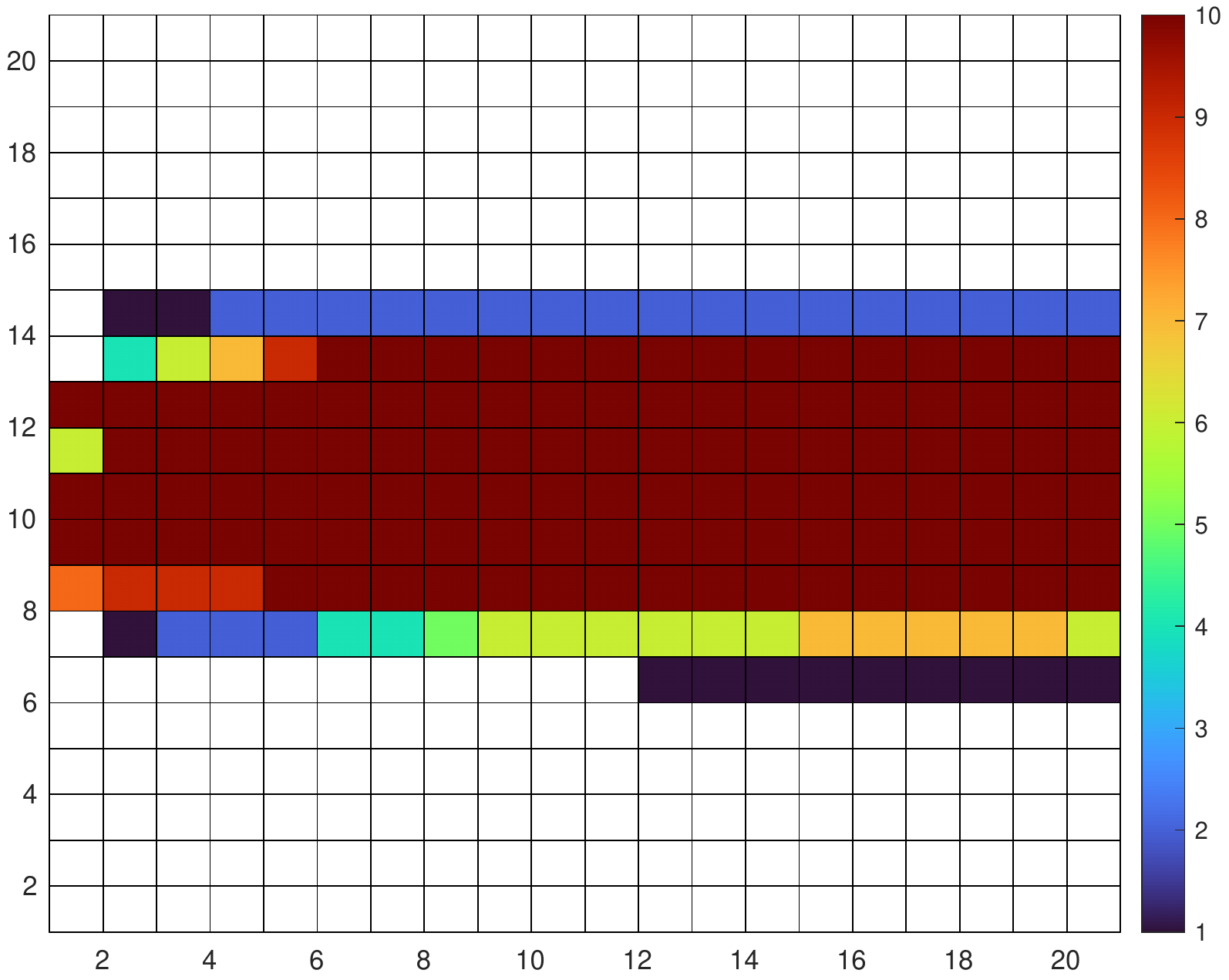}
\hskip5pt
\includegraphics[width=.23\columnwidth]{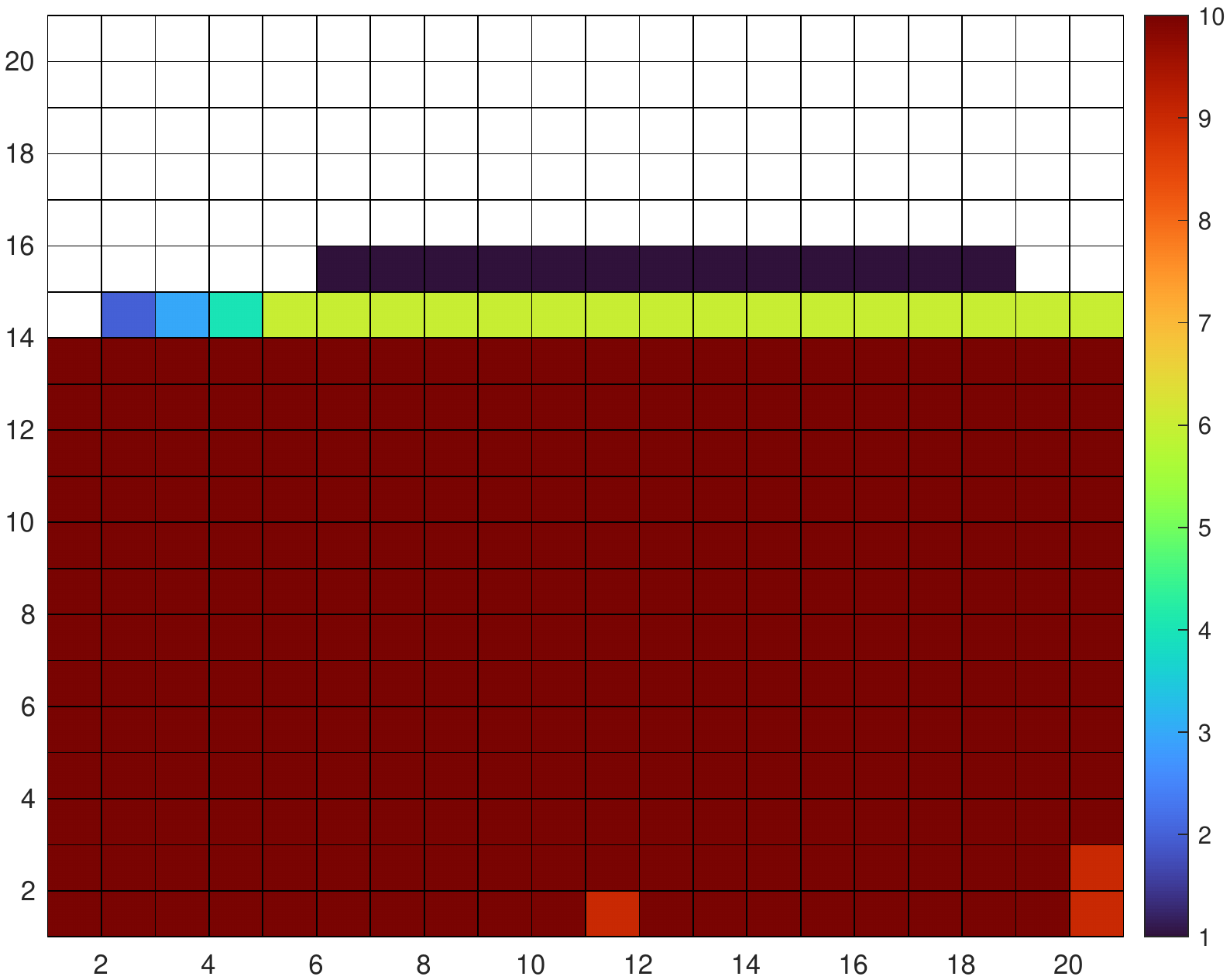}
\hskip5pt
\includegraphics[width=.23\columnwidth]{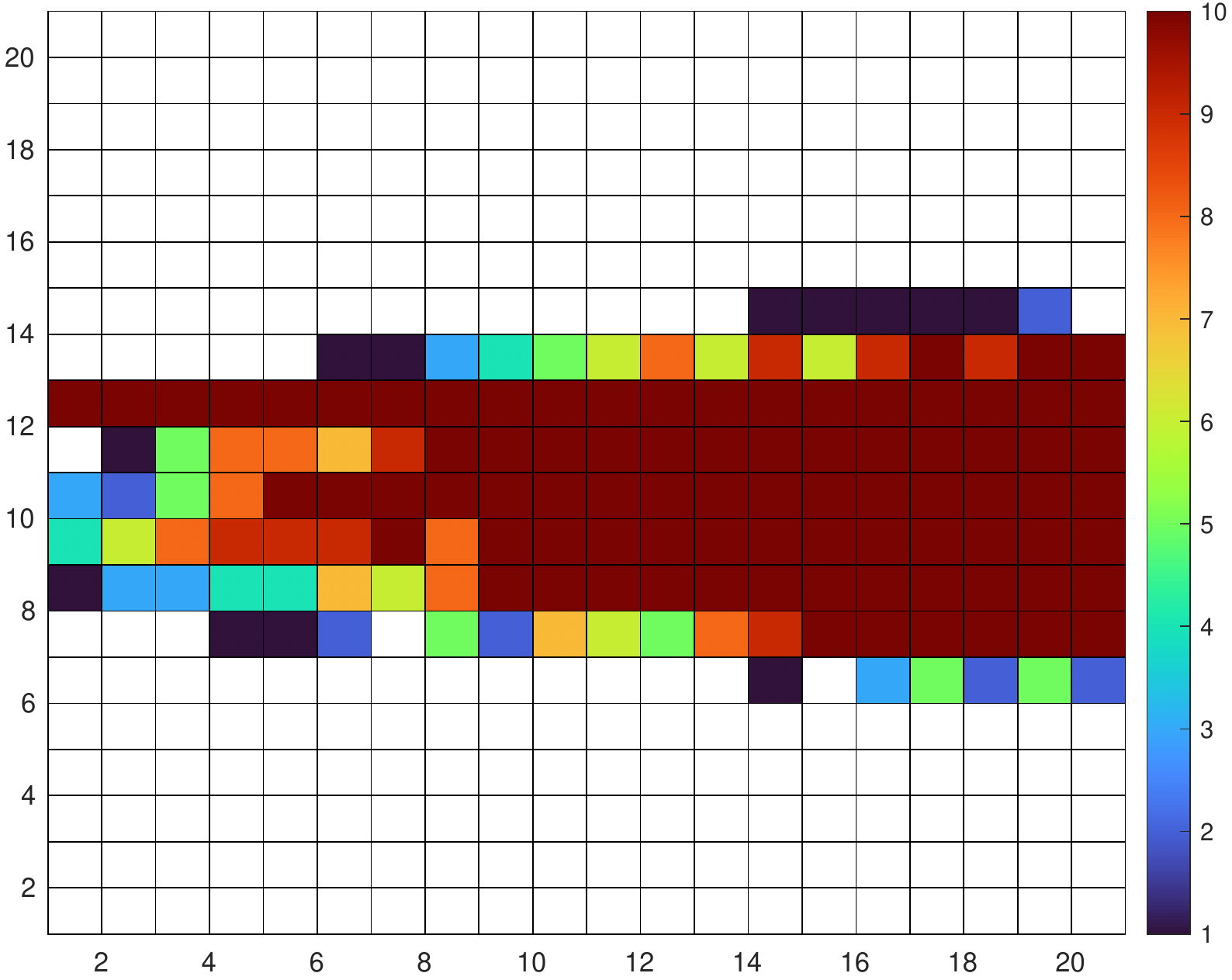}
\hskip5pt
\includegraphics[width=.23\columnwidth]{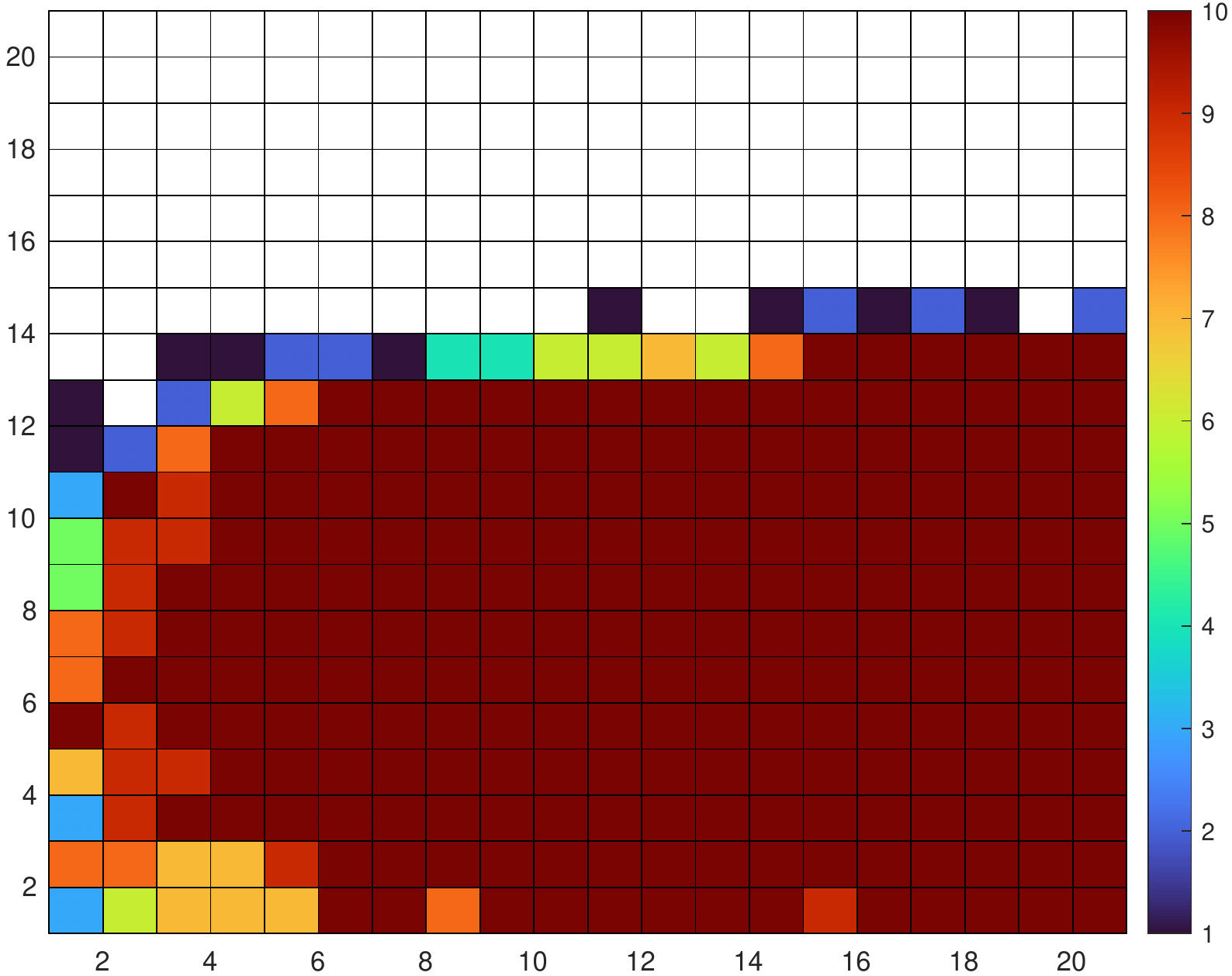}
\hskip5pt

% \begin{tikzpicture}
% \node[opacity=.55] (legend) at (0,0) {\includegraphics[width=1.2\columnwidth,trim=20pt 40pt 10pt 0,clip]{turbo_palette_legend.png}};
% %\node (tit) at (-4.4,0) {\scriptsize{Frequency:}};
% \node (low) at (-3.6,-0.3) {\scriptsize{\textcolor{gray!90}{low}}};
% \node (medium) at (0,-0.3) {\scriptsize{\textcolor{gray!90}{medium}}};
% \node (high) at (3.6,-0.3) {\scriptsize{\textcolor{gray!90}{high}}};
% \end{tikzpicture}
% } % scalebox
\squeezeup
\squeezeup
% \squeezeup
% \squeezeup
\caption{The frequency of cells housing a TTP solution over 10 independent runs on on instance eil51\_n250\_uncorr-similar-weights\_01 (top), pr152\_n453\_uncorr\_01 (middle), and a280\_n279\_bounded-strongly-corr\_01 (bottom).}
\label{fig:Mao_Frequency}
\squeezeup
\squeezeup
% \squeezeup
% \squeezeup
\end{figure*}
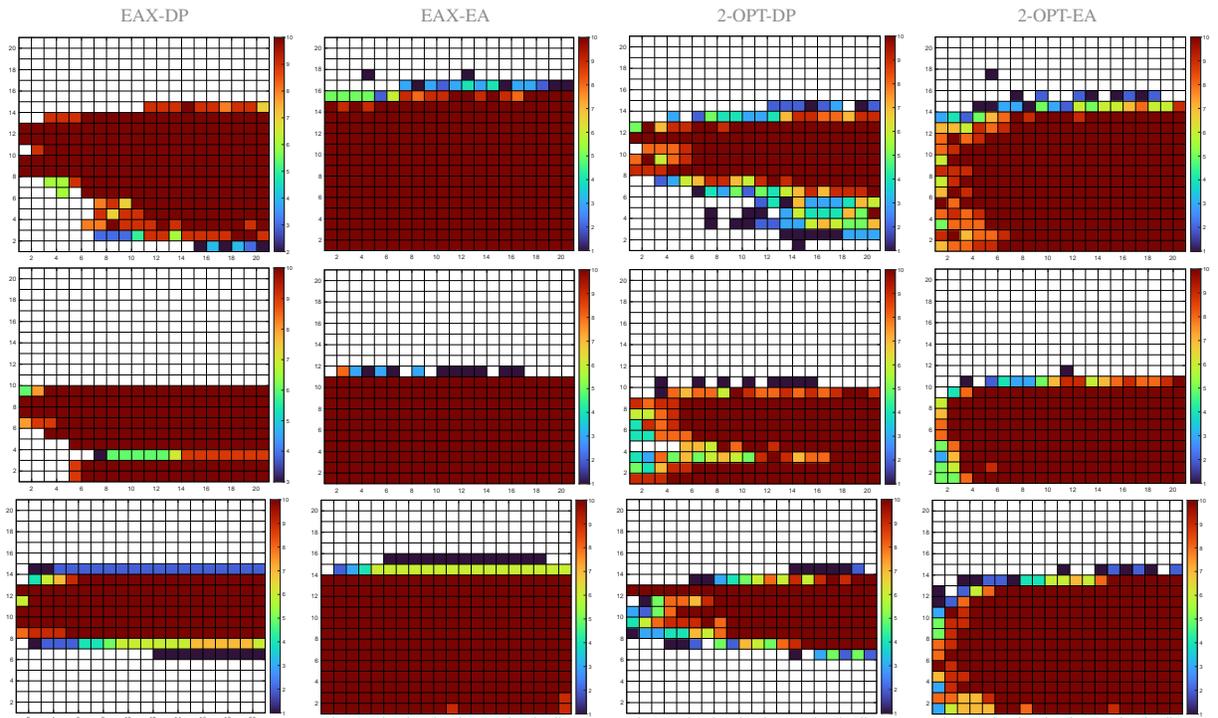
\subsection{Analysis of the maps}

This section visualises and scrutinises the final map obtained from the BMBEA using different search operators, namely EAX, 2-OPT, DP, and $(1+1)$EA. Figure~\ref{fig:Maps_Quality} visualises the final maps obtained from the four competitors in instances 6, 15, and 16. The TSP value increases when we move in the direction of $x$ axis, while moving in the $y$ axis results in a rise in the KP score. Since the TSP is a minimisation and the KP is a maximisation problem, Cell (1,20) consists of the solution with a BD closest to $f^*$ and $g^*$. The maps' cells are coloured based on the average TTP score of the solution within the cells over 10 independent runs; the hotter colour, the higher the TTP score. We can observe that the west part of the maps tends to contain better TTP solutions. In 8 out of 9 cases, the best solutions are located in a BD of $(1, 1.005) f^*$ and $(0.9, 0.95) g^*$. Moreover, the figure depicts that the maps obtained from BMBEAs using EAX have more hot-coloured cells than the ones with 2-OPT have, which shows the consistency of EAX in generating high-quality solutions. Turning to the comparison between DP and  $(1+1)$EA, the latter can populate a larger part of the map.   

Figure \ref{fig:Mao_Frequency} illustrates the frequency of cells containing a solution over ten independent runs. The instances are the same as Figure \ref{fig:Maps_Quality}. Here, a hotter colour indicates a higher frequency. The figure depicts that the algorithms cannot populate the cells close to the optimal KP. Because the algorithms compute the packing list as the second level of a bi-level optimisation procedure. Thus, the KP values are constrained by the given tour.
Interestingly, most cells corresponding to the KP values close to $(1-\alpha_2)g^*$ also remain empty for the same reason, especially when DP is used. Moreover, one may notice that the most red-coloured cells in Figure \ref{fig:Maps_Quality} are coloured red here as well. It illustrates a proportional relationship between the quality of solutions and the frequency. 
Furthermore, the cells associated with low TSP values (left) of maps are more likely to be empty than the other side.
% It shows that having a high-quality tour enables us to use a higher proportion of the knapsack capacity. Because the thief should travel faster to compensate for the low-quality of the tour, and to travel faster, they should travel lighter. In other words, the optimal packing lists usually utilise a higher proportion of the capacity for shorter tours, resulting in a higher KP score. This is one of the interesting inter-dependency between the KP and the TSP, which the map illustrates. More importantly, 
As the TSP value increases, so does the number of tours resulting in such a TSP value rise. This results in a more diverse set of tours and eventually a more diverse set of packing lists and a broader range of the KP score.   

\subsubsection{MAP-Elitism vs. $(\mu+1)$EA}
\label{subsec:exp_map_mu+1}
In this section, we compare the sets of solutions obtained by BMBEA and $(\mu+1)$EA. Both algorithms have identical initialisation, parent selection, and generating offspring. Therefore, the difference between these algorithms is limited to survival selection. The aim is to investigate the impact of MAP-Elitism.

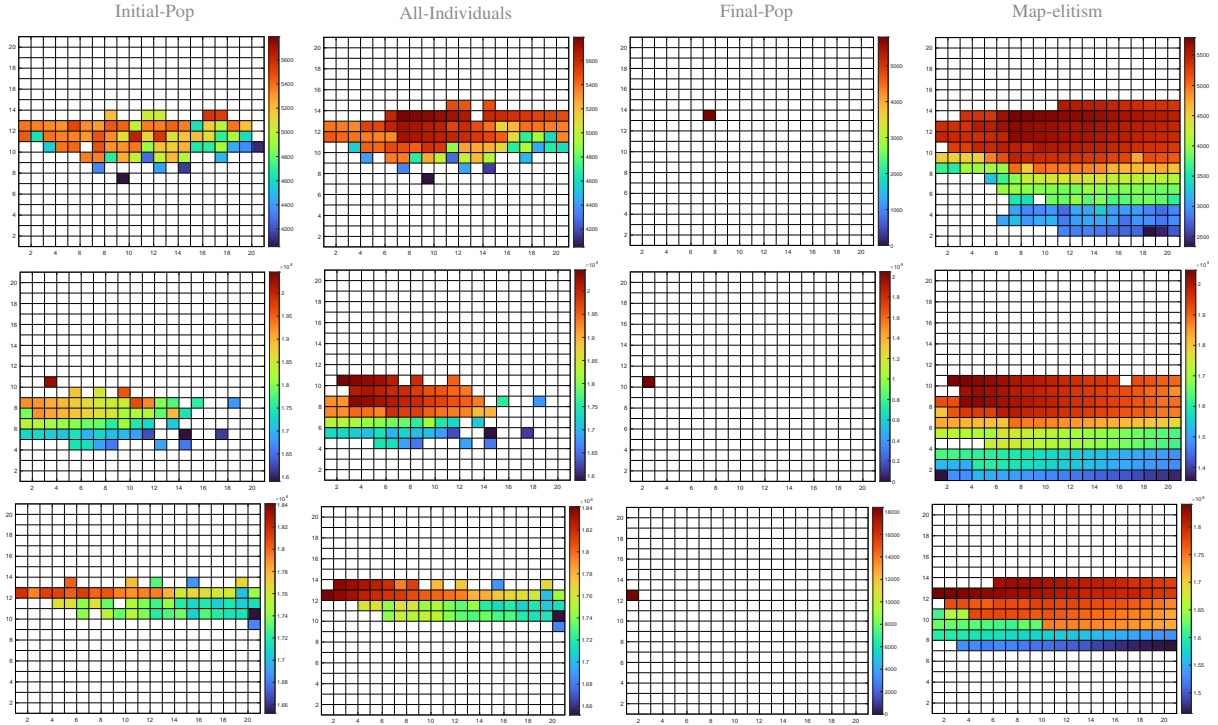
\begin{figure*}
\centering
\begin{tikzpicture}

%\node (tit) at (-4.4,0) {\scriptsize{Frequency:}};
\node (EAX-DP) at (-6,-0.3) {\scriptsize{\textcolor{gray!90}{Initial-Pop}}};
\node (EAX-EA) at (-2,-0.3) {\scriptsize{\textcolor{gray!90}{All-Individuals}}};
\node (2-OPT-DP) at (2,-0.3) {\scriptsize{\textcolor{gray!90}{Final-Pop}}};
\node (2-OPT-EA) at (6,-0.3) {\scriptsize{\textcolor{gray!90}{Map-elitism}}};
\end{tikzpicture}
% \scalebox{0.9}{
\includegraphics[width=.23\columnwidth]{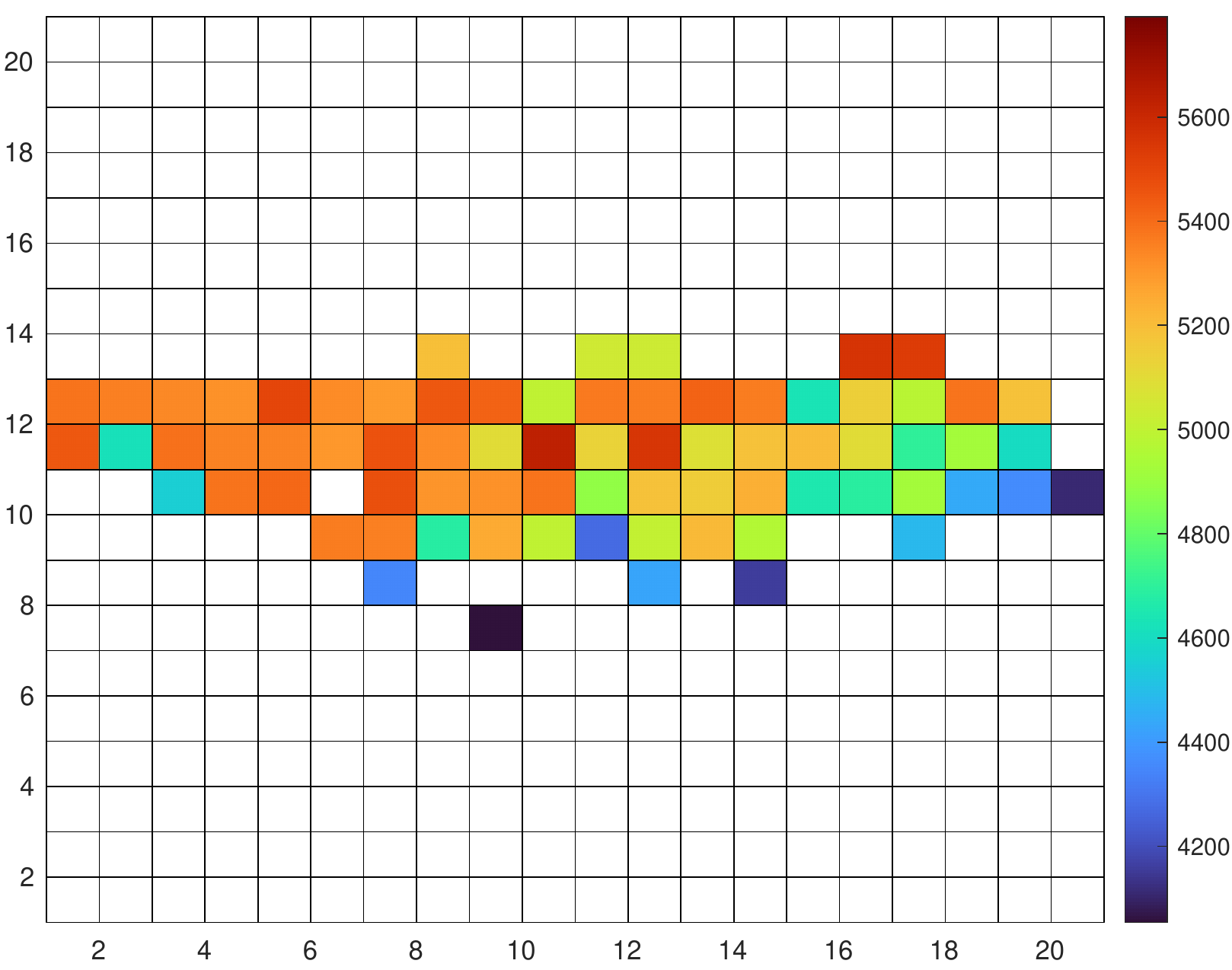}
\hskip5pt
\includegraphics[width=.23\columnwidth]{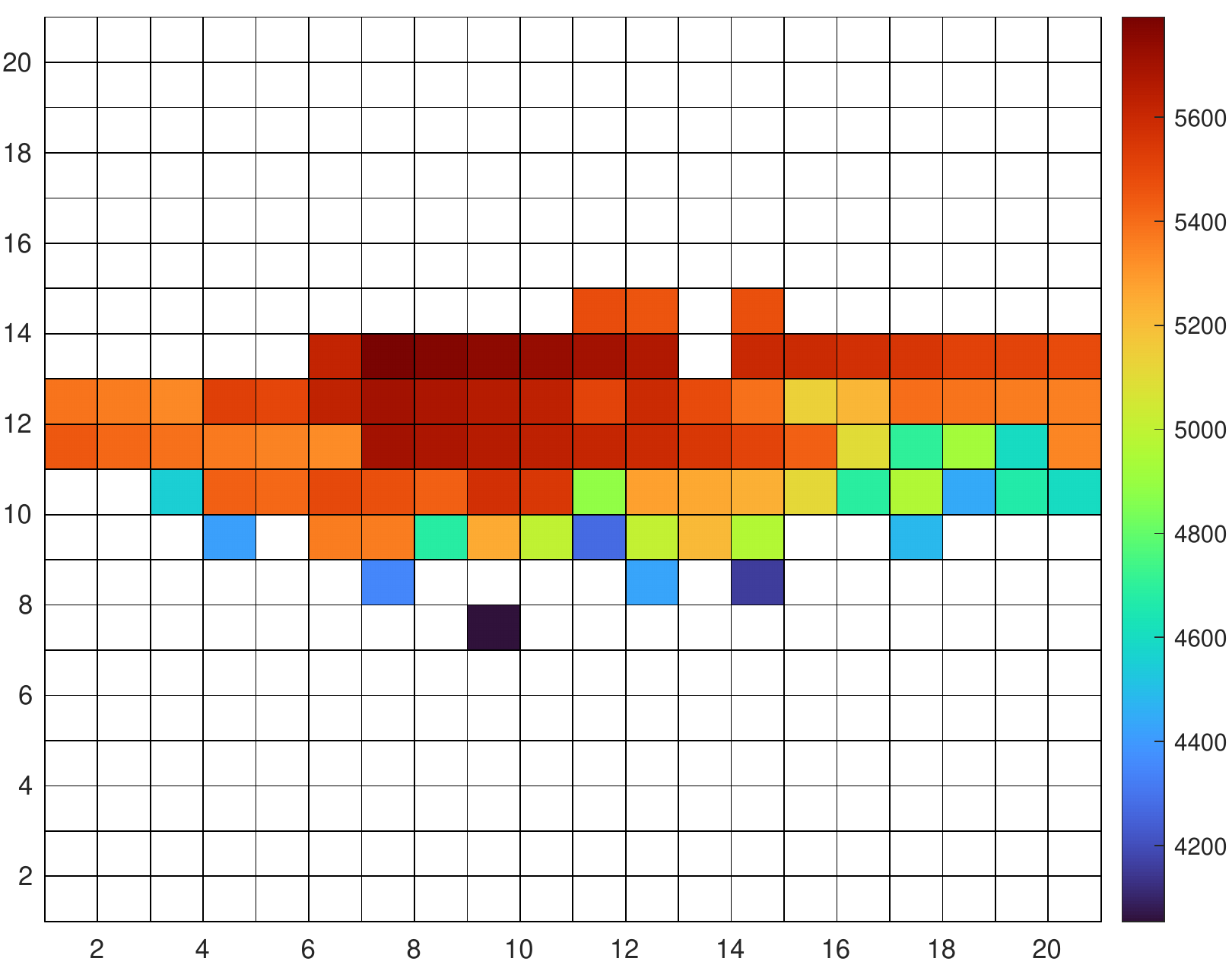}
\hskip5pt
\includegraphics[width=.23\columnwidth]{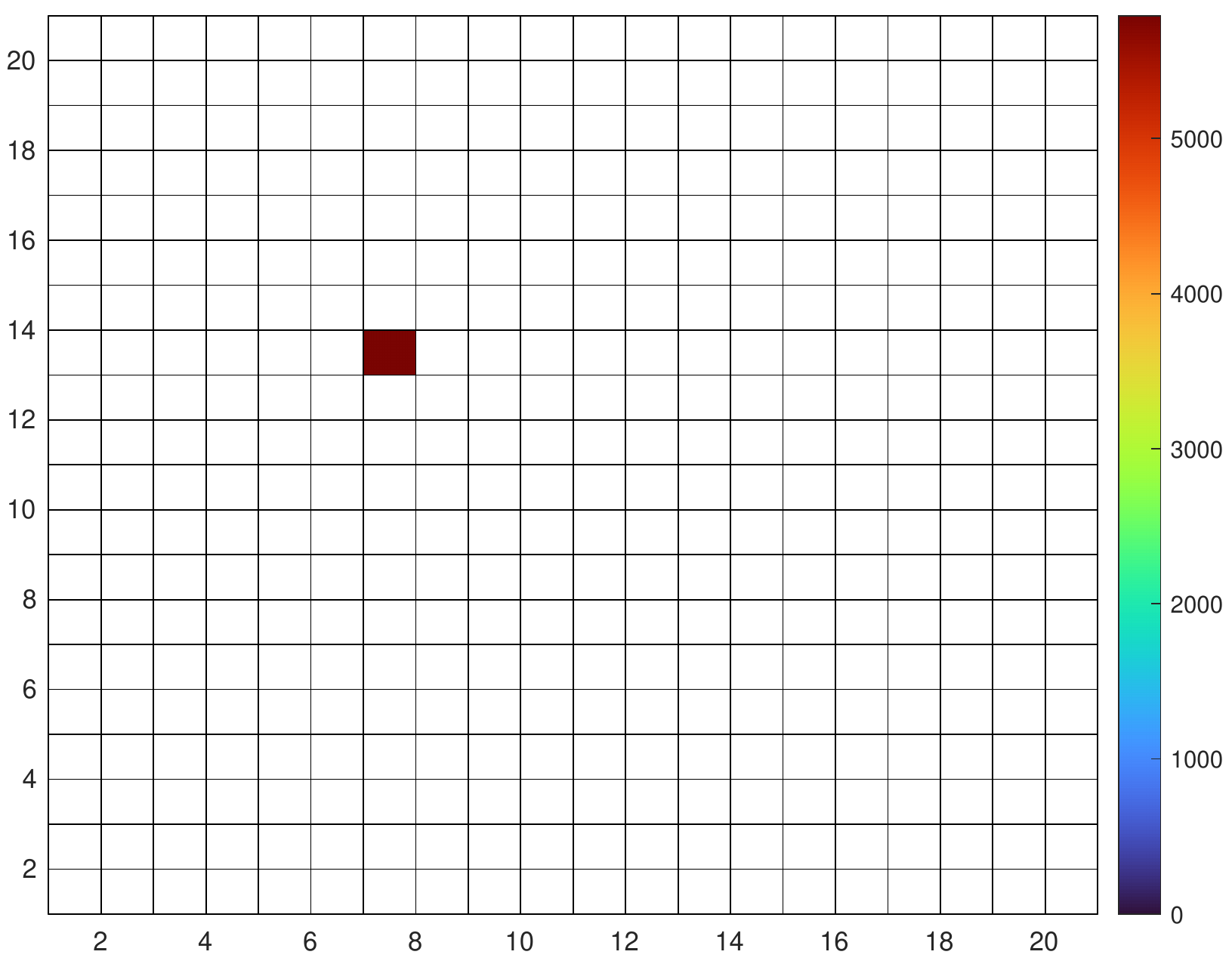}
\hskip5pt
\includegraphics[width=.23\columnwidth]{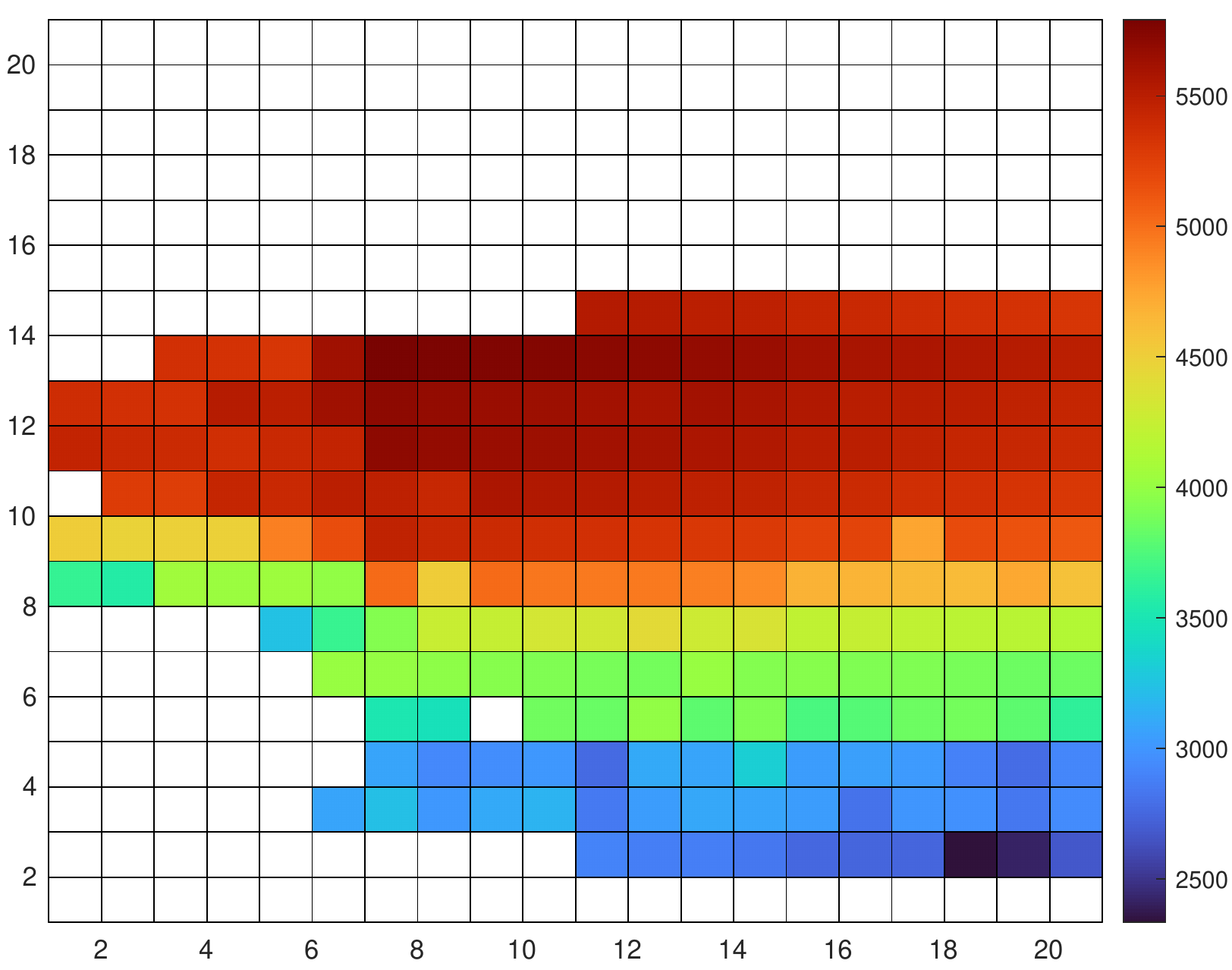}
\hskip5pt
\includegraphics[width=.23\columnwidth]{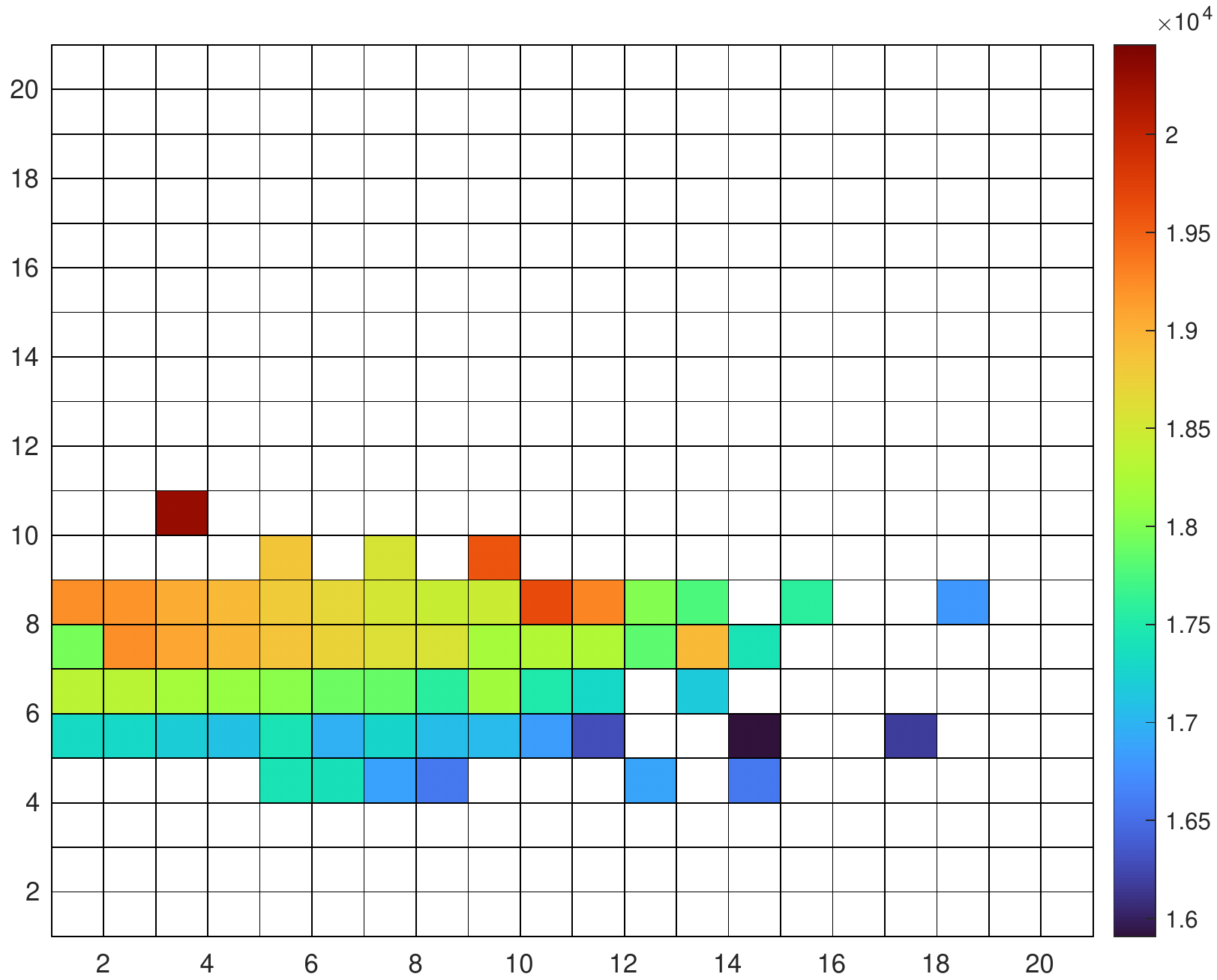}
\hskip5pt
\includegraphics[width=.23\columnwidth]{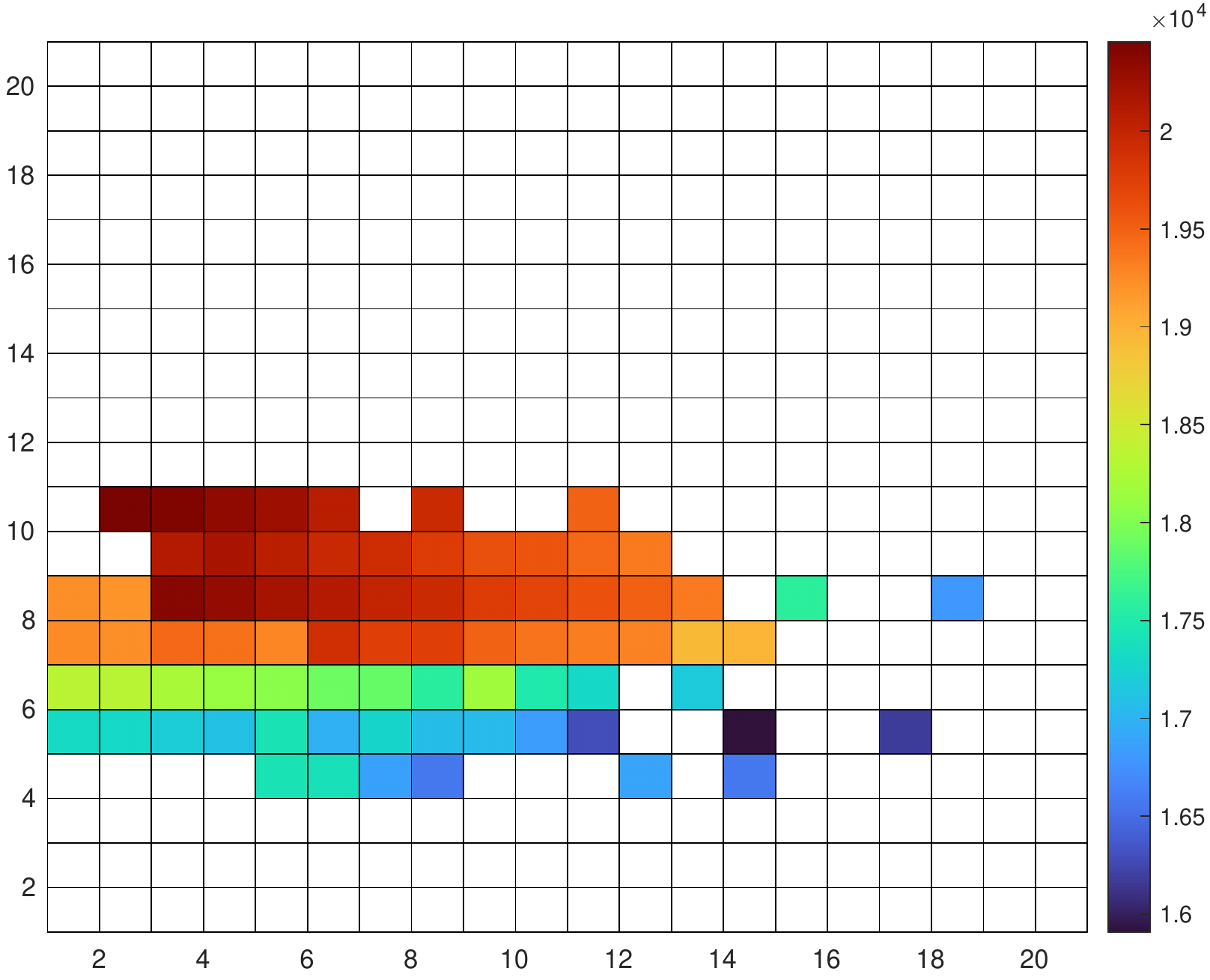}
\hskip5pt
\includegraphics[width=.23\columnwidth]{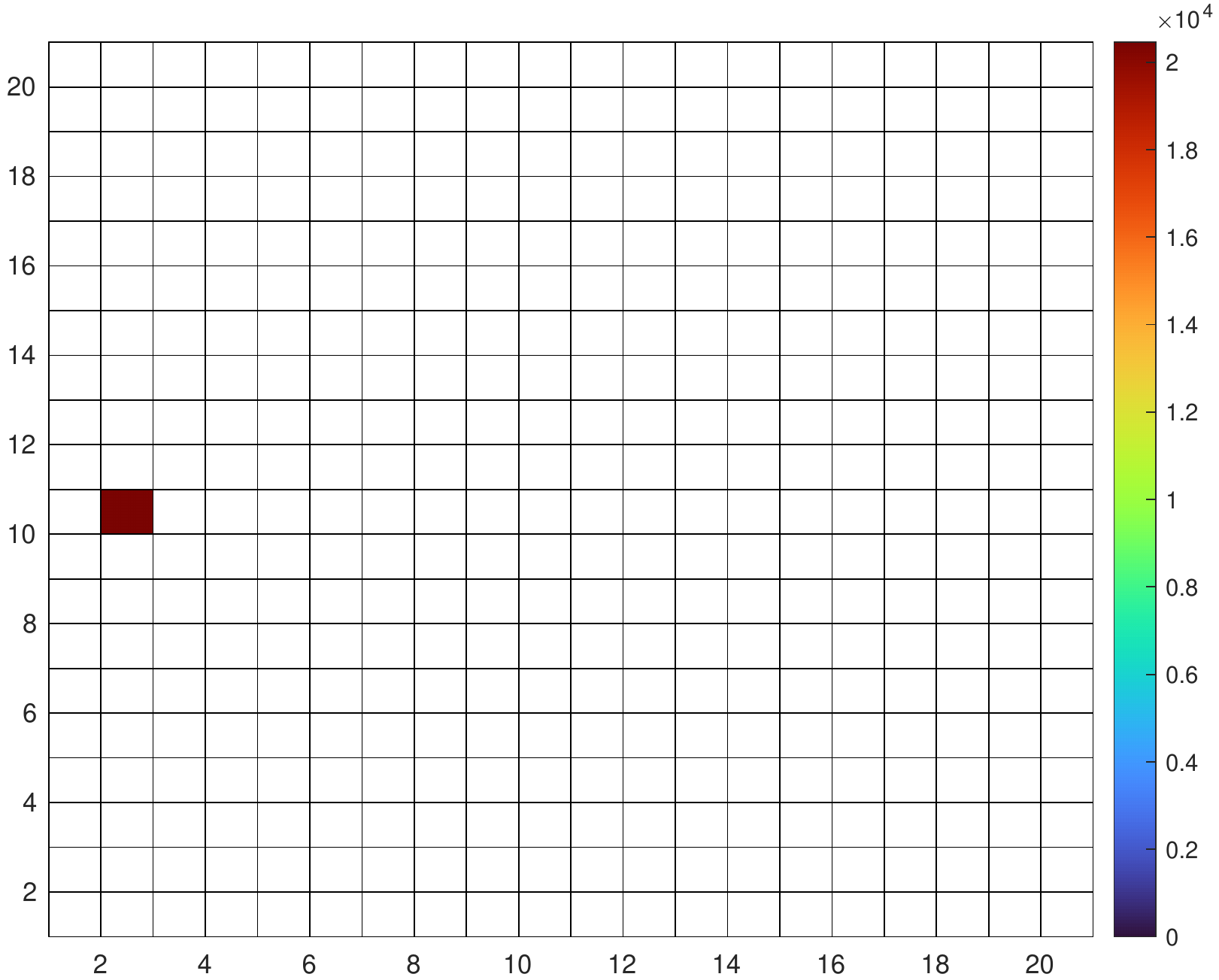}
\hskip5pt
\includegraphics[width=.23\columnwidth]{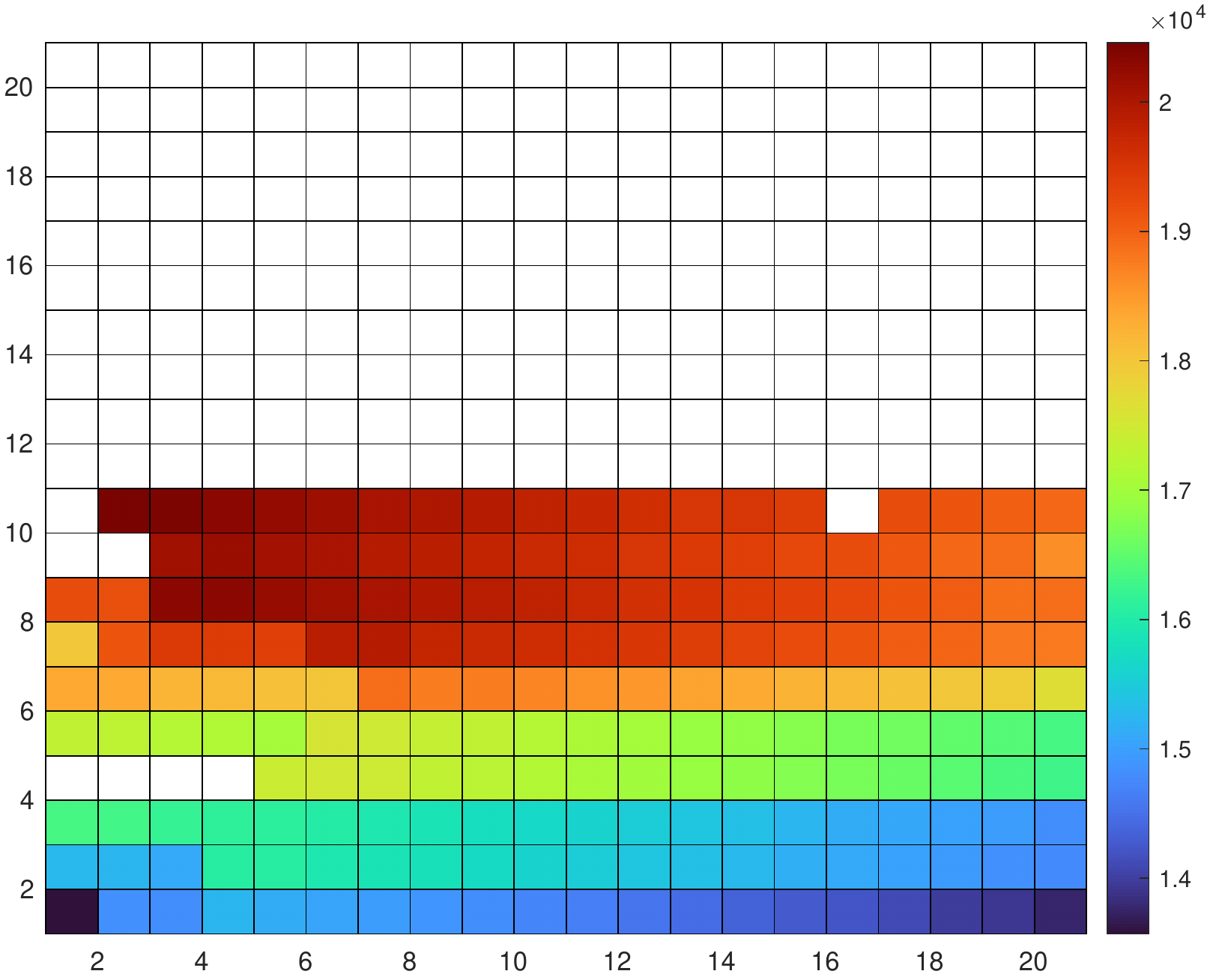}
\hskip5pt
\includegraphics[width=.23\columnwidth]{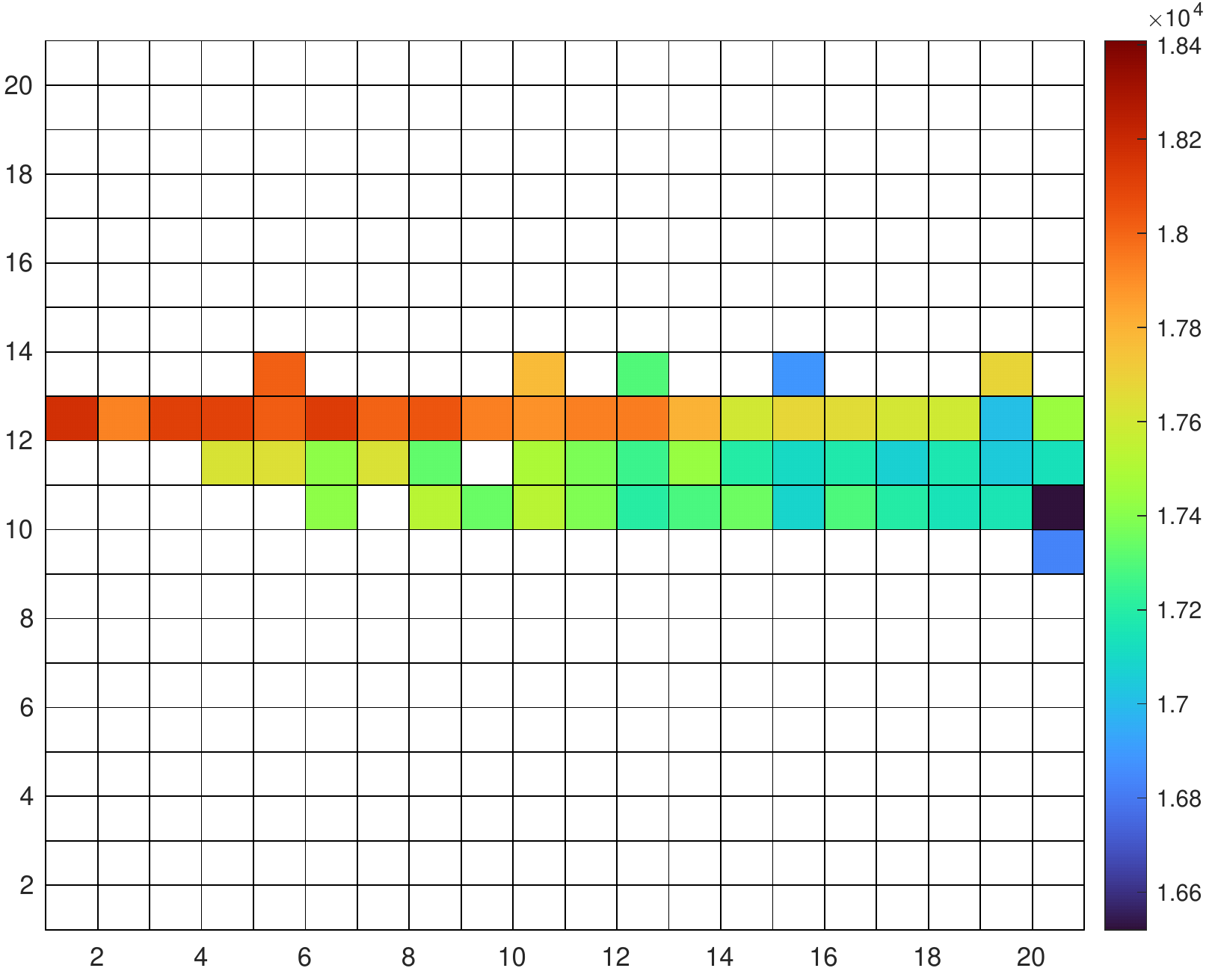}
\hskip5pt
\includegraphics[width=.23\columnwidth]{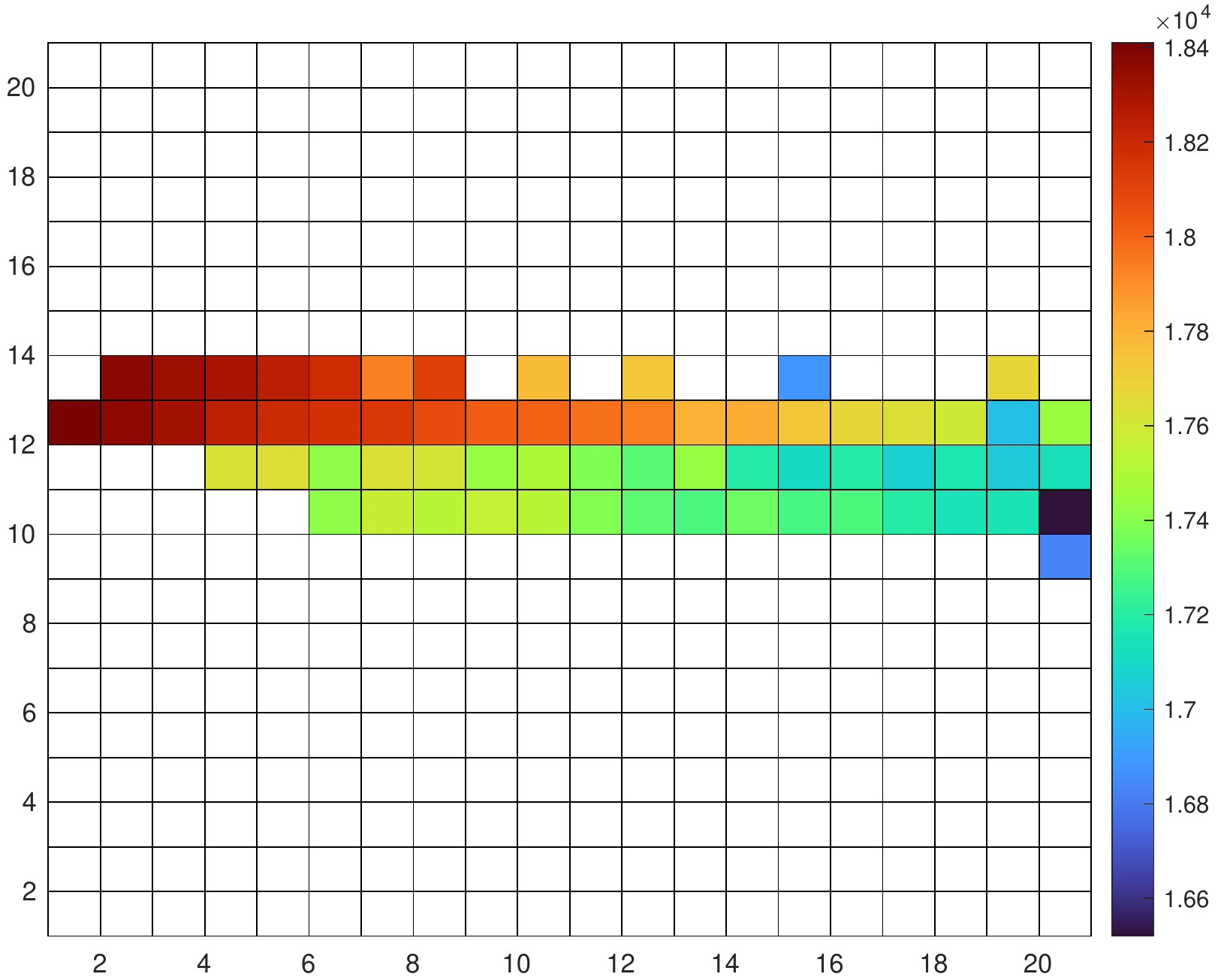}
\hskip5pt
\includegraphics[width=.23\columnwidth]{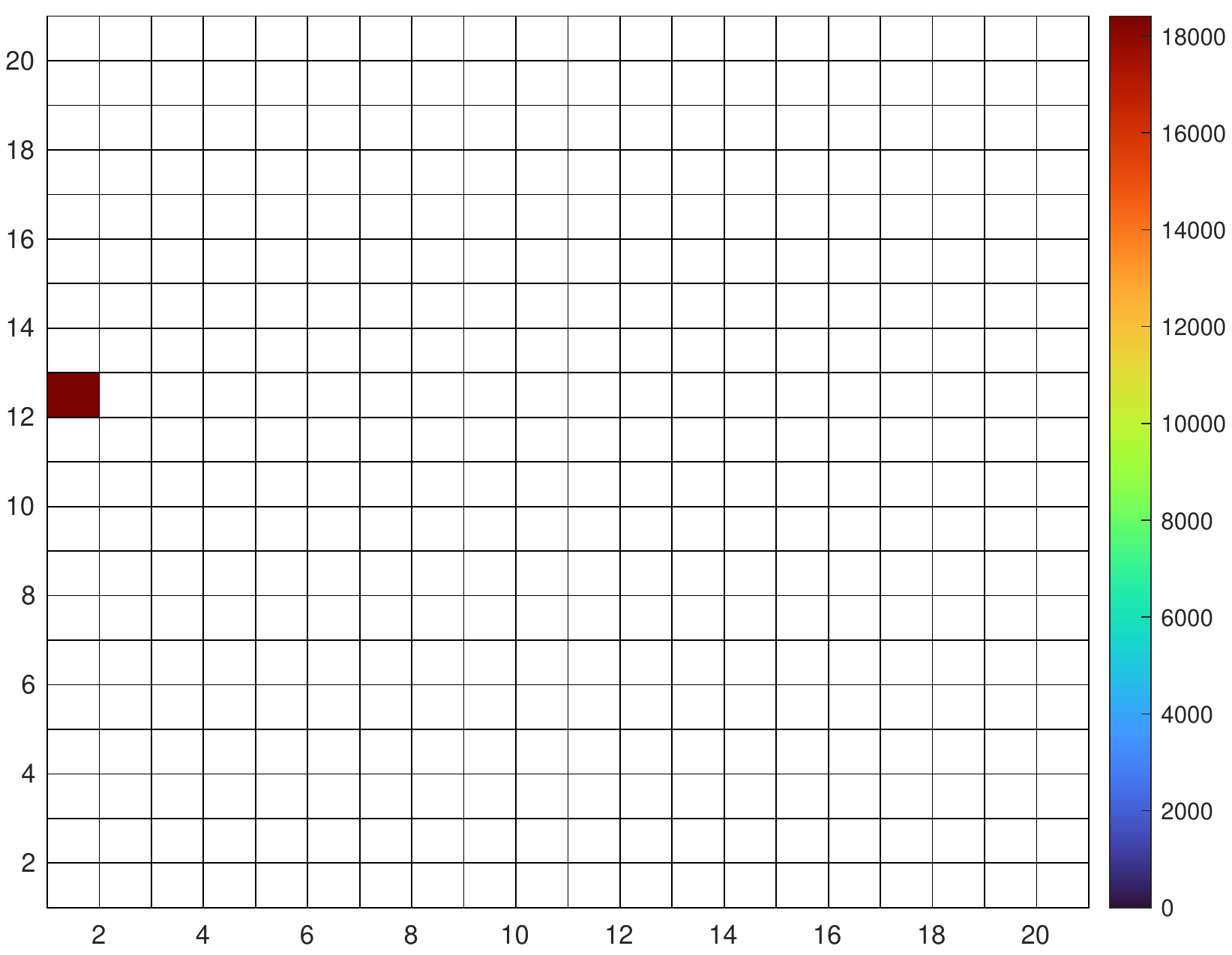}
\hskip5pt
\includegraphics[width=.23\columnwidth]{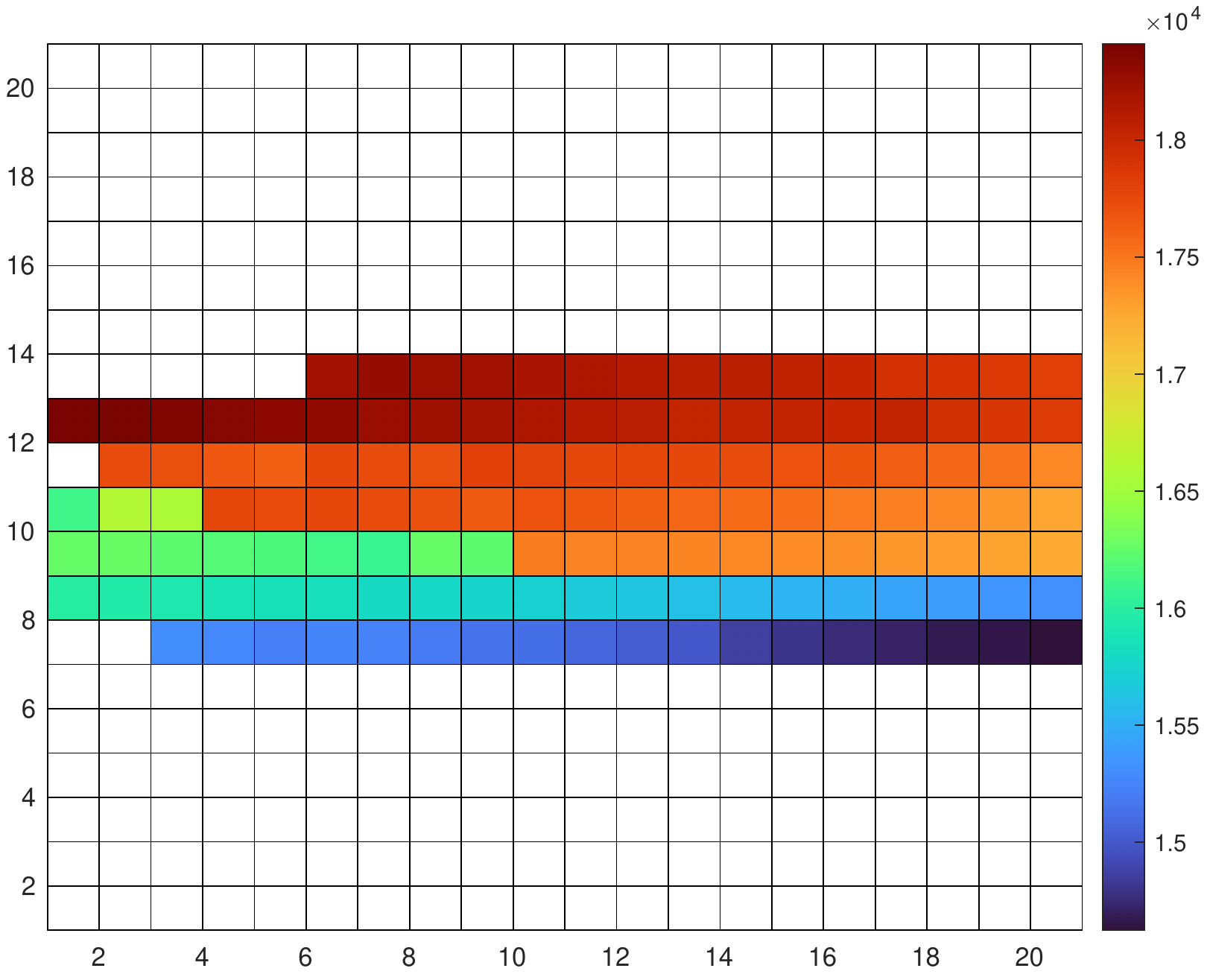}
\hskip5pt

% } % scalebox
% \squeezeup
% \squeezeup
% \squeezeup
% \squeezeup
\caption{The illustrations of solutions obtained by $(\mu+1)$EA and Map-elitism in the behavioral space on instance eil51\_n250\_uncorr-similar-weights\_01 (top), pr152\_n453\_uncorr\_01 (middle), and a280\_n279\_bounded-strongly-corr\_01 (bottom). The cells are coloured based on the TTP scores of the solutions in the cell over one single run. }
\label{fig:Maps_mu}
% \squeezeup
% \squeezeup
% \squeezeup
% \squeezeup
\end{figure*}

Figure \ref{fig:Maps_mu} compares MAP-Elitism and elitism in survival selection. The first three columns belong to $(\mu+1)$EA that show the initial population, all solutions generated during the search, and the final population, respectively. The fourth column illustrates the population obtained by BMBEA. One can observe that $(\mu+1)$EA converges to a single solution. Since the algorithm uses EAX crossover and DP as operators, it's impossible for the algorithm to find any other solutions from this point. On the other hand, the BMBEA's final population consists of a vast number of solutions with different properties. Thus, it can potentially find better-quality solutions if we continue the search. The other difference we can observe in Figure \ref{fig:Maps_mu} is that diversity of solutions decreases during the search $(\mu+1)$EA. On the contrary, the diversity increases using MAP-Elitism survival selection.     

\subsubsection{Maps With The Relaxed Approach}
\label{subsec:exp_map_relazed}
\begin{figure}
    \centering

    \includegraphics[width=0.95\textwidth]{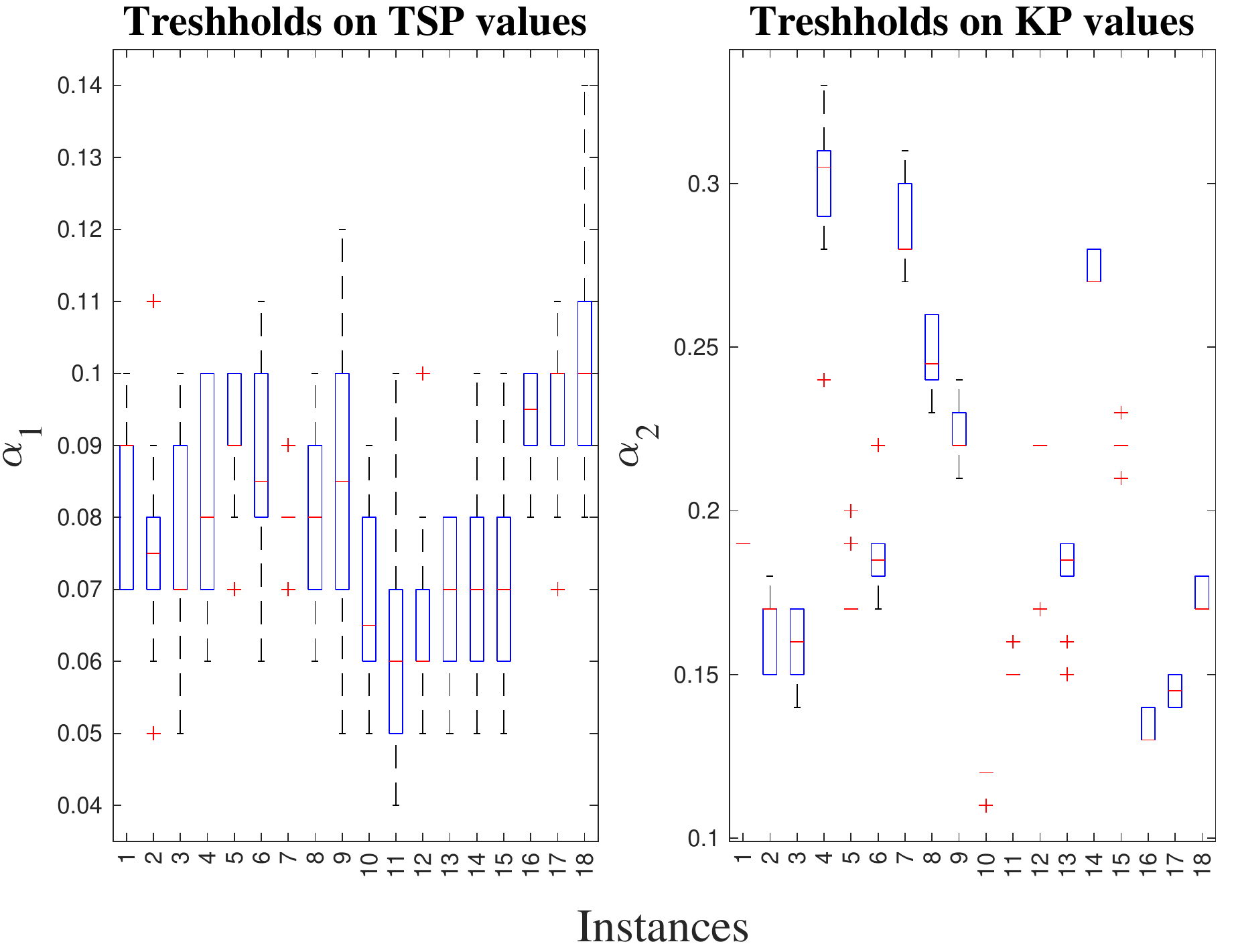}
   \caption{The values of $\alpha_1$ and $\alpha_2$ in the relaxed method}
   \label{fig:tresh}
\end{figure}

The maps obtained from the relaxed approach are similar to the prefixed method. The best quality solutions come from areas with similar distances to TSP and KP optimal values. However, $\alpha_1$ and $\alpha_2$ can differ in each run.
Figure \ref{fig:tresh} illustrates the $\alpha_1$ and $\alpha_2$ in 10 independent runs on instances 1 to 18. Here, after using EAX (\cite{nagata2013powerful}) to generate high-quality tours, we compute an optimal packing list for the tours by DP. Note that we used TSP optimal value as the termination criterion to boost time efficiency and diversity in tours. 
Figure \ref{fig:tresh} shows $\alpha_1$ and $\alpha_2$ belongs to $[0.04 \quad 0.14]$ and  $[0.11 \quad 0.33]$, respectively.
Since the TSP sub-problem is identical on instances 1 to 9, we can observe a similar trend for $\alpha_1$ on those instances. This argument is also true for instances 10 to 15 and 16 to 18. On the other hand, KP sub-problems are unique for all cases. Therefore, $\alpha_2$ is different in each instance. These observations can support our claim that $\alpha_1$ and $\alpha_2$ should be tuned for each instance separately in the prefixed method. 

\subsection{Best found TTP Solutions}

In this section, we analyse the performance of BMBEA in solving TTP. First, we look into the operators and draw a comparison in alternatively incorporation of the proposed operators into BMBEA. Then, we scrutinise the algorithm's survival selection by comparing BMBEA to $(\mu+1)$EA. Last, we study the method proposed for relaxing $\alpha_1$ and $\alpha_2$ and compared it to BMBEA with fixed  $\alpha_1$ and $\alpha_2$.

\begin{table*}[t]
\centering
\caption{Comparison of the search operators in terms of the TTP score and CPU time on the small size instances. In columns Stat the notation $X^+$ means the median of the measure is better than the one for variant $X$, $X^-$ means it is worse, and $X^*$ indicates no significant difference. Stat shows the results of Kruskal-Wallis statistical test at significance level $5\%$ and Bonferroni correction.}
% \begin{tiny}
\renewcommand{\tabcolsep}{2pt}
\renewcommand{\arraystretch}{0.5}
\begin{tabular}{l|cccc|cccc|c}
\toprule
              In. & EAX-DP  &(1) & &      & EAX-EA&(2) &  &        & Best-known\\
               & Average & Stat & Best & CPU time& Average & Stat & Best & CPU time   & value                   \\
\midrule
1&4267.1&$2^*3^+4^+$&\hl{4269.4}&86.5&4243.1&$1^*3^*4^+$&\hl{4269.4}&29.4&4269.4\\
2&7236&$2^*3^*4^+$&7252.8&2275&7086.6&$1^*3^*4^+$&7216.4&112.6&7532\\
3&11713&$2^*3^+4^+$&11733.9&18990.2&11621.1&$1^*3^*4^+$&11647.5&240.9&12804\\
4&1449.8&$2^*3^+4^+$&\hl{1460}&31.1&1443.6&$1^*3^+4^+$&\hl{1460}&26.9&1460\\
5&4250.1&$2^*3^+4^+$&4269.6&259.2&4235.1&$1^*3^*4^+$&4248.3&112.2&4365\\
6&5739.1&$2^*3^+4^+$&5792.2&884.3&5712.1&$1^*3^*4^+$&5792.2&219.7&6359\\
7&2808&$2^*3^+4^+$&2854.5&40.2&2781.2&$1^*3^*4^+$&2848.1&30.5&2871.1\\
8&6838.8&$2^*3^+4^+$&6884.4&456&6834.8&$1^*3^*4^+$&6884.4&98.7&7037\\
9&11753.2&$2^*3^*4^+$&11753.2&2185&11748.2&$1^*3^+4^+$&11753.2&214.7&12478\\
10&11140.5&$2^+3^+4^+$&\hl{11140.5}&3639.7&11113.3&$1^-3^*4^+$&11134.8&119.7&11117.4\\
11&25507.5&$2^*3^+4^+$&25525.5&224858.5&25148.2&$1^*3^+4^*$&25405.3&552.8&25664.4\\
12&3540.2&$2^*3^*4^+$&3669&253.6&3520.6&$1^*3^*4^+$&3665.4&112.8&3791.9\\
13&13374.8&$2^*3^*4^+$&\hl{13628.3}&5234.4&13210.6&$1^*3^*4^+$&13345.5&521.9&13556.9\\
14&5398.3&$2^*3^*4^+$&5398.3&386.8&5398.3&$1^*3^+4^+$&5398.3&99.5&5615\\
15&20456.8&$2^*3^+4^+$&20456.8&16047.8&20455.4&$1^*3^+4^+$&20456.8&476.2&20705.8\\
16&18449.6&$2^+3^*4^+$&18595.5&39024.7&18190.3&$1^-3^*4^*$&18244.5&254.5&19499\\
17&9163.3&$2^*3^+4^+$&9201.1&1496.5&9122&$1^*3^*4^+$&9176.8&241.9&9998\\
18&19419.2&$2^*3^+4^+$&19493.4&3588.3&19375.8&$1^*3^+4^+$&19495.4&215.8&20491\\
\midrule
              In. & 2-OPT-DP &(3) & &        & 2-OPT-EA &(4) & &        & Best-known \\
               & Average & Stat & Best & CPU time& Average & Stat & Best & CPU time   &                  value\\
\midrule
1&4208.3&$1^-2^*4^*$&4237.2&70.5&3969&$1^-2^-3^*$&4104.4&23.2&4296.4\\
2&7076&$1^*2^*4^+$&7252.8&2049.4&6510.7&$1^-2^-3^-$&6710.9&110.4&7532\\
3&11550.7&$1^-2^*4^*$&11647.5&16785&10516.4&$1^-2^-3^*$&10964.3&234&12804\\
4&1407&$1^-2^-4^*$&1428.6&20.1&1364.7&$1^-2^-3^*$&1409.7&21.5&1460\\
5&4185.9&$1^-2^*4^*$&4255.8&232&3903.8&$1^-2^-3^*$&4018.1&100.5&4365\\
6&5610.2&$1^-2^*4^*$&5699.2&797.9&5118.4&$1^-2^-3^*$&5277.1&219.1&6359\\
7&2663.1&$1^-2^*4^*$&2734.3&26.7&2550.6&$1^-2^-3^*$&2619.2&22.3&2871.1\\
8&6683&$1^-2^*4^*$&6858.8&375.4&6442&$1^-2^-3^*$&6688.4&97.6&7037\\
9&11480.3&$1^*2^-4^*$&11744.4&1956.5&10890.1&$1^-2^-3^*$&11157.8&205&12478\\
10&11124.4&$1^-2^*4^+$&11140.1&3070.9&10515.7&$1^-2^-3^-$&10623.3&98.4&11117.4\\
11&25138&$1^*2^*4^+$&25572.4&116932.7&22722.5&$1^-2^-3^-$&23539.2&553&25664.4\\
12&3627.5&$1^*2^*4^+$&3752.2&215.5&3266.5&$1^-2^-3^-$&3482.6&94.7&3791.9\\
13&13190.5&$1^*2^*4^+$&13560.9&4494.9&12407.3&$1^-2^-3^-$&12828.1&477.3&13556.9\\
14&5238.5&$1^*2^-4^*$&5397.9&321.9&5019.5&$1^-2^-3^*$&5319.3&86.3&5615\\
15&19725.3&$1^-2^-4^*$&20259.1&12978.6&18931.5&$1^-2^-3^*$&19764.6&474.9&20705.8\\
16&18273.2&$1^*2^*4^+$&18355.6&36393&17157.6&$1^-2^*3^-$&17307.9&244.2&19499\\
17&8972.7&$1^-2^*4^*$&9104.9&1479.1&8523.3&$1^-2^-3^*$&8714.4&217.9&9998\\
18&18901.1&$1^-2^-4^*$&19107.2&3370&18389.3&$1^-2^-3^*$&18763.1&209.9&20491\\
\bottomrule
\end{tabular}
% \end{tiny}
\label{tbl:Res_main}
\end{table*}
\begin{table}
\centering
\caption{Performance of the MAP-Elite based approach in terms of the TTP score. The notations are in line with Table \ref{tbl:Res_main}.}
\begin{small}
\renewcommand{\tabcolsep}{2pt}
\renewcommand{\arraystretch}{0.8}
\begin{tabular}{l|lccc|lccc|c}
\toprule
               In.& EAX-EA & (1)&   &    & 2-OPT-EA  &(2) &     &   &Best-known \\
               & Average & Stat & Best  & CPU time &Average & Stat & Best & CPU time  & value                 \\
\midrule
19&32625.3&$2^+$&\hl{33092}&835&29687.5&$1^-$&30065.8&728&32993.1\\
20&18975.9&$2^+$&19188.4&708&17622.2&$1^-$&17803.8&685&19379.7\\
21&35175.8&$2^+$&\hl{35512.2}&696&33456.6&$1^-$&34455.2&674&35015.2\\
22&642.3&$2^+$&\hl{1137.5}&1625&-4337&$1^-$&-2693.5&1696&893.4\\
23&51988.6&$2^+$&\hl{52651.8}&1624&48382.8&$1^-$&49830.4&1685&51303.4\\
24&29201.5&$2^+$&\hl{32072.9}&1627&25214.6&$1^-$&25618&1620&28304\\
25&104549.9&$2^+$&105434.5&7937&95300.4&$1^-$&96468.9&7975&105908.1\\
26&71829.9&$2^+$&\hl{73152.8}&6914&67954.6&$1^-$&69060.6&7285&72308.7\\
27&107975.3&$2^+$&\hl{109395.1}&6848&104852.4&$1^-$&106735&7091&108236.1\\
28&258901.5&$2^+$&260839.7&38669&238212.3&$1^-$&240916&45390&263040.2\\
29&129168.4&$2^+$&131072&36670&122606.8&$1^-$&123626.8&39520&131486.2\\
30&230888.9&$2^+$&\hl{237097.6}&32136&225694.2&$1^-$&227466.4&31796&233343\\
% 28&262758.6&$2^+$&\hl{264654.6}&244346.7&$1^-$&245597&263040.2\\
% 29&132208.5&$2^+$&\hl{132430}&127773&$1^-$&128588.1&131486.2\\
% 30&238596.9&$2^+$&\hl{241263.1}&234120.5&$1^-$&235595.3&233343\\

\bottomrule
\end{tabular}
\end{small}
\label{tbl:Res_large}
\end{table}
\begin{table}
\centering
\caption{Performance of the MAP-Elite based approach on the unbalanced instances. The notations are in line with Table \ref{tbl:Res_main}}
\begin{small}
\renewcommand{\tabcolsep}{2pt}
\renewcommand{\arraystretch}{0.7}
\begin{tabular}{l|lccc|lccc|c}
\toprule
               In. & EAX-EA &(1)& &       & 2-OPT-EA &(2) &        & &Best-known \\
               & Average & Stat & Best & CPU time & Average & Stat & Best & CPU time  & value                  \\
\midrule
31&-49282&$2^+$&\hl{-48622.8}&1555&-51704.2&$1^-$&-51635&1671&-49149.9\\
32&-7241&$2^+$&\hl{-4855}&1549&-10493&$1^-$&-9906.8&1709&-7714.6\\
33&-63137&$2^+$&-61797.3&1560&--66602&$1^-$&-63939&1709&-61709.1\\
34&-24508&$2^+$&-24263.5&1631&-27573&$1^-$&-26654&1684&-19215.2\\
\bottomrule
\end{tabular}
\end{small}
\label{tbl:Res_neg}
\end{table}
\subsubsection{Operators}
We now compare the search operators, EAX, 2OPT, DP, and $(1+1)$EA, in terms of the best-found TTP solution in this section. We consider instances in a range of 51 to 280 cities, and 50 to 453 items from \cite{PolyakovskiyB0MN14}. Table~\ref{tbl:Res_main} shows the average and the best TTP solutions and the average CPU time in ten independent runs for the four competitors and the best-known TTP values. The best-known values are obtained from \cite{abs-2011-05081}, and \cite{WuijtsT19}; both these papers compared their results to those of 21 algorithms analysed in \cite{WagnerLMNH18}. \citet{WagnerLMNH18} reported their results on all instances from Table \ref{tab:names}, while \citet{WuijtsT19} and \citet{abs-2011-05081} used some of the instances in their studies. The best-known values include \citet{abs-2011-05081} in instances 1 to 18, 22 to 24, and 28 to 34, and \citet{WuijtsT19} in instances 1 to 9, and 15 to 18. Note that our termination criterion differs from 10 minutes CPU time in \cite{WagnerLMNH18} and $2500$ local searches in \cite{WuijtsT19}. 
The results indicate that EAX outperforms 2-OPT in terms of TTP score in most cases. The observations are confirmed by a Kruskal-Wallis test at significance level $5\%$ and Bonferroni correction. Turning to the comparison of the KP operators, $(1+1)$EA yields very decent objective values and can compete with DP, which results in the optimal packing list.
In general, an increase in the size of instances severely affects the run time of the BMBEA using DP. On the other hand, the run times of $(1+1)$EA are significantly shorter. For example, the EAX-EA averagely finishes the 10000 iterations in 240.9 seconds on instance 3. The figure is about 18990.2 seconds for EAX-DP. This is while the algorithm's run time employing $(1+1)$EA remains reasonable. 
More Interestingly, Table~\ref{tbl:Res_main} also indicates that all variants of BMBEA result in very decent TTP scores. In instances 10 and 13, the introduced algorithms beat the best TTP scores and can hit the best known on instances 1 and 4.

Since the DP is not time-efficient in larger instances, we consider the $(1+1)$EA for computing the packing list. Table \ref{tbl:Res_large} shows the results on 12 instances from 575 to 4661 cities and 574 to 4460 items. As one can observe, EAX dominates 2-OPT in these instances. Moreover, the algorithm using EAX improved the best-found solution in 8 out of 12 cases. For example, the TTP score significantly increased from 893 to 1137.5 in instance 4. One can notice that the TTP score of the algorithm using 2-OPT is negative in this instance. \citet{PolyakovskiyB0MN14} balanced the instances in the TSP and the KP, but the TSP sub-problem is more dominating in some of the dsj1000 sub-group. The traveling cost is high in these particular instances, and the items do not compensate for the high cost. Having the TSP sub-problem more dominating, it is not surprising that the EAX outperforms the 2-OPT. Moreover, The domination is even stronger in four other instances of the dsj1000 sub-group in a way that the best-known values are negative. 
We investigate the four instances separated from the others due to the dominance of the TSP sub-problem over the KP.   
It means that the high-quality TTP solutions are closer to the TSP optimal value and more away from the KP optimal values. The current $\alpha_1$ and $\alpha_2$ are set for the balanced instances. Thus, we need to reset the $\alpha_1$ and $\alpha_2$ to populate the map. Based on initial experimental investigations, we set $\alpha_1$ and $\alpha_2$ to 2 and 60 percent, respectively.
Table~\ref{tbl:Res_neg} summarises the results on the four instances. The EAX, as expected, outperforms the 2-OPT in all four cases.
More importantly, the EAX-based algorithm improved the TTP values for instances 31, 32 by 1.1 and 37.1, respectively~\footnote{The TTP solutions can be accessed at https://github.com/NikfarjamAdel/Traveling-Thief-Problem}. 
Note that the results presented in Tables \ref{tbl:Res_main}, \ref{tbl:Res_large}, and \ref{tbl:Res_neg} are different from the ones presented in the conference version~(\cite{NikfarjamMap}) which are incorrect due to an implementation mistake.

\subsubsection{$(\mu+1)$EA vs. BMBEA}
\label{subsec:exp_bes_mu+1}
We now compare the BMBEA to $(\mu+1)$EA in solving the problem. As aforementioned, The difference between the two algorithms is in survival selection, where BMBEA select the next generation based on MAP-Elitism and $(\mu+1)$EA takes the most elite solutions. Table \ref{tbl:Res_muPone_DP} summarises the results of these algorithms where EAX and DP are considered as the operators. The table shows that the BMBEA has a higher average in 9 cases out of 18. The statistical test confirms a meaningful difference in the results in two instances favouring BMBEA. Both competitors result in the same average TTP score on four instances, and $(\mu+1)$EA brings about a higher mean on four instances. However, the statistical investigation finds no meaningful difference in those instances. Each algorithm outperforms the other in three cases regarding the best TTP score out of 10 runs. At the same time, they performed equally in the rest of the instances.

Figure \ref{fig:Maps_mu} shows that $(\mu+1)$EA converges to a single solution; therefore, there is no hope of finding better solutions by increasing the number of iterations. On the other hand, there is a good chance that an increase in the number of iterations results in better solutions for BMBEA. We investigate this by comparing the results in $10000$ and $100000$ iterations. Since DP can be time-consuming, we used $(1+1)$EA as the KP operator for this round of experiments. 
\begin{table}
\centering
\caption{
The comparison of $(\mu+1)EA$ and and Map-elite in solving TTP with using DP as KP operator. The notations are in line with Table \ref{tbl:Res_main}.}
\begin{small}
\renewcommand{\tabcolsep}{7pt}
\renewcommand{\arraystretch}{0.4}
\begin{tabular}{l|lcc|lcc}
\toprule
               In.& Map-Elitism & (1)&        & $(\mu+1)$  &(2) &        \\
               & Average & Stat & Best  & Average & Stat & Best                 \\
\midrule
1&\hl{4267.1}&$2^*$&4269.4&4264.9&$1^*$&4269.4\\
2&\hl{7236}&$2^+$&7252.8&7193.4&$1^-$&7252.8\\
3&\hl{11713}&$2^*$&11733.9&11693.5&$1^*$&11733.9\\
4&\hl{1449.8}&$2^*$&1460&1441.6&$1^*$&1460\\
5&\hl{4250.1}&$2^*$&4269.6&4243.8&$1^*$&4260.7\\
6&5739.1&$2^*$&5792.2&\hl{5761}&$1^*$&5792.2\\
7&\hl{2808}&$2^*$&2854.5&2785&$1^*$&2833.6\\
8&\hl{6838.8}&$2^*$&6884.4&6830.3&$1^*$&6884.4\\
9&11753.2&$2^*$&11753.2&11753.2&$1^*$&11753.2\\
10&11140.5&$2^*$&11140.5&11140.5&$1^*$&11140.5\\
11&25507.5&$2^*$&25525.5&\hl{25520.8}&$1^*$&25525.5\\
12&3540.2&$2^*$&3669&\hl{3549.4}&$1^*$&3687.1\\
13&\hl{13374.8}&$2^+$&13628.3&13243.4&$1^-$&13345.5\\
14&5398.3&$2^*$&5398.3&5398.3&$1^*$&5398.3\\
15&20456.8&$2^*$&20456.8&20456.8&$1^*$&20456.8\\
16&\hl{18449.6}&$2^*$&18595.5&18418.2&$1^*$&18444.1\\
17&9163.3&$2^*$&9201.1&\hl{9164.5}&$1^*$&9277.7\\
18&19419.2&$2^*$&19493.4&\hl{19447.7}&$1^*$&19507\\
\bottomrule
\end{tabular}
\end{small}
\label{tbl:Res_muPone_DP}
\end{table}

Table \ref{tbl:Res_muPone_EA} indicates results for BMBEA and $(\mu+1)$EA when the KP operator is altered to the $(1+1)$EA, and the number of iterations is equal to $10^3$ and $10^5$. The table shows that $(\mu+1)EA$ bring about a higher mean of TTP score in 10 cases compared to BMBEA when the termination criterion is set to $10^3$. The figure stands at 8 for BMBEA. The statistical investigation finds a meaningful difference in two instances in favour of $(\mu+1)EA$. One can notice the contradiction of this part of the table to Table \ref{tbl:Res_muPone_DP}. This is because the algorithms need more iterations to converge when we use $(1+1)$EA as the KP operator instead of DP. Giving more time to the algorithms can bring about similar results to Table \ref{tbl:Res_muPone_DP}. The other part of Table \ref{tbl:Res_muPone_EA} sheds light on this matter. 

We can see that BMBEA outperforms $(\mu+1)$ on 16 out of 18 instances n terms of average TTP score when the number of iterations is set to $10^5$. Also, a meaningful difference can be found in 12 instances, all favouring BMBEA. Increasing the number of iterations has no effect on the results of $(\mu+1)$EA since the algorithm converges within $10^3$ iterations. On the other hand, it considerably improves the BMBEA performance. This shows the efficiency of  MAP-Elitism in preserving diversity and preventing premature convergence.      
\begin{table}
\centering
\caption{
The comparison of $(\mu+1)EA$ and and Map-elite in solving TTP with using $(1+1)$EA as KP operator. The notations are in line with Table \ref{tbl:Res_main}.}
\begin{small}
\renewcommand{\tabcolsep}{2pt}
\renewcommand{\arraystretch}{0.8}
\begin{tabular}{l|lcc|lcc||lcc|lcc}
\toprule
                \multirow{3}{*}{In.}&\multicolumn{6}{c||}{$10^3$ iterations}&\multicolumn{6}{c}{$10^5$ iterations}\\ 
\cmidrule(l{2pt}r{2pt}){2-7}
\cmidrule(l{2pt}r{2pt}){8-13}
               & \multicolumn{3}{c|}{BMBEA (1)}& \multicolumn{3}{c||}{($\mu+1$) (2)} &\multicolumn{3}{c|}{BMBEA (1)}& \multicolumn{3}{c}{($\mu+1$) (2)}\\
               & Average & Stat & Best  & Average & Stat & Best & Average & Stat & Best  & Average & Stat & Best          \\
\midrule
1&\hl{4243.1}&$2^*$&4269.4&4188.9&$1^*$&4269.4&\hl{4269.4}&$2^+$&4269.4&4188.9&$1^-$&4269.4\\
2&\hl{7086.6}&$2^*$&7216.4&7067.4&$1^*$&7231.1&\hl{7202.5}&$2^+$&7252.8&7067.4&$1^-$&7231.1\\
3&11621.1&$2^*$&11647.5&\hl{11637.2}&$1^*$&11689.2&\hl{11651.9}&$2^*$&11733.9&11637.8&$1^*$&11695.5\\
4&\hl{1443.6}&$2^+$&1460&1434.2&$1^-$&1447.5&\hl{1458.9}&$2^+$&1460&1434.2&$1^-$&1447.5\\
5&\hl{4235.1}&$2^*$&4248.3&4202.2&$1^*$&4246.7&\hl{4255.2}&$2^+$&4274.5&4202.2&$1^-$&4246.7\\
6&\hl{5712.1}&$2^*$&5792.2&5663.4&$1^*$&5729.2&\hl{5776.6}&$2^+$&5792.2&5663.4&$1^-$&5729.2\\
7&\hl{2781.2}&$2^*$&2848.1&2748.8&$1^*$&2844.7&\hl{2804.7}&$2^*$&2854.5&2748.8&$1^*$&2844.7\\
8&\hl{6834.8}&$2^+$&6884.4&6776.9&$1^-$&6843.4&\hl{6858.8}&$2^+$&6884.4&6776.9&$1^-$&6843.4\\
9&\hl{11748.2}&$2^*$&11753.2&11641.9&$1^*$&11753.2&\hl{11749}&$2^+$&11753.2&11641.9&$1^-$&11753.2\\
10&11113.3&$2^-$&11134.8&\hl{11133.1}&$1^+$&11137.9&\hl{11137.7}&$2^+$&11140.4&11133.1&$1^-$&11137.9\\
11&25148.2&$2^*$&25405.3&\hl{25273.3}&$1^*$&25484&\hl{25396.9}&$2^+$&25524.3&25273.8&$1^-$&25484\\
12&\hl{3520.6}&$2^*$&3665.4&3431.2&$1^*$&3547.6&\hl{3528.3}&$2^+$&3687.1&3431.2&$1^-$&3547.6\\
13&13210.6&$2^*$&13345.5&\hl{13278.5}&$1^*$&13345.5&\hl{13326.6}&$2^*$&13345.5&13278.5&$1^*$&13345.5\\
14&5398.3&$2^*$&5398.3&5398.3&$1^*$&5398.3&5398.3&$2^*$&5398.3&5398.3&$1^*$&5398.3\\
15&\hl{20455.4}&$2^*$&20456.8&20267.7&$1^*$&20456.8&\hl{20456.8}&$2^*$&20456.8&20267.7&$1^*$&20456.8\\
16&18190.3&$2^-$&18244.5&\hl{18383.8}&$1^+$&18409.3&18381&$2^*$&18438.6&\hl{18392.9}&$1^*$&18409.3\\
17&9122&$2^*$&9176.8&\hl{9133.2}&$1^*$&9203.1&\hl{9249.3}&$2^+$&9331.3&9133.2&$1^-$&9203.1\\
18&\hl{19375.8}&$2^*$&19495.4&19359.9&$1^*$&19575.5&\hl{19559}&$2^+$&19738.3&19359.9&$1^-$&19575.5\\

\bottomrule
\end{tabular}
\end{small}
\label{tbl:Res_muPone_EA}
\end{table}
\subsubsection{Relaxed Method}
\label{subsec:exp_bes_relaxed}
Tuning $\alpha_1$ and $\alpha_2$ requires a lot of computational effort. It should be done for each instance separately to make sure the area of focus in the feature space is not infeasible and promising. As we observed in the previous section, the initial values ($\alpha_1 = 0.05$,  and $\alpha_2 = 0.2$) result in an infeasible focused area for instances 31 to 34. For this purpose,  we proposed the relax method where $\alpha_1$ and $\alpha_2$ are set based on the initial solutions. Here, we investigate the impact of this method on the performance of BMBEA. To do so, we compare the BMBEA using the relaxed method with the prefixed $\alpha_1$ and $\alpha_2$. 

Table \ref{tbl:Res_Chng_EA} summarise the experimental investigation. EAX and $(1+1)$EA are considered for the operators. When the termination criterion is set to $10^3$ iterations, the prefixed method performs statistically better in 4 out of 27 cases. A meaningful difference can be found in one instance in favour of the relaxed method. At the same time, there is no statistically significant difference in the rest of the cases. However, the average TTP score of the relaxed method is higher in four instances. We can observe that this method performs better in experiments with longer runs. It has a higher average TTP score in 19 cases, while the prefixed approach outperforms it in 5 instances. Both algorithms perform equally in three cases. Statistical tests confirm significant differences in the two instances for the relaxed approach. We can conclude that the relaxed method can result in a decent TTP score. At the same time, it eliminates the need to tune two influential parameters. It is noteworthy that both algorithms can beat the best-known TTP values in instances 20 and 25 in the longer runs.
\begin{table}
\centering
\caption{
The comparison of the Relaxed method with the prefixed BMBEA in solving TTP with using $(1+1)$EA as KP operator. The notations are in line with Table \ref{tbl:Res_main}.}
\begin{small}
\renewcommand{\tabcolsep}{2pt}
\renewcommand{\arraystretch}{0.8}
\begin{tabular}{l|lcc|lcc||lcc|lcc}
\toprule
                \multirow{3}{*}{In.}&\multicolumn{6}{c||}{$10^3$ iterations}&\multicolumn{6}{c}{$10^5$ iterations}\\ 
\cmidrule(l{2pt}r{2pt}){2-7}
\cmidrule(l{2pt}r{2pt}){8-13}
               & \multicolumn{3}{c|}{Prefixed (1)}& \multicolumn{3}{c||}{Relaxed (2)} &\multicolumn{3}{c|}{Prefixed (1)}& \multicolumn{3}{c}{Relaxed (2)}\\
               & Average & Stat & Best  & Average & Stat & Best & Average & Stat & Best  & Average & Stat & Best          \\
\midrule
1&4243.1&$2^*$&4269.4&\hl{4246}&$1^*$&4269.4&4269.4&$2^*$&4269.4&4269.4&$1^*$&4269.4\\
2&7086.6&$2^*$&7216.4&\hl{7116.7}&$1^*$&7231.&\hl{7202.5}&$2^*$&7252.8&7195.4&$1^*$&7252.8\\
3&\hl{11621.1}&$2^*$&11647.5&11619.5&$1^*$&11697.5&11651.9&$2^*$&11733.9&\hl{11667.9}&$1^*$&11733.3\\
4&\hl{1443.6}&$2^*$&1460&1437.1&$1^*$&1445.8&\hl{1458.9}&$2^*$&1460&1458.7&$1^*$&1460\\
5&4235.1&$2^*$&4248.3&4235.8&$1^*$&4260.&4255.2&$2^*$&4274.5&\hl{4264.2}&$1^*$&4286.3\\
6&\hl{5712.1}&$2^+$&5792.2&5655&$1^-$&5729.&\hl{5776.6}&$2^*$&5792.2&5762.1&$1^*$&5792.2\\
7&2781.2&$2^*$&2848.1&\hl{2799.1}&$1^*$&2840&2804.7&$2^*$&2854.5&\hl{2847.5}&$1^*$&2854.5\\
8&\hl{6834.8}&$2^*$&6884.4&6833.8&$1^*$&6884.4&6858.8&$2^-$&6884.4&\hl{6884.4}&$1^+$&6884.4\\
9&\hl{11748.2}&$2^*$&11753.2&11744.7&$1^*$&11753.&11749&$2^*$&11753.2&\hl{11753.2}&$1^*$&11753.2\\
10&\hl{11113.3}&$2^*$&11134.8&11103&$1^*$&11135.4&\hl{11137.7}&$2^*$&11140.4&11136.9&$1^*$&11137.9\\
11&\hl{25148.2}&$2^*$&25405.3&25036.9&$1^*$&25332.2&25396.9&$2^*$&25524.3&\hl{25441}&$1^*$&25525.5\\
12&\hl{3520.6}&$2^*$&3665.4&3487.4&$1^*$&3547.6&3528.3&$2^*$&3687.1&\hl{3564.7}&$1^*$&3789.8\\
13&13210.6&$2^*$&13345.5&\hl{13270}&$1^*$&13345.5&13326.6&$2^*$&13345.5&\hl{13345.5}&$1^*$&13345.5\\
14&\hl{5398.3}&$2^+$&5398.3&5387.6&$1^-$&5398.3&5398.3&$2^*$&5398.3&5398.3&$1^*$&5398.3\\
15&\hl{20455.4}&$2^*$&20456.8&20421.6&$1^*$&20456.8&20456.8&$2^*$&20456.8&20456.8&$1^*$&20456.8\\
16&\hl{18190.3}&$2^*$&18244.5&18187.2&$1^*$&18330.5&18381&$2^*$&18438.6&\hl{18388.1}&$1^*$&18472\\
17&\hl{9122}&$2^+$&9176.8&9074.2&$1^-$&9143.1&\hl{9249.3}&$2^*$&9331.3&9237.7&$1^*$&9294.1\\
18&\hl{19375.8}&$2^*$&19495.4&19247&$1^*$&19432.&19559&$2^*$&19738.3&\hl{19573.8}&$1^*$&19628.9\\
19&\hl{32625.3}&$2^*$&33092&32354.2&$1^*$&33080&33358.4&$2^*$&33759&\hl{33496.8}&$1^*$&34007\\
20&\hl{18975.9}&$2^*$&19188.4&18965.4&$1^*$&19092.8&19279.8&$2^*$&19538.3&\hl{19423.5}&$1^*$&19650.7\\
21&\hl{35175.8}&$2^*$&35512.2&35101.1&$1^*$&35429.5&35720.2&$2^*$&36021.4&\hl{35762.2}&$1^*$&36133.8\\
22&\hl{642.3}&$2^*$&1137.5&541.5&$1^*$&879.3&1535.5&$2^-$&2621.1&\hl{2356.1}&$1^+$&2964\\
23&\hl{51988.6}&$2^*$&52651.8&51885.5&$1^*$&52345.7&52292.8&$2^*$&52973.6&\hl{52954.2}&$1^*$&55260.9\\
24&\hl{29201.5}&$2^*$&32072.9&28711.7&$1^*$&29299.3&29492.5&$2^*$&32085.1&\hl{29500.7}&$1^*$&30411\\
25&\hl{104549.9}&$2^+$&105434.5&103863.4&$1^-$&105025&105442.9&$2^*$&106465.7&\hl{106146.9}&$1^*$&107250.6\\
26&\hl{71829.9}&$2^*$&73152.8&71615.1&$1^*$&73141&72110.7&$2^*$&73348.2&\hl{72149.4}&$1^*$&73291.9\\
27&107975.3&$2^-$&109395.1&\hl{108885.5}&$1^+$&109929&109402.2&$2^*$&110346&\hl{109846}&$1^*$&110681\\

\bottomrule
\end{tabular}
\end{small}
\label{tbl:Res_Chng_EA}
\end{table}
\section{Conclusion}
In this study, we incorporated the concept of QD into solving the TTP. To the best of our knowledge, this is the first time the QD concept has been used to solve a combinatorial problem. The behaviour descriptor for our approach is defined on the TSP and the KP scores of a TTP solution. Having described a 2D MAP-Elite based survival selection, we introduced the BMBEA algorithm to generate high-quality TTP solutions. BMBEA involves EAX crossover to create new tours. Afterwards, the algorithm computes a high-quality packing list by dynamic programming or the (1+1)~EA. By visualising the map obtained from BMBEA, we observed the distribution of high-performing TTP solutions over the behavioural space of TSP and KP.
Moreover, we conducted a comprehensive experimental comparison involving four different search operators for BMBEA. Moreover, we investigated the impact of Map-elitism on the final solutions by comparing it to a simple $(\mu+1)$EA. The results indicated that MAP-elitism boosts both the diversity and quality of solutions. 

It would be interesting to incorporate more complex MAP-Elite approaches such as CVT-MAP-Elites \cite{DBLP:VassiliadesCM18} into the introduced algorithm. Using such an approach can discretise the behavioural space more intelligently. Moreover, several multi-component combinatorial optimisation problems can be found in literature where QD is highly beneficial to understanding the inter-dependencies of components and the distribution of solutions in the behavioural space.

\section{Acknowledgements}
This work has been supported by the Australian Research Council (ARC) through grants DP190103894, FT200100536, and by the South Australian Government through the Research Consortium “Unlocking Complex Resources through Lean Processing”.

\bibliographystyle{abbrvnat}
\bibliography{arxiv}
\pagebreak

\end{document}